\documentclass[opre,nonblindrev]{informs3} 

\DoubleSpacedXI 

\usepackage{subfigure}
\usepackage{latexsym}
\usepackage{mathptmx}
\usepackage{amsfonts,amsbsy}
\usepackage{mathrsfs,booktabs,mdwlist,multirow,colortbl,bm}
\usepackage{multicol}
\usepackage{algorithmic,algorithm}

\usepackage{natbib}
 \bibpunct[, ]{(}{)}{,}{a}{}{,}%
 \def\bibfont{\small}%
\TheoremsNumberedThrough     
\ECRepeatTheorems

\EquationsNumberedThrough    

\begin{document}


\RUNAUTHOR{Jinyang Jiang, Jiaqiao Hu, and Yijie Peng}

\RUNTITLE{Quantile-Based Deep Reinforcement Learning}

\TITLE{Quantile-Based Deep Reinforcement Learning using Two-Timescale Policy Gradient Algorithms}

\ARTICLEAUTHORS{%
\AUTHOR{Jinyang Jiang}
\AFF{Department of Management Science and Information Systems,\\ Guanghua School of Management, Peking University, Beijing 100871, CHINA, \EMAIL{jinyang.jiang@stu.pku.edu.cn}}
\AUTHOR{Jiaqiao Hu}
\AFF{Department of Applied Mathematics and Statistics,\\
State University of New York at Stony Brook, Stony Brook, NY 11794, U.S.A., \EMAIL{jqhu@ams.stonybrook.edu}}
\AUTHOR{Yijie Peng}
\AFF{Department of Management Science and Information Systems,\\ Guanghua School of Management, Peking University, Beijing 100871, CHINA, \EMAIL{pengyijie@pku.edu.cn}}
} 

\ABSTRACT{%
Classical reinforcement learning (RL) aims to optimize the expected cumulative reward. In this work, we consider the RL setting where the goal is to optimize the quantile of the cumulative reward. We parameterize the policy controlling actions by neural networks, and propose a novel policy gradient algorithm called Quantile-Based Policy Optimization (QPO) and its variant Quantile-Based Proximal Policy Optimization (QPPO) for solving deep RL problems with quantile objectives. QPO uses two coupled iterations running at different timescales for simultaneously updating quantiles and policy parameters, whereas QPPO is an off-policy version of QPO that allows multiple updates of parameters during one simulation episode, leading to improved algorithm efficiency. Our numerical results indicate that the proposed algorithms outperform the existing baseline algorithms under the quantile criterion. 
}%



\maketitle

%


\section{INTRODUCTION}
\begin{figure}[t]
	\centering
	\includegraphics[scale=0.48]{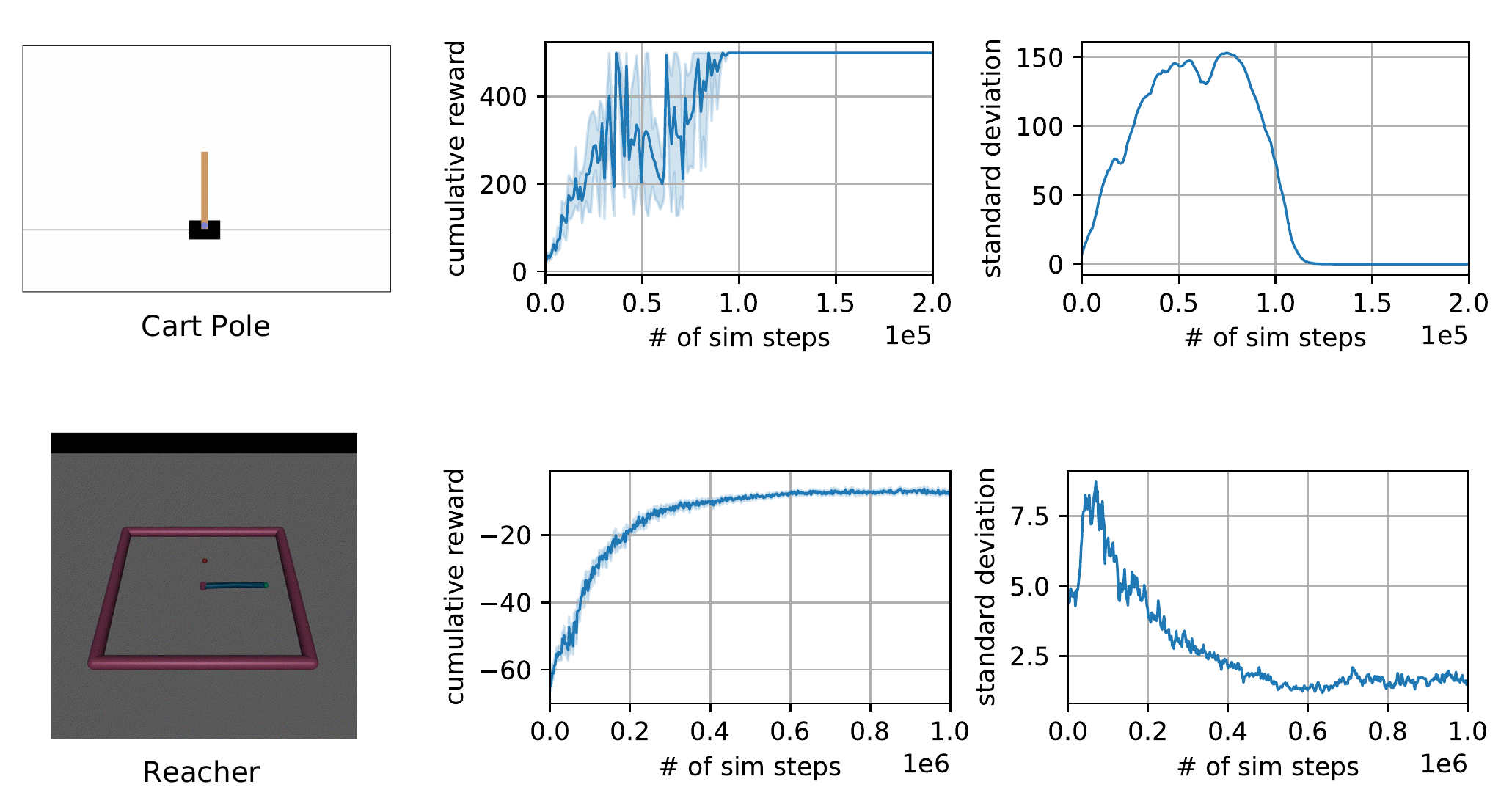}
	\caption{Learning curves of cumulative reward and its standard deviation in single RL training experiment for robotic control tasks. The agent's performance is evaluated by 50 simulation episodes for Cart Pole and 25 for Reacher.}
	\label{fig:classical_rl}
\end{figure}

Deep reinforcement learning (RL) has recently made significant successes in games \citep{mnih2015human,silver2016mastering}, robotic control \citep{levine2016end}, recommendation \citep{zheng2018drn} and many other fields. RL formulates a complex sequential decision-making task as a Markov Decision Process (MDP) and attempts to learn an optimal policy by interacting with the environment via sampling rewards and state transitions. In the classical RL framework, the goal is to optimize the expectation/mean of the cumulative reward. 
Under such a criterion, the optimal policy of the MDP satisfies the well-known Bellman equation, which forms the basis for many popular RL algorithms.
In application domains such as robotic control and board games, where RL achieves great successes, the reward functions and the underlying system dynamics are often deterministic, and random transitions are merely introduced artificially to characterize complex physical environments. For these settings, optimizing the expected reward turns out to be a desirable goal because a well-trained policy would typically result in the underlying dynamics being acted in a nearly deterministic fashion, leading to a cumulative reward that contains little or no variation; see Figure.\ref{fig:classical_rl} for an illustration.

The output of a business system, however, is usually the random outcome of group human behaviors, which could be extremely difficult to predict. 
For instance, in inventory management, customer demands, arrival times, and order lead times are intrinsically  random, so even under the (mean-based) optimal ordering policy, the inventory cost may still be subject to large uncertainty.
In such cases, it is important to question whether the mean is an appropriate objective for optimization because mean only measures the average performance of a random system but not its extreme or ``tail'' behavior. The tail performance may actually reflect a catastrophic outcome for a system. For example, the joint defaults of subprime mortgages led to the 2008 financial crisis, and in the post-crisis era, the Basel accord requires major financial institutes to maintain a minimal capital level for sustaining the loss under extreme market circumstances. A quantile, on the other hand, can be used to capture the tail behavior of a random system, and thus could be a more useful alternative in applications involving risk minimization or the prevention of damages/losses caused by the occurrence of extreme events. For example, managers in service industry may try to optimize the $0.01$ quantile level of system reliability that guarantees the service for $99\%$ of the customers \citep{decandia2007dynamo}; physicians may want to make a treatment decision that maximizes the $0.1$ percentile of health improvement rate \citep{beyerlein2014quantile}. In finance, quantiles are also known as value-at-risk (VaR) and can be directly translated into the minimal capital requirement.

\begin{figure*} [h]
	\centering
	\subfigure[]{
		\includegraphics[scale=0.42]{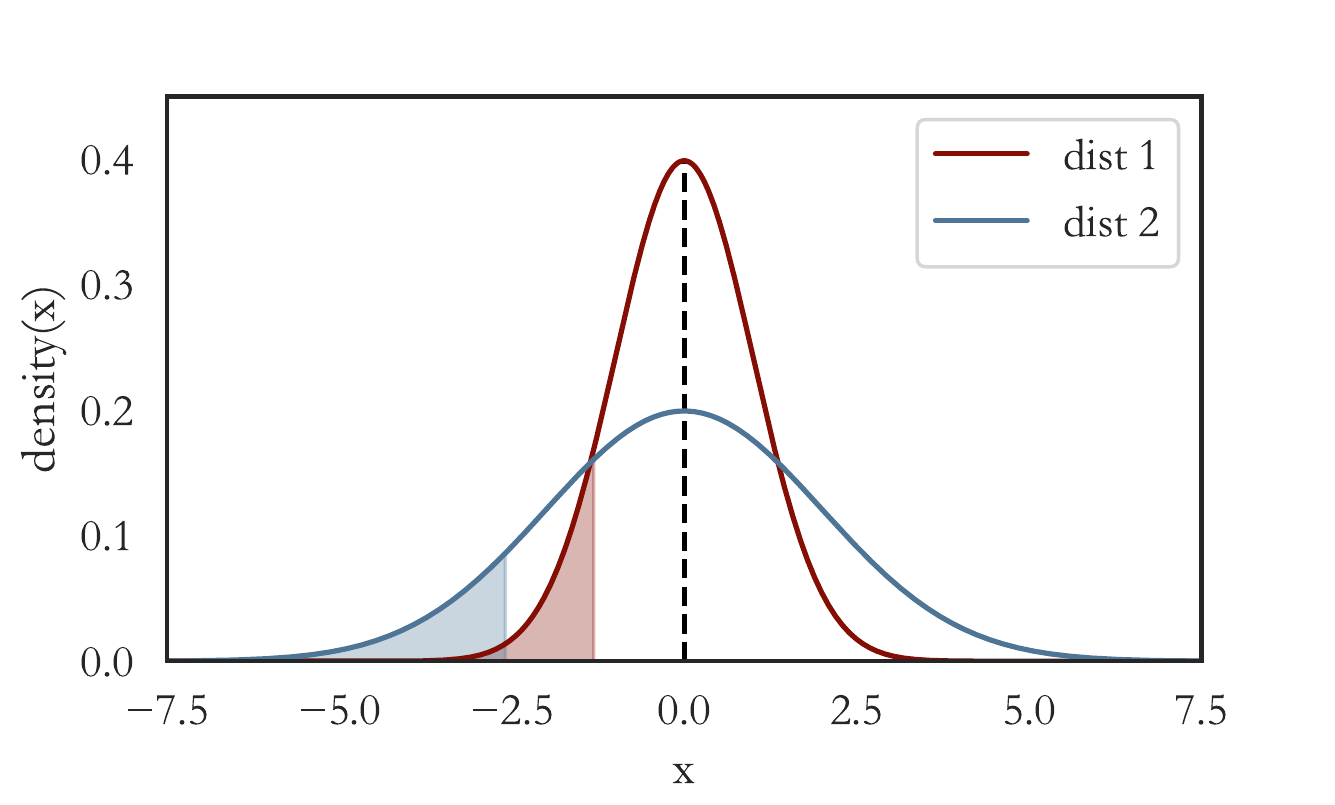}
            \label{fig:quantile_mean}}
	\subfigure[]{
		\includegraphics[scale=0.42]{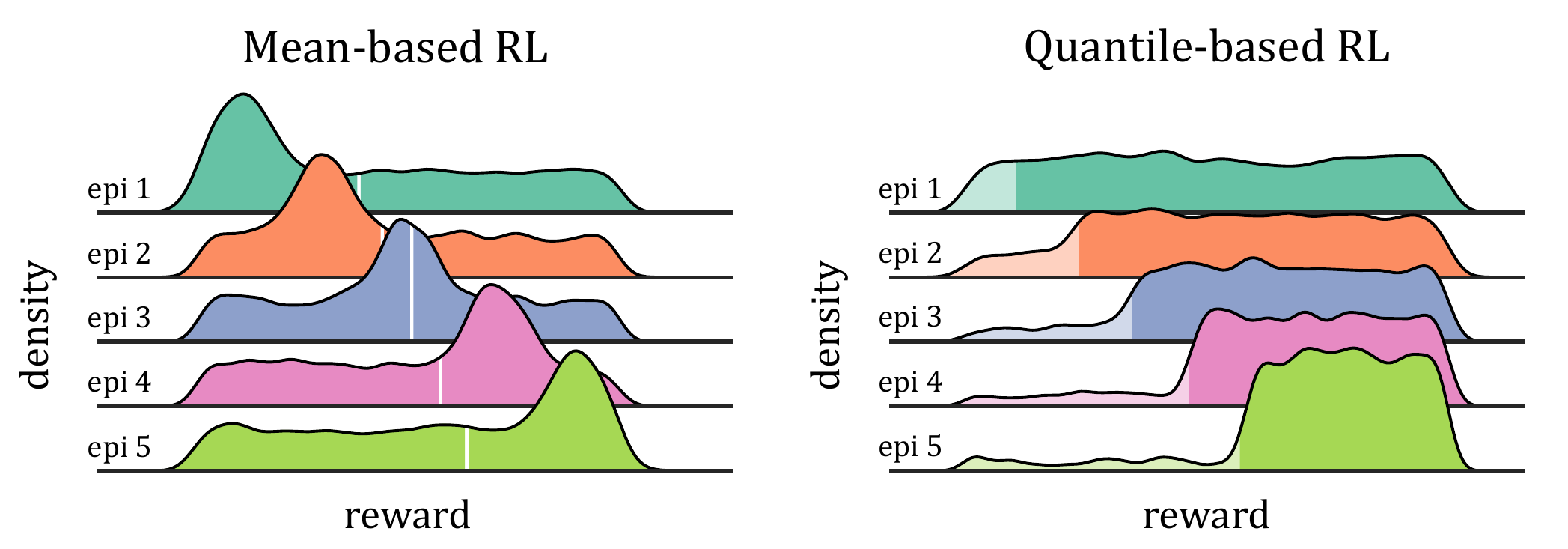}
            \label{fig:quantile_mean_opt}}
	\caption{(a) Probability density plots of two normal distributions $\mathcal{N}(0,1)$ and $\mathcal{N}(0,2)$. The right boundaries of shaded areas represent the value of $10\%$-quantile. (b) Comparison of mean-based RL and quantile-based RL.}
\end{figure*}

In this paper, we consider the setting where the goal is to optimize the quantiles of the cumulative rewards. As a simple illustration of the difference between mean and quantile,  Figure.\ref{fig:quantile_mean} shows the return distributions of two portfolios. It can be clearly observed that although both distributions have the same mean, the $10\%$-quantile of distribution 1 is significantly larger than that of distribution 2, meaning that portfolio 2 would require more capital to avoid irreversible loss or bankruptcy at the $90\%$ confidence level. The difference between these two criteria in the RL setting is also highlighted in Figure.\ref{fig:quantile_mean_opt}: quantile optimization improves the tail performance while mean optimization improves the average value.
For some distributions, e.g. the Cauchy distribution, the mean does not even exist and hence cannot be used as a meaningful performance measure. In contrast, quantiles are always well-defined.

Unfortunately, quantile measures are known to be extremely difficult to optimize \citep{rockafellar2020risk}, and solving MDPs with quantile objectives is even more challenging because unlike in the mean case, a quantile of the cumulative reward
cannot be decomposed into the sum of quantiles of rewards at different decision epochs, making it impossible to apply conventional RL methods rooted in the Bellman equation. This issue does not occur under nested risk measures \citep{ruszczynski2010risk,jiang2018risk} or other measures that allow for a recursive structure decomposition. For MDPs under the conditional value-at-risk (CVaR) criterion, some recent work \citep{chow2013stochastic,chow2014algorithms} considers the use of state-space augmentation techniques. However, none of these techniques can be suitably applied to the quantile setting. More recently,  \cite{li2022quantile} also propose a dynamic programming algorithm for quantile-based MDP, but their work requires explicit specification of the model parameters and does not fit into the RL context considered in this paper.

We parameterize the policy controlling actions by neural networks. When the parameterization is smooth but entails a large number of parameters, it is natural to adopt a gradient descent-based technique for solving the nonlinear optimization problem required in policy training.
However, the difficulty is that, in contrast to the gradient of a mean, which can be estimated without bias using a single sample/simulation trajectory, the quantile gradient relies on the true quantile value and hence does not allow for an unbiased estimation. To address this issue, we propose two novel policy gradient algorithms. The first algorithm, called Quantile-Based Policy Optimization (QPO), uses two coupled stochastic approximation (SA) iterations running at different timescales for simultaneously computing quantiles and policy parameters, allowing quantile estimation and parameter optimization to be conducted in a coherent manner. Each update of policy parameters in QPO requires a single sample trajectory (episode) of the system.
For MDPs with complicated system dynamics and/or long planning horizons, updating parameters once at the end of an episode may not be practical if observing/simulating the entire trajectory is computationally demanding. The second algorithm draws ideas from the state-of-art mean-based RL techniques, in particular, the Proximal Policy Optimization (PPO) approach of Schulman et al. (2017), yielding an off-policy variant of QPO we call Quantile-Based Proximal Policy Optimization (QPPO) that allows multiple updates of policy parameters during one simulation episode. This, in effect, leads to improved efficiency in data utilization and significantly enhanced algorithm convergence behavior. Our empirical study on several business applications indicates that the proposed algorithms are promising and outperform some of the existing algorithms under the quantile criterion.

We summarize our main contributions as follows.
\begin{itemize}
    \item We derive a quantile-based policy gradient estimator, propose a two-timescale policy gradient algorithm, QPO, for quantile optimization in the deep RL setting, and establish the strong convergence and rates of convergence of the algorithm.
    \item We introduce an enhanced version of QPO that allows multiple updates of policy parameters during a single episode, show its convergence, and provide new error bounds to  characterize its performance.
    \item We carry out simulation studies to empirically investigate and compare the performance of our new algorithms with existing baseline techniques on realistic business applications, including financial investment and inventory management problems.
\end{itemize}

To the best of our knowledge, our work is the first to develop quantile-based deep RL algorithms capable of training policies parameterized by large-scale neural networks. Preliminary versions of the two algorithms, QPO and QPPO, have been presented in the conference paper \cite{jiang2022quantile} but without convergence analysis. In this work, in addition to providing complete convergence proofs, we perform detailed analysis to characterize the performance of the algorithms in terms of convergence rates/error bounds and conduct more comprehensive numerical experiments to illustrate the algorithms on realistic applications.

The rest of the paper is organized as follows. In Section 2, we review the related work on gradient estimation and deep RL. In Section 3, we describe the MDP problem and present our
quantile-based RL optimization model. Section 4 contains a detailed description of the proposed QPO algorithm, accompanied by its theoretical convergence and rate results. In Section 5, we introduce an accelerated variant of QPO and further provide error bounds on its performance. Simulation studies are carried out in Section 6, and we conclude the paper and discuss future directions in Section 7. The proofs of all theoretical results are given in the online appendix.

\section{RELATED WORK}
\subsection{Mean-based RL}

In mean-based RL, there are two major categories of methods. The first category is value-based methods, which learn the value/Q-function and rely on the Bellman equation to select the action with the best value. Pioneer work in deep RL includes Deep Q-learning (DQN) and its variants \citep{mnih2015human,hessel2018rainbow}.  The second category of methods are based on policy optimization, where the policies are parameterized in various ways, e.g., through basis functions or neural networks, and optimized by stochastic gradient descent. REINFORCE is an early policy gradient algorithm \citep{williams1992simple}.
Trust Region Policy Optimization (TRPO) introduces importance sampling techniques into RL to increase data utilization efficiency \citep{schulman2015trust}.
PPO further improves upon TRPO through optimizing a clipped surrogate objective and has currently become a commonly used baseline algorithm \citep{schulman2017proximal}.
There are also algorithms that combine the advantages of these two types of methods such as Deep Deterministic Policy Gradient (DDPG) and Soft Actor-Critic (SAC) \citep{Lillicrap,haarnoja2018soft}.

Recently, mean-based RL has been studied actively in the domain of operations research and management science. On the application side, there is a growing literature applying RL techniques to problems in fields such as market making \citep{baldacci2022market}, business management \citep{moon2022manufacturing}, consumer targeting \citep{wang2022deep}, network optimization \citep{qu2022scalable}, and collusion avoidance \citep{abada2020artificial}. In terms of theory,
\cite{sinclair2022adaptive} improve the training cost of value-based online RL algorithms through adaptive discretization of problem space; \cite{wang2022reliable} derive confidence bounds for off-policy evaluation
by leveraging methodologies from distributionally robust optimization; other recent advancements on important concepts in RL can be found in, e.g., \cite{cen2022fast, bhandari2021finite}.

\subsection{Gradient Estimation of Risk Measures}
In classical gradient estimation problems, the focus has been on mean-based performance measures \citep{fu2006gradient}, whereas gradient estimation of risk measures such as quantiles and CVaRs is considered to be more difficult and is an area of active research. A number techniques have been developed over the past two decades, including infinite perturbation analysis \citep{hong2009estimating,jiang2015estimating}, kernel estimation \citep{liu2009kernel,hong2009simulating}, and measure-valued differentiation \citep{heidergott2016measure}.
Recently, \cite{glynn2021computing} also consider the use of the generalized likelihood ratio method for estimating the gradient of a general class of distortion risk measures.
Unfortunately, applying these methods in an MDP environment would require the analytic expressions for the transition and reward functions, which are typically not available in a RL setting.
A well-known black-box gradient estimation approach is Simultaneous Perturbation Stochastic Approximation (SPSA) \citep{spall1992multivariate}. However, tuning parameters such as perturbation- and step-sizes in optimization requires care and it could be hard to apply SPSA to high-dimensional optimization problems in deep RL when policies are parameterized by neural networks.

\subsection{Risk-sensitive RL}
Risk measures have been introduced into RL either in the forms of objectives or constraints, which are referred as risk-sensitivity RL in the literature. With risk measures as the objective functions, well-trained agents can be expected to perform more robustly under extreme events.
For example, an expected exponential utility approach is taken by \cite{borkar2001sensitivity}; \cite{petrik2012approximate} and \cite{tamar2014policy} study CVaR-based objectives;
\cite{prashanth2013actor} aim to optimize several
variance-related risk measures using SPSA and a smooth function approach;
\cite{prashanth2016cumulative} apply SPSA to optimize an objective function in the cumulative prospect theory. For a comprehensive discussion on policy optimization under various risk measures, we refer the reader to \cite{prashanth2022risk}.

There are also studies that incorporate a risk measure as the constraint to an RL problem.
For example, a Lagrangian approach has been used in \cite{bertsekas1997nonlinear} to solve RL problems subject to certain risk measure constraints; dynamic and time-consistent risk constraints are considered in \cite{chow2013stochastic}; \cite{borkar2014risk} use CVaR as the constraint; \cite{chow2017risk} also develop policy gradient and actor-critic algorithms under VaR and CVaR constraints.

\section{Problem Formulation \& Preliminaries}\label{Section Problem Formulation and Preliminaries}
\subsection{Markov Decision Process}
An MDP can be defined as a 5-tuple $(\mathcal{S},\mathcal{A},p,u,\eta)$, where $\mathcal{S}$ and $\mathcal{A}$ are the state and action spaces, $p(s'|s,a)$ is the transition probability, $u(s',a,s)$ is the reward function, and $\eta\in(0,1)$ is the reward discount factor. We consider the class of stationary, randomized, Markovian policies parameterized by a vector $\theta$, where each $\pi(\cdot|s;\theta)$ represents a probability distribution over $\mathcal{A}$ for every $s\in\mathcal{S}$.
Denote the state and action encountered at time $t\in\{0,1,2,\cdots,T\}$ by $s_t$ and $a_t$, where $s_t\sim p(\cdot |s_{t-1}, a_{t-1})$, $a_t\sim\pi(\cdot| s_{t};\theta)$, and $T$ is the decision horizon. Then the trajectory generated by following a policy $\pi$ can be defined as $\tau=\{s_0,a_0,s_1,\cdots,a_{T-1},s_T\} \sim \Pi(\cdot;\theta)$, where $s_0$ is the initial state and
$\Pi(\tau;\theta)=p(s_0)\prod_{t=0}^{T-1}\pi(a_t| s_{t};\theta)p(s_{t+1}|s_{t},a_{t})$. The accumulated total reward can thus be written as $R=\sum_{t=0}^{T-1}\eta^t u(s_{t+1},a_{t},s_{t})=U(\tau)$, which is being viewed as a function of a random trajectory $\tau$ and follows a distribution denoted by $ F_R(\cdot;\theta)$.

\subsection{Mean-based Criterion for RL}
In the classical RL setting, the objective is to determine the optimal choice of $\theta$ that  maximizes the expected cumulative reward, i.e.,
\begin{align}
    \max\limits_{\theta\in\Theta}\mathbb{E}_{R\sim F_R(\cdot;\theta)}[R]=\max\limits_{\theta\in\Theta}\mathbb{E}_{\tau\sim\Pi(\cdot;\theta)}[U(\tau)],\label{pf.eq1}
\end{align}
where $\Theta$ is the parameter space.

An important class of techniques for solving (\ref{pf.eq1}) is the policy gradient algorithms, where the likelihood-ratio method is frequently used to derive unbiased gradient estimators that do not rely on knowledge of the transition probabilities.
Specifically, the gradient of the objective function in (\ref{pf.eq1}) can be reformulated as
\begin{align}
    \nabla_{\theta}\mathbb{E}[U(\tau)]&=\nabla_{\theta}\int_{\Omega_\tau}U(\tau)\Pi(\tau;\theta)d\tau 
    =\int_{\Omega_\tau}U(\tau)\frac{\nabla_{\theta}\Pi(\tau;\theta)}{\Pi(\tau;\theta)}\Pi(\tau;\theta)d\tau\nonumber\\
    &=\mathbb{E}\bigg[U(\tau)\sum_{t=0}^{T-1}\nabla_{\theta}\log\pi(a_t|s_t;\theta)\bigg], \label{pf.eq2}
\end{align}
 where the interchange of the gradient and integral in the second equality can be justified by the dominated convergence theorem \citep{l1992experimental}. This yields an unbiased estimator  $U(\tau)\sum_{t=0}^{T-1}\nabla_{\theta}\log\pi(a_t|s_t;\theta)$ for estimating $\nabla_{\theta}\mathbb{E}[U(\tau)]$. In addition, by noticing that for $t'<t$,
$\mathbb{E}_{\tau\sim\Pi(\cdot;\theta)}[u(s_{t'+1},a_{t'},s_{t'})\nabla_{\theta}\log\pi(a_t|s_t;\theta)]=0$, the gradient estimator can be further reduced to
$\sum_{t=0}^{T-1}\nabla_{\theta}\log\pi(a_t|s_t;\theta)\sum_{t'=t}^{T-1}\eta^{t'-t}u(s_{t'+1},a_{t'},s_{t'})$,
where $\sum_{t'=t}^{T-1}\eta^{t'-t}u(s_{t'+1},a_{t'},s_{t'})$ is the reward-to-go when starting from state $s_t$ and action $a_t$.

\subsection{Quantile-Based Criterion for RL}\label{sec33}
For a given probability level $\alpha\in(0,1)$, the $\alpha$-quantile for distribution $F_R(\cdot;\theta)$ is defined as
\begin{align*}
    q(\alpha;\theta)=\arg\inf\{r:P(R(\theta)\leq r)=F_R(r;\theta)\geq\alpha\}.
\end{align*}
We assume that $F_R(r;\theta)$ is continuously differentiable on $\mathbb{R}$, i.e., $F_R(r;\theta)\in C^1(\mathbb{R})$, so that the $\alpha$-quantile can be written as the inverse of the distribution function, i.e., $q(\alpha;\theta)=F_R^{-1}(\alpha;\theta)$.
Our goal is to maximize the $\alpha$-quantile of the distribution on a compact convex set $\Theta\subset\mathbb{R}^m$, i.e.,
\begin{align}
    \max\limits_{\theta\in\Theta}q(\alpha;\theta)=\max\limits_{\theta\in\Theta}F^{-1}_R(\alpha;\theta). \label{pf.eq3}
\end{align}

As in a typical policy gradient algorithm, we consider solving (\ref{pf.eq3}) by a stochastic gradient ascent method that makes use of the gradient information. To this end, note that by the definition of $q(\alpha;\theta)$, we have $F_R(q(\alpha;\theta);\theta)=\alpha$. Thus, an analytical expression for its gradient can be readily obtained by taking gradients at both sides of the equation \citep[see, e.g.,][]{fu2009conditional},
\begin{align}
    \nabla_{\theta} q(\alpha;\theta)=-\frac{\nabla_{\theta} F_R(r;\theta)}{f_R(r;\theta)}\bigg|_{r=q(\alpha;\theta)}. \label{pf.eq4}
\end{align}
The simplest way to use this relationship would be to construct two separate estimators for the numerator and denominator of (\ref{pf.eq4}). However, there are two major difficulties:
\begin{itemize}
\item[$\bullet$] The right-hand-side of (\ref{pf.eq4}) contains the $\alpha$-quantile itself, which is unknown and changes value as the underlying parameter vector $\theta$ changes;
\end{itemize}
\begin{itemize}
\item[$\bullet$] The density function $f_R(\cdot;\theta)$ of the cumulative reward usually does not have an analytical form. 
\end{itemize}

\section{Quantile-Based Policy Optimization}\label{Section Quantile-Based Policy Optimization}
Following our discussion in Section~\ref{sec33}, we propose a viable approach for approximating  the quantile gradient (\ref{pf.eq4}) and present a new on-policy policy gradient algorithm, QPO,
for solving (\ref{pf.eq3}). We then establish the strong convergence of the algorithm and characterize its rates of convergence.

\subsection{On-Policy RL Algorithm  for Optimizing Quantiles}
\cite{hu2022stochastic} recently propose a three-timescale SA algorithm for stochastic optimization problems with quantile objectives. However, it is difficult to apply their algorithm in our setting because the density estimation in the denominator of (4) relies on the analytical forms of the transition probability and reward function that are typically not available in an RL context. In addition, tuning and finding the best set of algorithm parameters in a three-timescale SA method could be elusive. Algorithms with less hyperparameters consume smaller amounts of trial-and-error tuning data and effort, and hence are easier and more efficient to implement
on large scale RL problems that require expensive simulation models for performance evaluation. 

The QPO algorithm we propose is a two-timescale approach that does not require estimating the density $f_R(r;\theta)$. In particular, a simple but important observation is that the density in the denominator of (\ref{pf.eq4}) is always non-negative so that the quantile gradient shares the same direction as $-\nabla_{\theta} F_R(r;\theta)\big|_{r=q(\alpha;\theta)}$. This allows us to estimate the best ascent direction in searching for the optimum by constructing an estimator for $\nabla_{\theta} F_R(r;\theta)$ and then replacing $r$ by an estimate of $q(\alpha;\theta)$.

The quantile estimates can be computed using the following recursive procedure proposed in \cite{hu2022stochastic}:
\begin{align}
    q_{k+1}=q_k + \beta_k(\alpha-\mathbf{1}\{U(\tau_{k})\leq q_{k})\}),\label{iter_q}
\end{align}
where $\beta_k$ is the step-size, $\tau_k$ is the trajectory simulated by following the current policy $\pi(\cdot|\cdot;\theta_k)$ parameterized by $\theta_k$, and $q_k$ is an estimate of $q(\alpha;\theta_k)$ at step $k$. When $\theta_k$ is fixed, it can be seen that (\ref{iter_q}) is essentially an SA iteration for solving the root-finding problem
$F_R(q;\theta_k)=\alpha$ and hence converges to $q(\alpha;\theta_k)$ under mild conditions on $\beta_k$.

On the other hand, we note that a similar likelihood ratio technique as in (\ref{pf.eq2}) can be applied to derive the gradient $\nabla_{\theta} F_R(r;\theta)$, yielding
\begin{align}
\nabla_{\theta} F_R(r;\theta)&=\nabla_{\theta} \mathbb{E}[\mathbf{1}\{R\leq r\}]=\nabla_{\theta} \mathbb{E}[\mathbf{1}\{U(\tau)\leq r\}] 
=\nabla_{\theta}\int_{\Omega_{\tau}}\mathbf{1}\{U(\tau)\leq r\}\Pi(\tau;\theta)d\tau \nonumber\\
&=\mathbb{E}[\mathbf{1}\{U(\tau)\leq r\}\nabla_{\theta}\log\Pi(\tau;\theta)] 
=\mathbb{E}\bigg[\mathbf{1}\{U(\tau)\leq r\}\sum_{t=0}^{T-1}\nabla_{\theta}\log\pi(a_t|s_t;\theta)\bigg] \nonumber\\
&\approx\frac{1}{N}\sum_{n=0}^{N-1}\mathbf{1}\{U(\tau_n)\leq r\}\sum_{t=0}^{T-1}\nabla_{\theta}\log\pi(a_t^n|s_t^n;\theta). \nonumber
\end{align}
Consequently, an unbiased estimator for $-\nabla_{\theta} F_R(r;\theta)$ is given by $$D(\tau;\theta,r)=-\mathbf{1}\{U(\tau)\leq r\} \sum_{t=0}^{T-1}\nabla_{\theta}\log\pi(a_t|s_{t};\theta).$$
The estimator, when combined with the quantile estimate $q_k$ obtained in (\ref{iter_q}), can be effectively integrated into the following gradient ascent method for solving (\ref{pf.eq3}):
\begin{equation}\label{iter_theta}
\theta_{k+1}\leftarrow \varphi\left(\theta_k+\gamma_kD(\tau_k;\theta_k,q_k)\right),
\end{equation}
where $\gamma_k$ is the gradient search step-size and $\varphi(\cdot)$ represents a projection operations that bring an iterate $\theta_{k+1}$ back to the parameter space $\Theta$ whenever it becomes infeasible. This leads to our proposed QPO algorithm whose detailed steps are presented in Algorithm~\ref{alg.qpo} below.

\begin{algorithm}[h]
    \caption{Quantile-Based Policy Optimization (QPO)}
    \label{alg.qpo}
\begin{algorithmic}[1]
    \STATE {\bfseries Input:} Policy network $\pi(\cdot|\cdot;\theta)$, quantile parameter $\alpha\in(0,1)$.
    \STATE {\bfseries Initialize:} Policy parameter $\theta_0\in\Theta$ and quantile estimate $q_0\in\mathbb{R}$.
    \FOR{$k=0,\cdots,K-1$}
    \STATE Generate one episode $\tau_k=\{s_0^k,a_0^k,s_1^k,\cdots,a_{T-1}^k,s_{T}^k\}$ following policy $\pi(\cdot|\cdot;\theta_k)$;
    \STATE $q_{k+1}\leftarrow q_k + \beta_k\big(\alpha-\mathbf{1}\{U(\tau_k)\leq q_{k}\}\big)$;
    \STATE $\theta_{k+1}\leftarrow \varphi\left(\theta_k+\gamma_kD(\tau_k;\theta_k,q_k)\right)$.
    \ENDFOR
    \STATE {\bfseries Output:} Trained policy network $\pi(\cdot|\cdot;\theta_{K})$.
\end{algorithmic}
\end{algorithm}

\subsection{Strong Convergence of QPO}\label{Subsection Strong Convergence of QPO}
Let $(\Omega,\mathcal{F},P)$ be a probability space. We define $\mathcal{F}_{k}=\sigma(\theta_0,q_0,\cdots,\theta_k,q_k)$ as the filtration generated by our algorithm for $k=0,1,\cdots$. For a given vector $x$, let $\Vert x\Vert$ be the $L^2$ norm of $x$; for a matrix $A$, let $\Vert A \Vert$ be the Frobenius norm of $A$.
Here we introduce some assumptions before the analysis.

\begin{assumption} \label{a1}
For any $\alpha\in(0,1)$, $q(\alpha;\theta)\in C^1(\Theta)$.
\end{assumption}

\begin{assumption} \label{a2}
$\nabla_{\theta}F_R(q;\theta)$ is Lipschitz continuous with respect to both $q$ and $\theta$, i.e., there exists a constant $C$ such that $\Vert\nabla_{\theta}F_R(q_1;\theta_1)-\nabla_{\theta}F_R(q_2;\theta_2)\Vert\leq C\Vert(q_1,\theta_1)-(q_2,\theta_2)\Vert$ for any $(q_i,\theta_i)\in\mathbb{R}\times\Theta$, $i=1,2$.
\end{assumption}

\begin{assumption} \label{a3}
The step-size sequences $\{\gamma_k\}$ and $\{\beta_k\}$ satisfy

(a) $\gamma_k>0$, $\sum_{k=0}^{\infty}\gamma_k=\infty$, $\sum_{k=0}^{\infty}\gamma_k^2<\infty$;
(b) $\beta_k>0$, $\sum_{k=0}^{\infty}\beta_k=\infty$, $\sum_{k=0}^{\infty}\beta_k^2<\infty$;
(c) $\gamma_k=o(\beta_k)$.
\end{assumption}

\begin{assumption} \label{a4}
The log gradient of the neural network output with respect to $\theta$ is bounded, i.e.,  for any state-action pair $(s,a)\in\mathcal{S}\times\mathcal{A}$ and parameter $\theta\in\Theta$, $\sup_{a,s,\theta}\Vert\nabla_{\theta}\log\pi(a|s;\theta)\Vert<\infty$ .
\end{assumption}

Assumption \ref{a1} requires that the objective function is smooth enough, which is commonly assumed in continuous optimization. Assumptions \ref{a2} and \ref{a3} are standard in stochastic approximation analysis.

Since $\gamma_k = o(\beta_k)$ in Assumption \ref{a3} (c), recursion (\ref{iter_q}) is updated on a faster timescale than  recursion (\ref{iter_theta}). Let
$$g_1(q,\theta)=\alpha-F_R(q;\theta),\quad g_2(q,\theta)=-\nabla_{\theta'}F_R(q;\theta')\big|_{\theta'=\theta},$$ and we expect them to track two coupled ODEs:
\begin{align}
    \dot{q}(t)=g_1(q(t),\theta(t)),\quad \dot{\theta}(t)=\tilde{\varphi}(g_2(q(t),\theta(t))), \label{ca.eq3}
\end{align}
where $\tilde{\varphi}(\cdot)$ is a projection function satisfying
\begin{align*}
  \tilde{\varphi}(g_2(q(t),\theta(t)))=g_2(q(t),\theta(t))+p(t),
\end{align*}
where $p(t)\in-C(\theta(t))$ is the vector with the smallest norm needed to keep $\theta(t)$ in $\Theta$, and $C(\theta)$ is the normal cone to $\Theta$ at $\theta$. When $\theta(t)\in\partial\Theta$, $\tilde{\varphi}(\cdot)$ projects the gradient onto $\partial\Theta$.

Intuitively, $\theta(t)$ can be viewed as static for analyzing the dynamic of the process $q(t)$. Suppose that for some constant $\bar{\theta}\in\Theta$, the unique global asymptotically stable equilibrium of the ODE
\begin{align}
    \dot{q}(t)=g_1(q(t),\bar{\theta}) \label{ca.eq4}
\end{align}
is $q(\alpha;\bar{\theta})$.
Recursion (\ref{iter_theta}) can be viewed as tracking the ODE
\begin{align}
    \dot{\theta}(t)=\tilde{\varphi}(g_2(q(\alpha;\theta(t)),\theta(t))).\label{ca.eq5}
\end{align}
If $\theta^{*}=\arg\max_{\theta\in\Theta}q(\alpha;\theta)$ is the unique global asymptotically stable equilibrium of this ODE, then the global convergence of QPO to $\theta^{*}$ can be proved; otherwise, the sequence $\{\theta_k\}$ generated by recursion (\ref{iter_theta}) converges to some limit set of the ODE, which consists of local maxima.
For simplicity, we prove the unique global asymptotically stable equilibriums for ODEs (\ref{ca.eq4}) and (\ref{ca.eq5}) under certain conditions, the details of which can be found in Appendix \ref{Appendix Proofs in "Strong Convergence of QPO"}.

\begin{lemma}\label{lem.ode1}
$q(\alpha;\bar{\theta})$ is the unique global asymptotically stable equilibrium of ODE (\ref{ca.eq4})  for all $\bar{\theta}\in\Theta$.
\end{lemma}

\begin{lemma}\label{lem.ode2}
If $q(\alpha;\theta)$ is strictly convex on $\Theta$, then $\theta^{*}$ is the unique global asymptotically stable equilibrium of ODE (\ref{ca.eq5}).
\end{lemma}

To prove that recursions (\ref{iter_q}) and (\ref{iter_theta}) track coupled ODE (\ref{ca.eq3}), we apply the convergence theorem of the two-timescale stochastic approximation as below:
\begin{theorem}\label{th.bokar}
\citep{borkar1997stochastic} Consider two coupled recursions:
\begin{align*}
    q_{k+1}&=q_k+\beta_k (g_1(q_k,\theta_k)+\epsilon_{1,k}),\\
    \theta_{k+1}&=\varphi(\theta_k+\gamma_k (g_2(q_k,\theta_k)+\epsilon_{2,k})),
\end{align*}
where $\varphi$ is a projection function, $g_1$ and $g_2$ are Lipschitz continuous, $\{\beta_k\}$ and $\{\gamma_k\}$ satisfy Assumption \ref{a3}, $\{\epsilon_{1,k}\}$ and $\{\epsilon_{2,k}\}$ are random variable sequences satisfying
\begin{align*}
    \sum_k\beta_k\epsilon_{1,k}<\infty,\ \sum_k\gamma_k\epsilon_{2,k}<\infty,\quad\ a.s.
\end{align*}
If ODE (\ref{ca.eq4}) has a unique global asymptotically stable equilibrium $q(\alpha;\bar{\theta})$ for each $\bar{\theta}\in\Theta$, then the coupled recursions converge to the unique global asymptotically stable equilibrium of the ODE $\dot{\theta}(t)=\tilde{\varphi}(g_2(q(\alpha;\theta(t)),\theta(t)))$ a.s. provided that the sequence $\{q_k\}$ is bounded.
\end{theorem}

Next we show that the sequence $\{q_k\}$ is almost surely bounded under our assumptions. Then the conditions in Theorem \ref{th.bokar} can be verified in Theorem \ref{th.main}.

\begin{lemma}\label{lem.q}
If Assumptions \ref{a1} and \ref{a3}(b) hold, then the sequence $\{q_k\}$ generated by recursion (\ref{iter_q}) is bounded w.p.1, i.e., $\sup_k|q_k|<\infty$ w.p.1.
\end{lemma}

\begin{theorem}\label{th.main}
If Assumptions \ref{a1}-\ref{a4} hold and $q(\alpha;\theta)$ is strictly convex on $\Theta$, then the sequence $\{\theta_k\}$ generated by recursions (\ref{iter_q}) and (\ref{iter_theta}) converges to the unique optimal solution $\{q(\alpha;\theta^*),\theta^*\}$ of problem (\ref{pf.eq3}) w.p.1.
\end{theorem}

\subsection{Rate of Convergence}\label{Subsection Rate of Convergence}
We first establish a central limit theorem for the two-timescale SA algorithm, the proof of which can be found in Appendix \ref{Appendix Proofs in "Rate of Convergence"}. We consider specific step-sizes of the forms $\beta_k=bk^{-\beta}$ and $\gamma_k=rk^{-\gamma}$ for $k\geq1$, where $b,r>0$ and $\frac{1}{2}<\beta<\gamma<1$ and introduce an assumption on the Hessian of the distribution function. All conclusions in this section on the convergence rates of the algorithm are based on assuming that QPO converges to $\theta^*$ that lies in the interior of $\Theta$.

\begin{assumption}\label{clt.a1}
The smallest eigenvalue of $\nabla^2_{\theta'}F_R(q(\alpha;\theta^*);\theta')\big|_{\theta'=\theta^*}$ is greater than 0.
\end{assumption}
To derive the joint weak convergence rate of the sequences $\{q_k\}$ and $\{\theta_k\}$, we apply a central limit theorem for two-timescale stochastic approximation in \cite{mokkadem2006convergence}.
Let $Q$ be the block coefficient matrix in the first order Taylor polynomial of $(g_1(q,\theta),g_2(q,\theta))^{\top}$ at $\{q(\alpha;\theta^*),\theta^*\}$, i.e.,
\begin{align*}
    \begin{pmatrix}
    g_1(q,\theta)\\g_2(q,\theta)
    \end{pmatrix}
    = \begin{pmatrix}
    Q_{11} & Q_{12} \\ Q_{21} & Q_{22}
    \end{pmatrix}
    \begin{pmatrix}
    q-q(\alpha;\theta^*) \\ \theta-\theta^*
    \end{pmatrix}
    +O\left(\left\Vert \begin{matrix}
    q-q(\alpha;\theta^*) \\ \theta-\theta^*
    \end{matrix}\right\Vert^2\right),
\end{align*}
where
\begin{align*}
    Q_{11} &= -f_R(q(\alpha;\theta^*);\theta^*),&Q_{12} &= -\nabla_{\theta'}F_R(q(\alpha;\theta^*);\theta')\big|_{\theta'=\theta^*},\\
    Q_{21} &= -\nabla_{q'}\left(\nabla_{\theta'}F_R(q';\theta')\big|_{\theta'=\theta^*}\right)\big|_{q'=q(\alpha;\theta^*)},&Q_{22} &= -\nabla^2_{\theta'}F_R(q(\alpha;\theta^*);\theta')\big|_{\theta'=\theta^*},
\end{align*}
and $Q_{12}=0$ by the first order necessary condition for optimality.

\begin{theorem} \label{clt}
If Assumptions \ref{a1}-\ref{clt.a1} hold and $\theta^*\in\Theta$, then we have
\begin{align}
    \begin{pmatrix}
    \sqrt{\beta_k^{-1}}(q_k-q(\alpha;\theta^*))\\\sqrt{\gamma_k^{-1}}(\theta_k-\theta^*)
    \end{pmatrix}\stackrel{\mathcal{D}}{\longrightarrow}\mathcal{N}
    \left(0,
    \begin{pmatrix}
    \Sigma_{q} & 0 \\ 0 & \Sigma_{\theta}
    \end{pmatrix}
    \right),\label{clt_rate}
\end{align}
where
\begin{align*}
\Sigma_q&=\int_0^{\infty} \exp \left\{Q_{11} t\right\}  \text{Var}(\mathbf{1}\{R\leq q(\alpha;\theta^*)\}) \exp \left\{Q_{11}^{\top}t\right\} d t ,\\
\Sigma_\theta&=\int_0^{\infty} \exp \left\{Q_{22} t\right\} \text{Var}(D(\tau;\theta^*, q(\alpha;\theta^*)) \exp \left\{Q_{22}^{\top} t\right\} d t.
\end{align*}
\end{theorem}

In addition to the above asymptotic normality result, we further characterize the finite-time performance of the algorithm in terms of its mean-squared errors. We make the following assumptions:

\begin{assumption} \label{a5}
Let $Q$ be a set containing all $q_k$. There exist $C_f^{-}, C_f^{+}>0$, such that $f(q;\theta)\in[ C_f^{-}, C_f^{+}]$ for $q\in Q$ and $\theta\in\Theta$.
\end{assumption}

\begin{assumption} \label{a6}
Let $H(\theta)=\nabla^2_{\theta}q(\alpha;\theta)$ and $\lambda(\theta)$ be its {\color{black}largest} eigenvalue. $\exists\  C_{\lambda}>0$, such that $\lambda(\theta){\color{black}<-C_{\lambda}}$ for all $\theta\in\Theta$.
\end{assumption}

We first derive the order of the mean squared errors for recursion (\ref{iter_q}).
\begin{theorem} \label{th.q_rate}
If Assumptions \ref{a1}-\ref{a5} hold, then the sequence $\{q_k\}$ generated by recursion (\ref{iter_q}) satisfies
\begin{align*}
    \mathbb{E}[\Vert q_k-q(\alpha;\theta_k)\Vert^2]=O(\frac{\gamma_k^2}{\beta_k^2})+O(\beta_k).
\end{align*}
\end{theorem}

Next, we establish the convergence rate of recursion (\ref{iter_theta}).

\begin{theorem}\label{th.theta_rate}
If Assumptions \ref{a1}-\ref{a6} hold, then the sequence $\{\theta_k\}$ generated by recursions (\ref{iter_q}) and (\ref{iter_theta}) satisfies
\begin{align*}
    \mathbb{E}[\Vert \theta_k-\theta^*\Vert^2] = O(\gamma_k)+O(\frac{\gamma_k^2}{\beta_k^2})+O(\beta_k).
\end{align*}
\end{theorem}

\section{Acceleration Technique}\label{Section Acceleration Technique}
In this section, we propose an acceleration technique for the vanilla QPO to improve the data utilization efficiency and prove its convergence.
We begin by introducing in Section \ref{Parameter Update with Truncated Trajectory} a variant of QPO that allows the quantile estimation and parameter update to be carried out based on truncated trajectories. This is then used in conjunction with an importance sampling technique in Section \ref{Off-Policy RL Algorithm for Optimizing Quantiles} to arrive at an off-policy procedure that reuses simulation samples and performs multiple policy updates during a single episode. A clipped surrogate objective approach is subsequently considered in Section \ref{Clipped Surrogate Objective} to further reduce the variance of the estimates in order to achieve better learning capacity and stability.

\subsection{Parameter Update with Truncated Trajectory}\label{Parameter Update with Truncated Trajectory}
For the state-of-art mean-based RL algorithms such as the one in \cite{schulman2017proximal}, the parameters can be updated multiple times during one episode, whereas the quantile-based RL algorithm (\ref{iter_theta}) only updates the parameters once in an episode, which is inefficient in utilizing the simulated data. To alleviate the issue, we use truncated subsequences rather than the entire trajectory to update policy parameters.

Denote the trajectory generated at the $k$-th simulation episode with length $l$ as  $\tau_k^l=\{s_0^k,a_0^k,s_1^k,\cdots,a_{l-1}^k,s_{l}^k\}$ and the corresponding accumulated reward as $R_k^l=\sum_{t=0}^{l-1}\eta^{t} u(s_{t+1}^k,a_{t}^k,s_{t}^k)=U(\tau_k^l)$, which follows the distribution $F_{R^l}(\cdot;\theta)$. Assume that the MDP has a finite time horizon $T$, and our objective (\ref{pf.eq3}) can be rewritten as:
\begin{align}
    \max\limits_{\theta\in\Theta}q^T(\alpha;\theta)= \max\limits_{\theta\in\Theta}F_{R^T}^{-1}(\alpha;\theta). \label{prob.aqpo}
\end{align}
We keep track of not only the quantile estimate of $F_{R^T}(\cdot;\theta)$ but also the quantile estimates $\{q_k^l\}_{l=T_0}^T$ of distributions $\{F_{R^l}(\cdot;\theta)\}_{l=T_0}^T$ with $T_0$ being a truncation parameter. In the $k$-th iteration, we randomly select an integer $l_k$ from $\{T_0,\cdots,T\}$ with equal probabilities and then run a simulation of length $l_k$ to obtain $\tau_k^{l_k}=\{s^k_0,a^k_0,s_1^k,\cdots,a_{l_k-1}^k,s_{l_k}^k\}$. The parameters are updated as follows:
\begin{align}
    q_{k+1}^{l_k} &= q_k^{l_k} + \beta_k (\alpha - \mathbf{1}\{U(\tau_k^{l_k})\leq q_k^{l_k}\}),\ q_{k+1}^{l} = q_k^{l},\ \forall\ l\neq {l_k},
    \label{iter_q_trun}\\
    \theta_{k+1}&=\varphi(\theta_k+\gamma_k D(\tau_k^{l_k};\theta_k,q_k^{l_k})\label{iter_theta_trun}),
\end{align}
where the trajectories $\tau_k^{l_k}$ are generated independently for each iteration $k$. Note that in recursion (\ref{iter_theta_trun}), we use $\tau^l$ to compute descent direction $D(\tau^l;\theta,q^l)$, which is a biased estimate of $-\nabla_{\theta'}F_{R^T}(q^T;\theta')\big|_{\theta'=\theta}$.

Next, we establish the convergence of this algorithm in a similar manner as that of Algorithm 1.
We assume $F_{R^l}(r;\theta)\in C^1(\mathbb{R})$ for all $l=T_0,\cdots,T$ and show that recursions (\ref{iter_q_trun}) and (\ref{iter_theta_trun}) track the following coupled ODEs:
\begin{equation}
   \begin{aligned}
    \dot{q}^l(t)&=g_1^l(q^l(t),\theta(t)),\ \forall\ l=T_0,\cdots,T,\\
    \dot{\theta}(t) &= \tilde{\varphi}\left(\frac{1}{T-T_0+1}\sum_{l=T_0}^T g_2^l(q^l(t),\theta(t))\right),
\end{aligned}\label{ode.aqpo}
\end{equation}
where $g_1^l(q,\theta) = \alpha - F_{R^l}(q;\theta)$ and $g_2^l(q,\theta) = -\nabla_{\theta'}F_{R^l}(q;\theta')\big|_{\theta'=\theta}$. Here we introduce some additional assumptions before the analysis.

\begin{assumption}\label{a7}
For any $\alpha\in(0,1)$ and $l=T_0,\cdots,T$, $q^l(\alpha;\theta)\in C^1(\Theta)$.
\end{assumption}

\begin{assumption} \label{a8}
For all $l=T_0,\cdots,T$, $\nabla_{\theta}F_{R^l}(q;\theta)$ is Lipschitz continuous with respect to both $q$ and $\theta$, i.e., there exists a constant $C$ such that $\Vert\nabla_{\theta}F_{R^l}(q_1;\theta_1)-\nabla_{\theta}F_{R^l}(q_2;\theta_2)\Vert\leq C\Vert(q_1,\theta_1)-(q_2,\theta_2)\Vert$ for any $(q_i,\theta_i)\in\mathbb{R}\times\Theta$, $i=1,2$.
\end{assumption}

Suppose that for $\bar{\theta}\in\Theta$, $q^l(\alpha;\bar{\theta})$ is the unique global asymptotically stable equilibrium of the ODE
\begin{align}
    \dot{q}^l(t)=g_1^l(q^l(t),\bar{\theta}),\ \forall\ l=T_0,\cdots,T. \label{qode.aqpo}
\end{align}
Then recursion (\ref{iter_theta_trun}) tracks the ODE
\begin{align}
    \dot{\theta}(t)= \tilde{\varphi}\left(\frac{1}{T-T_0+1}\sum_{l=T_0}^T g_2^l(q^l(\alpha;\theta(t)),\theta(t))\right).\label{thetaode.aqpo}
\end{align}
Let $\bar{q}(\alpha;\theta)=\sum_{l=T_0}^T q^l(\alpha;\theta)$ and assume that $\theta^{**}=\arg\max_{\theta\in\Theta} \bar{q}(\alpha;\theta)$ is the unique global asymptotically stable equilibrium of this ODE, so that we can establish the global convergence of the coupled recursions to $\theta^{**}$.
Note that if we fix $\theta(t)$ to a constant $\bar{\theta}$, then $\{q^l(t)\}_{l=T_0}^T$ is decoupled, and we immediately have the following lemmas:
\begin{lemma}\label{lem.ode3} For all $l=T_0,\cdots,T$ and $\bar{\theta}\in\Theta$,
$q^l(\alpha;\bar{\theta})$ is the unique global asymptotically stable equilibrium of ODE (\ref{qode.aqpo}). And if $\bar{q}(\alpha;\theta)$ is strictly convex on $\Theta$, then $\theta^{**}$ is the unique global asymptotically stable equilibrium of ODE (\ref{thetaode.aqpo}).
\end{lemma}

To apply the convergence theorem for two-timescale SAs, we next show in Lemmas \ref{lem.aq} and \ref{lem.aqpo} below that the sequence $\{q_k^l\}_{l=T_0}^T$ is almost surely bounded and the simulation error accumulated over the iterations remains bounded.
\begin{lemma}\label{lem.aq}
If Assumptions \ref{a3}(b) and \ref{a7} hold, then the sequence $\{q_k^l\}_{l=T_0}^T$ generated by recursion (\ref{iter_q_trun}) is bounded w.p.1, i.e., $\sup_k|q_k^l|<\infty$ w.p.1, $\forall\ l = T_0,\cdots T$.
\end{lemma}

Denote $M_k'' = \sum_{i=0}^k \gamma_i\delta_i''$, where $\delta_i'' = D(\tau_i^{l_i};\theta_i,q_i^{l_i}) +\frac{1}{T-T_0+1}\sum_{l=T_0}^T \nabla_{\theta'}F_{R^l}(q_i^l;\theta')\big|_{\theta'=\theta_i}$ and $l_i$ is randomly chosen from $\{T_0,\cdots,T\}$ with equal probabilities.
\begin{lemma}\label{lem.aqpo}
If Assumptions \ref{a3}(a), \ref{a4}, \ref{a7} and \ref{a8} hold, then $\{M_k''\}$ is bounded w.p.1.
\end{lemma}

\begin{theorem}\label{main.aqpo}
If Assumptions \ref{a3}, \ref{a4}, \ref{a7} and \ref{a8} hold and $\sum_{l=T_0}^T q^l(\alpha;\theta)$ is strictly convex on $\Theta$, then the the sequences $\{q_k^l\}_{l=T_0}^T$ and $\{\theta_k\}$ generated by recursions (\ref{iter_q_trun}) and (\ref{iter_theta_trun}) converge to the unique optimal solution $\{\{q^l(\alpha;\theta^{**})\}_{l=T_0}^T,\theta^{**}\}$ w.p.1.
\end{theorem}

Finally, we show that the error introduced by truncation is asymptotically negligible as $T\rightarrow\infty$ under different reward settings. We consider uniformly bounded and Gaussian rewards, as well as the more general sub-Gaussian case. In each case, we provide an explicit bound on the error caused by trajectory truncation. Proofs of all three cases can be found in Appendix \ref{Appendix Acceleration Technique}.
\begin{theorem}\label{q_error_1}
If the reward function is bounded, i.e., there exists a constant $C_r>0$, for any $(s',a,s)\in\mathcal{S}\times\mathcal{A}\times\mathcal{S}, |u(s',a,s)|\leq C_r$, then $|q^T(\alpha;\theta^*)-q^T(\alpha;\theta^{**})|\leq \frac{2}{T-T_0+1}\cdot\frac{\eta^{T_0}}{(1-\eta)^2}C_r$.
\end{theorem}
The reward setting in Theorem \ref{q_error_1} is the most common situation in classical RL problems. For video games and robotic control, rewards are often chosen to be finite, such as the position of an agent, or the time it takes to finish a task.

\begin{theorem}\label{q_error_2}
If the reward is normally distributed, i.e., there exist constants $C_{\mu},C_{\sigma}>0$, for any $(s',a,s)\in\mathcal{S}\times\mathcal{A}\times\mathcal{S}$, $ u(s',a,s)\sim\mathcal{N}(\mu(s',a,s),\sigma^2(s',a,s))$, where $|\mu(s',a,s)|<C_{\mu}$, $\sigma(s',a,s)<C_{\sigma}$, then $|q^T(\alpha;\theta^*)-q^T(\alpha;\theta^{**})|\leq \frac{2}{T-T_0+1}\cdot\frac{\eta^{T_0}}{(1-\eta)^2}C_r'$, where $C_r'=C_{\sigma}|q_{\mathcal{N}(0,1)}(\alpha)|+C_{\mu}$ and $q_{\mathcal{N}(0,1)}(\alpha)$ is the $\alpha$-quantile of the standard normal distribution.
\end{theorem}
In financial portfolio management, the rewards (returns) of assets are often assumed to be Gaussian.

\begin{theorem}\label{q_error_3}
If the reward is a sub-Gaussian random variable, i.e., there exist constants $c,C_{\mu}>0$, for any $(s',a,s)\in\mathcal{S}\times\mathcal{A}\times\mathcal{S}$, $P(|u(s',a,s)-\mu(s',a,s)|\geq \xi)\leq 2\exp(-c\xi^2)$, where the mean of $u(s',a,s)$ is finite, i.e., $|\mu(s',a,s)|<C_{\mu}$, then $|q^T(\alpha;\theta^*)-q^T(\alpha;\theta^{**})| = O\left(\frac{\sqrt{\eta^{T_0}}}{T-T_0+1}\right)$.
\end{theorem}
Sub-Gaussian family of distributions includes Gaussian, Bernoulli, bounded distributions, and Beta or Dirichlet distributions under certain conditions \citep{marchal2017sub}. Moreover, the sub-Gaussianity is preserved by linear operations, which enables us to construct more complex distributions belonging to the family.

\subsection{Off-Policy RL Algorithm  for Optimizing Quantiles}\label{Off-Policy RL Algorithm for Optimizing Quantiles}

\begin{figure}[t]
	\centering
	\includegraphics[scale=0.48]{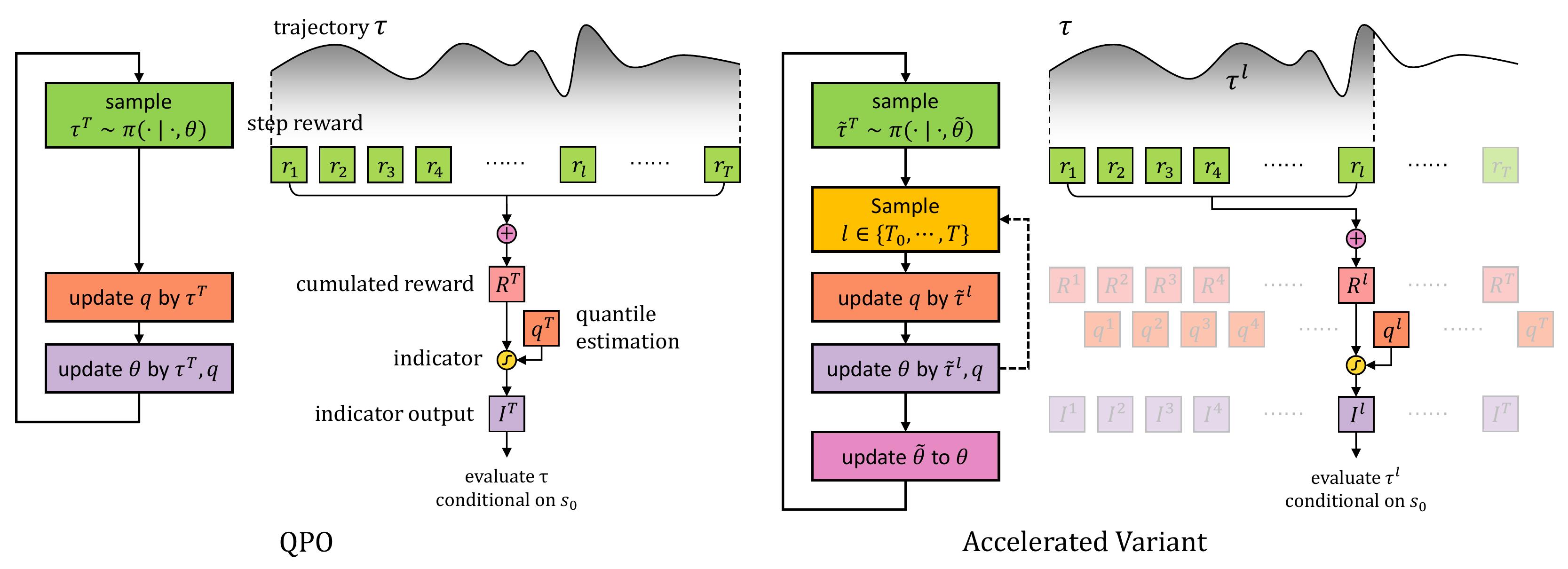}
	\caption{Comparison of QPO and its accelerated variant in terms of algorithm structure and data flow.}
	\label{fig:alg_comparison}
\end{figure}

By applying the importance sampling method, we can use a fixed policy $\pi(\cdot|\cdot,\hat{\theta})$ to interact with the environment, and then recursions (\ref{iter_q_trun}) and (\ref{iter_theta_trun}) can be transformed into
\begin{align}
q_{k+1}^{l_k} &= q_k^{l_k} + \beta_k (\alpha - \rho(\hat{\tau}^{l_k}_k;\theta_{k+1},\hat{\theta}) \mathbf{1}\{U(\hat{\tau}^{l_k}_k)\leq q_k^{l_k}\}),\ q_{k+1}^{l} = q_k^{l},\ \forall\ l\neq {l_k} \label{iter_q_is}\\
\theta_{k+1}&=\varphi(\theta_k+\gamma_k \rho(\hat{\tau}^{l_k}_k;\theta_{k+1},\hat{\theta}) D(\hat{\tau}^{l_k}_k;\theta_k,q_k^{l_k})) \label{iter_theta_is},
\end{align}
where trajectories $\{\hat{\tau}^{l_k}_k\}$ are generated independently following $\pi(\cdot|\cdot,\hat{\theta})$ and the importance sampling ratio is defined as
\begin{align*}
     \rho(\hat{\tau};\theta,\hat{\theta})=\frac{\Pi(\hat{\tau};\theta)}{\Pi(\hat{\tau};\hat{\theta})}=\prod_{t}\rho(\hat{a}_t,\hat{s}_t;\theta,\hat{\theta}),\quad  &\rho(\hat{a}_t,\hat{s}_t;\theta,\hat{\theta})=\frac{\pi(\hat{a}_t|\hat{s}_t;\theta)}{\pi(\hat{a}_t|\hat{s}_t;\hat{\theta})}.
\end{align*}
Recursions (\ref{iter_q_is}) and (\ref{iter_theta_is}) are equivalent to recursions (\ref{iter_q_trun}) and (\ref{iter_theta_trun}) in a probability sense \citep{glynn1989importance}. And the convergence results in Section 3.1 hold for recursions (\ref{iter_q_is}) and (\ref{iter_theta_is}). The differences in logical structure and data flow between vanilla QPO and the accelerated variant are shown in Figure.\ref{fig:alg_comparison}.

We consider modified versions of (\ref{iter_q_is}) and (\ref{iter_theta_is}) that improve data utilization efficiency by allowing multiple updates on $\theta$ in one episode for quantile-based RL. Let $\theta^{T-T_0}_k$ be the policy parameter obtained just prior to the $k$-th episode.
In the $k$-th episode, the latest policy $\pi(\cdot|\cdot,\theta_k^{T-T_0})$ is used to interact with the environment and generate a trajectory ${\tau}_k^T$. Then the policy parameter is updated multiple times based on subsequences of ${\tau}_k^T$ generated in one simulation episode.
For $j=0,\cdots,T-T_0$, we randomly select ${l_j}$ from $\{T_0,\cdots,T\}$ with equal probabilities in the $k$-th iteration. Then we use the subsequence ${\tau}_k^{l_j}=\{{s}_0^k,{a}_0^k,{s}_1^k,\cdots,{a}_{{l_j}-1}^k,{s}_{{l_j}}^k\}$ of trajectory ${\tau}_k^T$ to perform the following recursions:
\begin{align}
    q_{k+1}^{l_j} &= q_k^{l_j} + \beta_k\big(\alpha-\rho({\tau}^{l_j}_k;\theta_{k+1}^{j},\theta_k^{T-T_0})\mathbf{1}\{U({\tau}_k^{l_j})\leq q_{k}^{l_j}\}\big),
    \label{iter_q_acc}\\
    \theta_{k+1}^{j+1} &= \varphi\left(\theta_{k+1}^{j}+\gamma_k \rho({\tau}^{l_j}_k;\theta_{k+1}^{j},\theta_k^{T-T_0}) D({\tau}_k^{l_j};\theta_{k+1}^{j},q_k^{l_j})\right),\label{iter_theta_acc}
\end{align}
where $\theta_{k+1}^0 =\theta_k^{T-T_0}$. Since subsequences of $\tau_k^T$ in recursions (\ref{iter_q_acc}) and (\ref{iter_theta_acc}) share common random variables, the stochastic gradient estimates at different iterations of recursions (\ref{iter_q_acc}) and (\ref{iter_theta_acc}) become correlated. The convergence result of SA with correlated noise are rather technical and can be referred to Chapter 6 in \cite{kushner2003stochastic}.

\subsection{Clipped Surrogate Objective}\label{Clipped Surrogate Objective}
In Section \ref{Off-Policy RL Algorithm for Optimizing Quantiles}, the on-policy QPO is transformed to an off-policy algorithm by employing the importance sampling method.
However, the importance ratio term $\rho({\tau}^{l_j}_k;\theta_{k+1}^{j},\theta_k^{T-T_0})$ may inflate the variance significantly.
This issue has been considered in mean-based RL algorithms. In TRPO \citep{schulman2015trust}, an optimization problem with a surrogate objective of (\ref{pf.eq1}) is proposed by applying the importance sampling method, i.e.,
\begin{align*}
    \max_{\theta\in\Theta} \mathbb{E}_{t,\hat{\tau}\sim\Pi(\cdot;\hat\theta)}[\rho(\hat{a}_t,\hat{s}_t;\theta,\hat{\theta})\hat{A}_t],\quad
    \text{s.t.}\ \text{KL}(\Pi(\cdot;\hat\theta)\Vert\Pi(\cdot;\theta))\leq\delta,
\end{align*}
where $A_t$ is the advantage function of the $t$-th decision calculated by some variant of the “reward-to-go" and the KL divergence constraint is used to enforce the distribution under $\theta$ to stay close to $\Pi(\cdot,\hat\theta)$ so that the variance of the importance ratio does not become excessively large. To simplify the computation, \cite{schulman2015trust} use a penalty rather than a hard constraint, whereas PPO introduced by \cite{schulman2017proximal} is based on an optimization problem with a clipped surrogate objective that is much easier to handle, i.e.,
\begin{align*}
    \max_{\theta\in\Theta} \mathbb{E}_{t,\hat{\tau}\sim\Pi(\cdot;\hat\theta)}[\min\{\rho(\hat{a}_t,\hat{s}_t;\theta,\hat{\theta})\hat{A}_t,\text{clip}(\rho(\hat{a}_t,\hat{s}_t;\theta,\hat{\theta}),1-\varepsilon,1+\varepsilon)\hat{A}_t\}],
\end{align*}
where the clip function is denoted as $\text{clip}(x,x^{-},x^{+})$, and $x^{-}$ and $x^{+}$ are the lower and upper truncation bounds. 

Note that given $l_j$ and $q_k^{l_j}$, recursion (\ref{iter_theta_acc}) optimizes the surrogate objective
\begin{align}
    \mathbb{E}_{\tau_k^{T}\sim\Pi(\cdot;\theta_k^{T-T_0})}[-\mathbf{1}\{U(\tau_k^{l_j})\leq q_k^{l_j}\}\rho(\tau_k^{l_j};\theta,\theta_k^{T-T_0})].\label{surrogate}
\end{align}
Therefore, to constrain the difference between $\theta$ and $\tilde{\theta}$, we adapt the technique used in PPO to our quantile-based algorithm. Specifically, we introduce a clip operation to the ratio term in surrogate problem (\ref{surrogate}).
To further reduce the variance, we employ a baseline network $B(s_0^k, l_j|w)$, that does not depend on $\theta$ and is updated by minimizing the mean-squared error (MSE) of the difference from $-\mathbf{1}\{u(\tau_k^{l_j})\leq q_{k}^{l_j}\}$.
Then, the clipped surrogate objective can be written as
\begin{equation}
\begin{aligned}
    {\mathbb{E}}_{\tau_k^{T}\sim\Pi(\cdot;\theta_k^{T-T_0})}\bigg[\min\{&\rho(\tau_k^{l_j};\theta_{k+1}^{j},\theta_k^{T-T_0})(-\mathbf{1}\{U(\tau_k^{l_j})\leq q_k^{l_j}\}-B(s_0^k, l_j|w)), \\
    &\text{clip}(\rho(\tau_k^{l_j};\theta_{k+1}^{j},\theta_k^{T-T_0}),1-\varepsilon,1+\varepsilon)(-\mathbf{1}\{U(\tau_k^{l_j})\leq q_k^{l_j}\}-B(s_0^k, l_j|w))\}\bigg]. \label{surrogate_clip}
\end{aligned}
\end{equation}

By applying the technique developed in Section \ref{Off-Policy RL Algorithm for Optimizing Quantiles} to optimization problem (\ref{surrogate_clip}), we propose a quantile-based off-policy deep RL algorithm, named QPPO. The pseudo code of QPPO is presented in Algorithm \ref{alg.qppo}.

\begin{algorithm}[h]
   \caption{Quantile-Based Proximal Policy Optimization (QPPO)}
   \label{alg.qppo}
\begin{algorithmic}[1]
    \STATE {\bfseries Input:} Policy network $\pi(\cdot|\cdot;\theta)$,  quantile parameter $\alpha\in(0,1)$.
    \STATE {\bfseries Initialize:} Policy parameter $\theta_0^{T-T_0}$, and quantile estimators $\{q_0^l\}_{l=T_0}^T\subset\mathbb{R}$.
    \FOR{$k=0,\cdots,K-1$}
    \STATE Generate one episode ${\tau}_k^T=\{s_0^k,a_0^k,s_1^k,\cdots,a_{T-1}^k,s_{T}^k\}$ following policy $\pi(\cdot|\cdot;\theta_k^{T-T_0})$;
    \STATE Randomly shuffle the list $\{T_0,\cdots,T\}$ to generate $\{l_j\}_{j=0}^{T-T_0}$;
    \STATE $\theta_{k+1}^0 \leftarrow \theta_k^{T-T_0}$.
    \FOR{$j=0,\cdots,T-T_0$}
        \STATE $q_{k+1}^{l_j} \leftarrow q_k^{l_j} + \beta_k\big(\alpha-\rho({\tau}^{l_j}_k;\theta_{k+1}^{j},\theta_k^{T-T_0})\mathbf{1}\{U({\tau}_k^{l_j})\leq q_{k}^{l_j}\}\big)$;
        \STATE
        $\theta_{k+1}^{j+1}\leftarrow \max_{\theta\in\Theta}\ \min\{\rho(\tau_k^{l_j};\theta_{k+1}^{j},\theta_k^{T-T_0})(-\mathbf{1}\{U(\tau_k^{l_j})\leq q_k^{l_j}\}-B(s_0^k, l_j|w)),$ \\
        $\quad\quad\quad\quad\quad\quad\quad\quad\quad \text{clip}(\rho(\tau_k^{l_j};\theta_{k+1}^{j},\theta_k^{T-T_0}),1-\varepsilon,1+\varepsilon)(-\mathbf{1}\{U(\tau_k^{l_j})\leq q_k^{l_j}\}-B(s_0^k, l_j|w))\}$.
    \ENDFOR
    \ENDFOR
    \STATE {\bfseries Output:} Trained policy network $\pi(\cdot|\cdot;\theta_{K-1}^{T-T_0})$.
\end{algorithmic}
\end{algorithm}

\section{Numerical Experiments}\label{Section Numerical Experiments}

In this section, we conduct simulation experiments on different RL tasks to demonstrate the effectiveness of proposed algorithms. 
We first compare our QPO and QPPO with baseline RL algorithms REINFORCE and PPO, as well as the SPSA-based algorithm, in a simple example for illustration.
In the more complicated financial investment and inventory management examples, we focus on comparing QPPO with one of the state-of-art mean-based algorithms, PPO, to demonstrate the behavioral patterns of agents under different criteria and the high learning efficiency of QPPO.
Exogenous parameters of simulation environments and hyperparameters of algorithms used in all three examples can be found in Appendix \ref{Appendix Numerical Experiments}.

The selection of $\{\beta_k\}$ affects the convergence speed of the algorithms, since the quantile estimation is the reference for the agent to evaluate the action. To simplify parameter tuning, we relax the divergence conditions in Assumption \ref{a3} and let $\{\gamma_k\}$ and $\{\beta_k\}$ decay exponentially at rates $\eta$ and $\frac{1+\eta}{2}$, respectively. Then we 
employ an Adam optimizer for the quantile estimation recursion and use a small number of simulation episodes to provide an initial point for the quantile estimation.

\subsection{Toy Example: Zero Mean}\label{Subsection Toy Example: Zero Mean}

In a simulation environment with $T$ time steps and $N$ alternative choices, the reward of an agent at the $t$-th step is given by $r_t \sim \text{Uniform}(-s_t^{a_t},s_t^{a_t})$,
where the state $s_t=(s_t^1,\cdots,s_t^N)\in\mathbb{R}^N$ is a vector obtained by randomly reshuffling the entries of $s_{t-1}$, and $a_t\in\{1,\cdots,N\}$ is the action chosen by the agent. The agent observes the current state and chooses the coordinate index of the state vector which determines the support of the uniform distribution of the reward. It is obvious that the cumulative reward follows a distribution with zero mean.
Therefore, optimizing the expected cumulative reward is ineffective, whereas the quantile performance can be optimized by choosing $a_t^*=\arg\min_{i\in\{1,\cdots,N\}} s_t^i$.

\begin{figure}[!htb]
    \centering
    \subfigure[0.25-quantile learning curves]{
        \begin{minipage}[t]{0.4\linewidth}
            \centering
            \includegraphics[width=\linewidth]{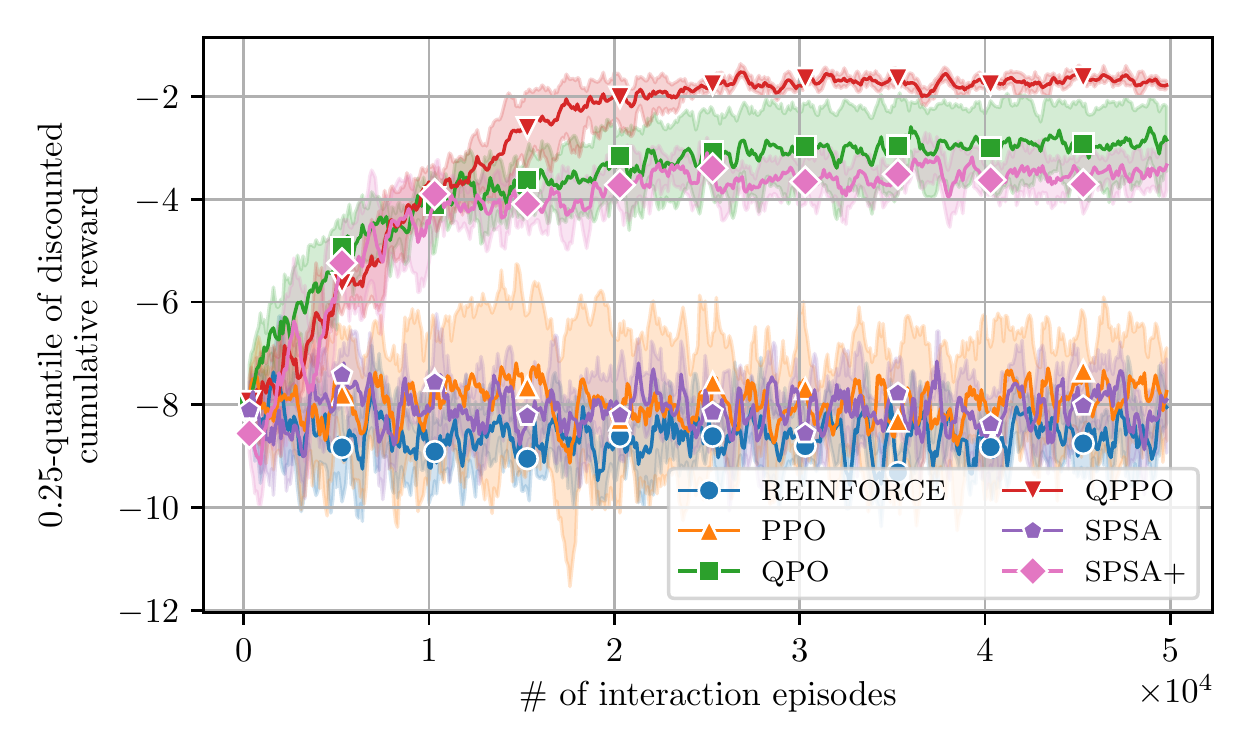}\\
        \end{minipage}
    }
    \subfigure[Accuracy learning curves]{
        \begin{minipage}[t]{0.4\linewidth}
            \centering
            \includegraphics[width=\linewidth]{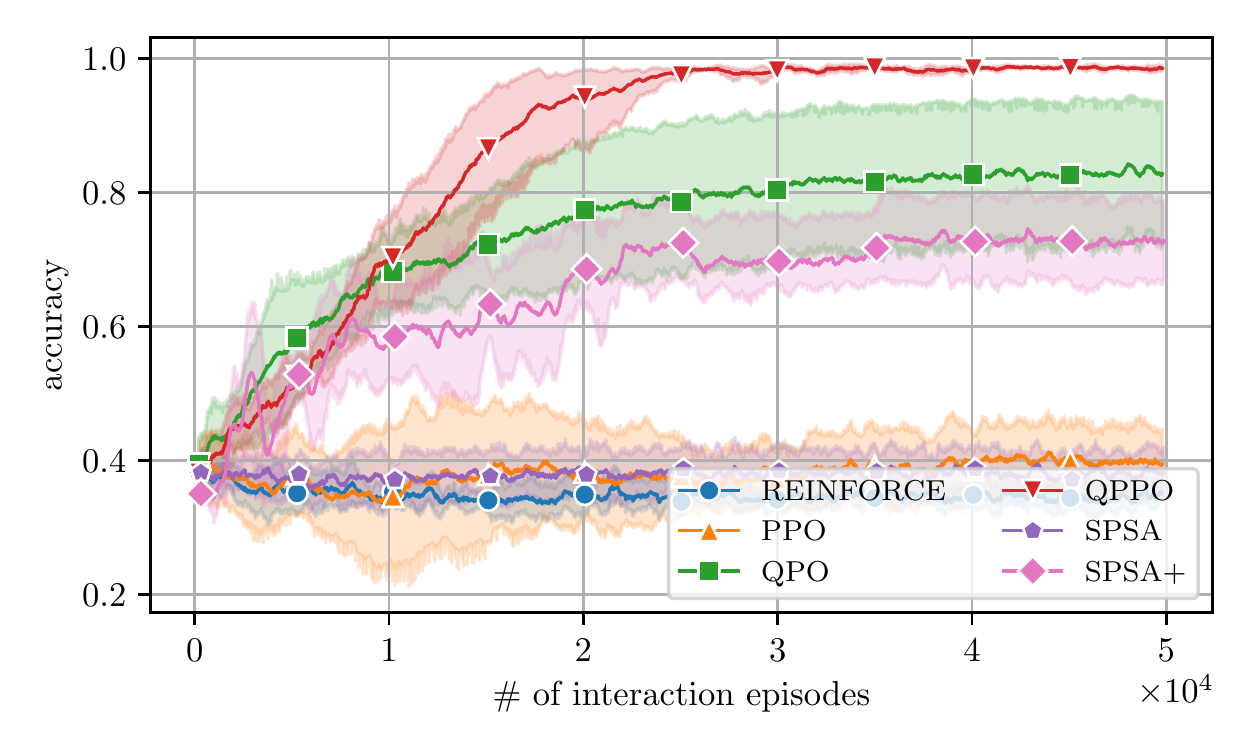}\\
        \end{minipage}
    }
    \caption{Learning curves for quantile ($\alpha=0.25$) and accuracy of REINFORCE, PPO, QPO, QPPO and SPSA in simple Zero Mean example by 5 independent experiments.}
    \label{fig: learning_curve_easy}
\end{figure}

We first test REINFORCE, PPO, QPO, QPPO and SPSA on a simple setting in which the state vector takes values in $\{1,4,9\}$.
The policy and baseline are represented by multi-layer perceptrons. The agent makes decisions in $20$ time steps.
The learning curves are presented in Figure \ref{fig: learning_curve_easy}. The quantile performance is estimated by rewards in the past $100$ episodes. The accuracy curves report the probabilities of the agent selecting the minimum alternative. The shaded areas represent the $95\%$ confidence intervals, and the curves are appropriately smoothed for better visibility. 
SPSA uses a batch of 10 samples for empirical estimation of quantile. All algorithms have the same learning rate for efficiency comparison. Due to the low efficiency of SPSA, we also use a learning rate 50 times larger than that of other algorithms in SPSA$+$ to confirm the validity of SPSA in this simple example.
We can see that QPO and QPPO significantly improve the agent’s $0.25$-quantile performance, while REINFORCE and PPO do not learn anything. The quantile performance of SPSA$+$ improves slowly, but the learning curve of SPSA is almost flat. In independently replicated macro experiments, QPPO outperforms others in terms of both stability and efficiency due to the improvement in data utilization.

\begin{figure}[!htb]
    \centering
    \subfigure[0.25-quantile learning curves]{
        \begin{minipage}[t]{0.4\linewidth}
            \centering
            \includegraphics[width=\linewidth]{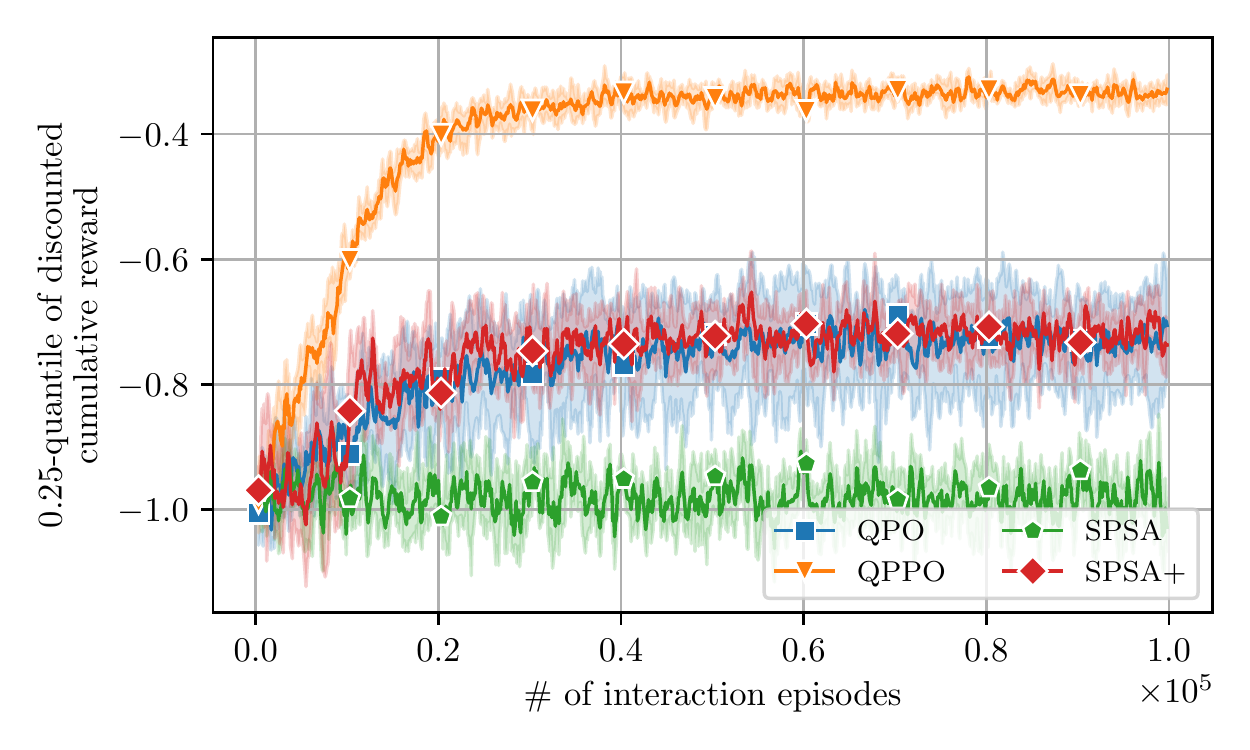}\\
        \end{minipage}
    }
    \subfigure[Accuracy learning curves]{
        \begin{minipage}[t]{0.4\linewidth}
            \centering
            \includegraphics[width=\linewidth]{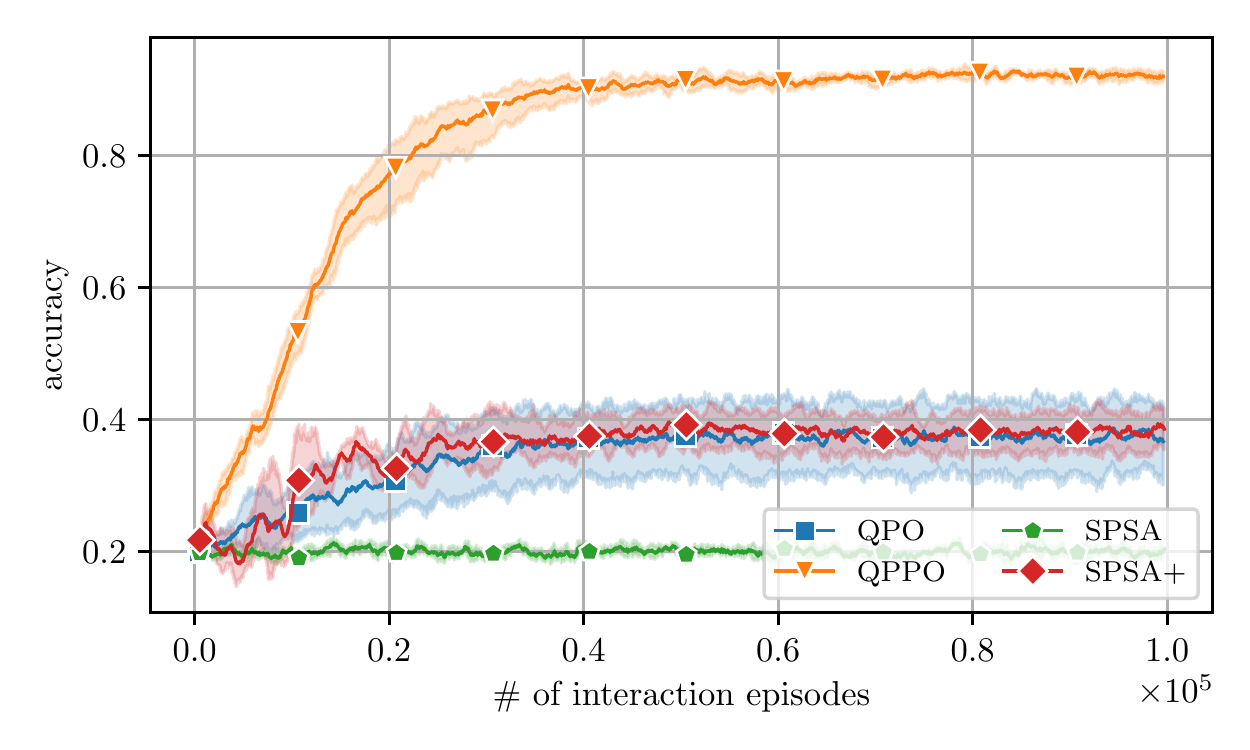}\\
        \end{minipage}
    }
    \caption{Learning curves for quantile ($\alpha=0.25$) and accuracy of QPO, QPPO and SPSA in hard Zero Mean example by 5 independent experiments.}
    \label{fig: learning_curve_hard}
\end{figure}

Next, we further test the performance of quantile-based algorithms in a harder setting where the state vector takes values in $\{0.1,0.2,0.3,0.4,0.5\}$. We increase the size of the neural networks. The results are shown in Figure \ref{fig: learning_curve_hard}, with the same plotting settings as in Figure \ref{fig: learning_curve_easy}.
SPSA$+$ represents SPSA with a learning rate 100 times larger than others. The advantage of QPPO over QPO and SPSA is even more obvious in the hard scenario. The results also indicate that the ability of SPSA-based algorithm is limited as the policy and environment become more complex.
Due to the low data efficiency of QPO and SPSA, and the difficulty in tuning SPSA, we will no longer test these two algorithms in subsequent more complicated experiments. 

\subsection{Financial Investment}\label{Financial Investment}
We consider a portfolio management problem with $N$ simulated prices and $T$ time steps, where high returns are associated with high risks. The price vector $p_t=(p_t^1,\cdots,p_t^N)$ of the alternative assets follows a multivariate geometric Brownian motion:
\begin{align}
    dp_t = \mu p_t dt + \Sigma^{\frac{1}{2}} dW_t, \label{eq:price}
\end{align}
where $\mu$ and $\Sigma^{\frac{1}{2}}$ are the drift and volatility, and $W_t$ is a Wiener process, which is widely used in finance to model stock prices, e.g., in the Black–Scholes model \citep{black1973pricing}. We generate the price paths by Monte Carlo simulation, following the Euler-Maruyama discretization \citep{kloeden1992stochastic} of the stochastic process (\ref{eq:price}), i.e., for $t=0,\cdots,T-1$, which leads to
\begin{align}
   \Delta p_t = p_{t+1} - p_t = (\mu\Delta t + \Sigma^{\frac{1}{2}} \sqrt{\Delta t} \varepsilon_t) \odot p_t,\quad \varepsilon_t\sim\mathcal{N}(0,I_N), \label{eq:price_sim}
\end{align}
where $\Delta t$ is the time interval for adjusting positions, and $\odot$ denotes the Hadamard product.

The agent initially holds a portfolio with a vector of random positions on $N$ assets $w_0 = (w_0^1,\cdots,w_0^N)$ and a fixed
total value $v_0$ and can control the proportion $a_t=(a_t^1,\cdots,a_t^N)$ of the value $v_{t}$ invested
into each asset. In addition, the market has friction, i.e., there is a transaction fee of proportion $f$ every time the agent buys. Therefore, for $t=0,\cdots,T-1$, the new position of the $i$-th asset is given by
\begin{align*}
    w_{t+1}^i = w_t^i + (1-f) \left[ \frac{a_{t}^i v_t^i}{p_t^i}- w_t^i \right]^{+} - \left[ \frac{a_{t}^i v_t^i}{p_t^i}- w_t^i \right]^{-},
\end{align*}
where $x^+$ and $x^-$ represent $\max(x,0)$ and $\max(-x,0)$, respectively.
At every time step $t$, the agent first makes
investment decisions $a_t$, and calculates the total value of the
portfolio under current prices $p_{t}$ and that under new
prices $p_{t+1}$ simulated by recursion (\ref{eq:price_sim}). The agent is rewarded by the differences of two total
values, i.e., $r_t = v_{t+1}-v_{t}=\sum_{i=1}^N p_{t+1}^i w_{t+1}^i - \sum_{i=1}^N p_{t}^i w_{t}^i$. 
The observation state contains the current vector of asset positions $w_t$, prices $p_t$ and some statistics of profit margin $m_t$, where $m_t^i = \Delta p_t^i/p_t^i$, over a certain window width.

\begin{figure}[!htb]
    \centering
    \subfigure[0.1-quantile learning curves]{
        \begin{minipage}[t]{0.25\linewidth}
            \centering
            \includegraphics[width=\linewidth]{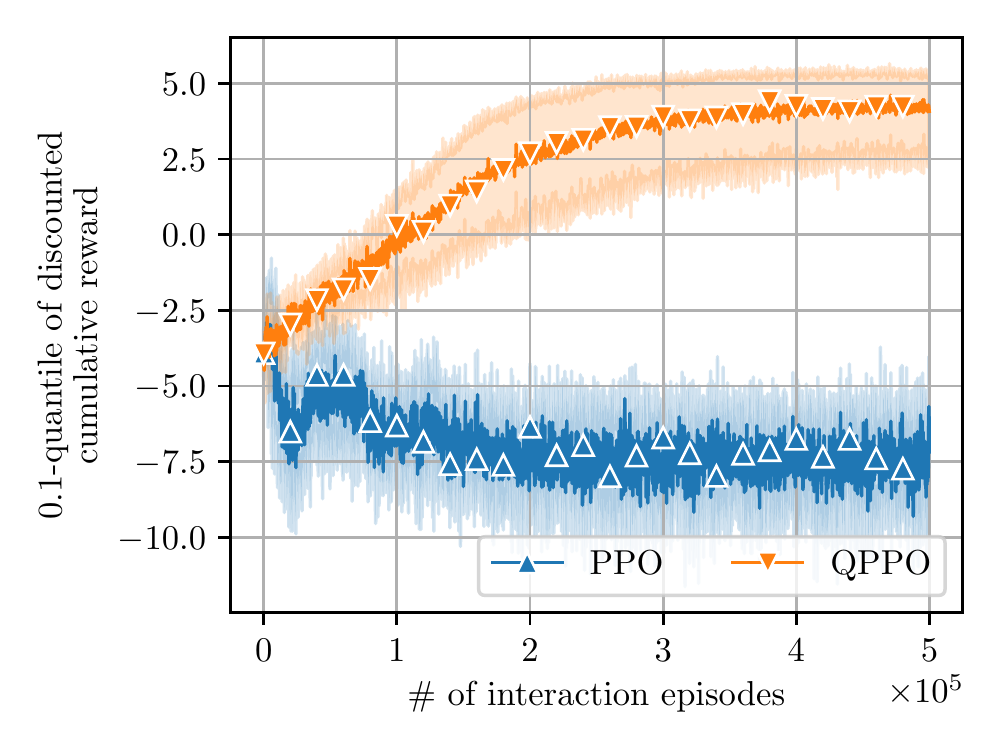}\\
        \end{minipage}
    }
    \subfigure[Average learning curves]{
        \begin{minipage}[t]{0.25\linewidth}
            \centering
            \includegraphics[width=\linewidth]{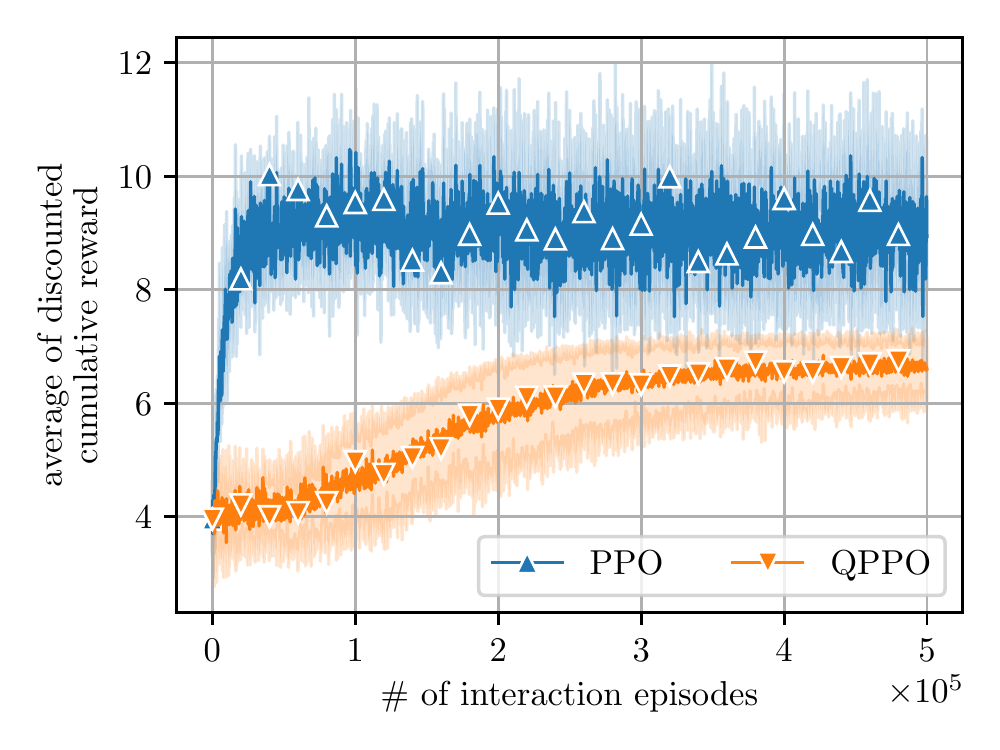}\\
        \end{minipage}
    }
    \subfigure[KDE plots]{
        \begin{minipage}[t]{0.25\linewidth}
            \centering
            \includegraphics[width=\linewidth]{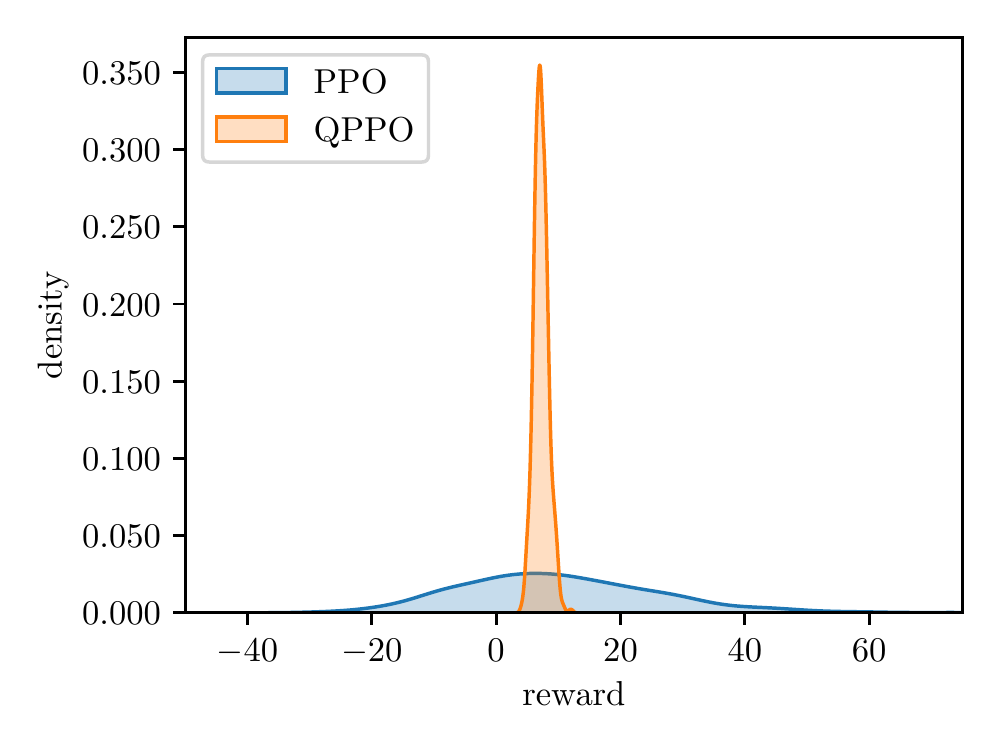}\\
        \end{minipage}
    }
    \caption{Comparison of PPO and QPPO in perfectly hedgeable Portfolio Management example.}
    \label{fig: learning_curve_kde_3}
\end{figure}

\begin{figure}[!htb]
    \centering
    \subfigure[0.1-quantile learning curves]{
        \begin{minipage}[t]{0.25\linewidth}
            \centering
            \includegraphics[width=\linewidth]{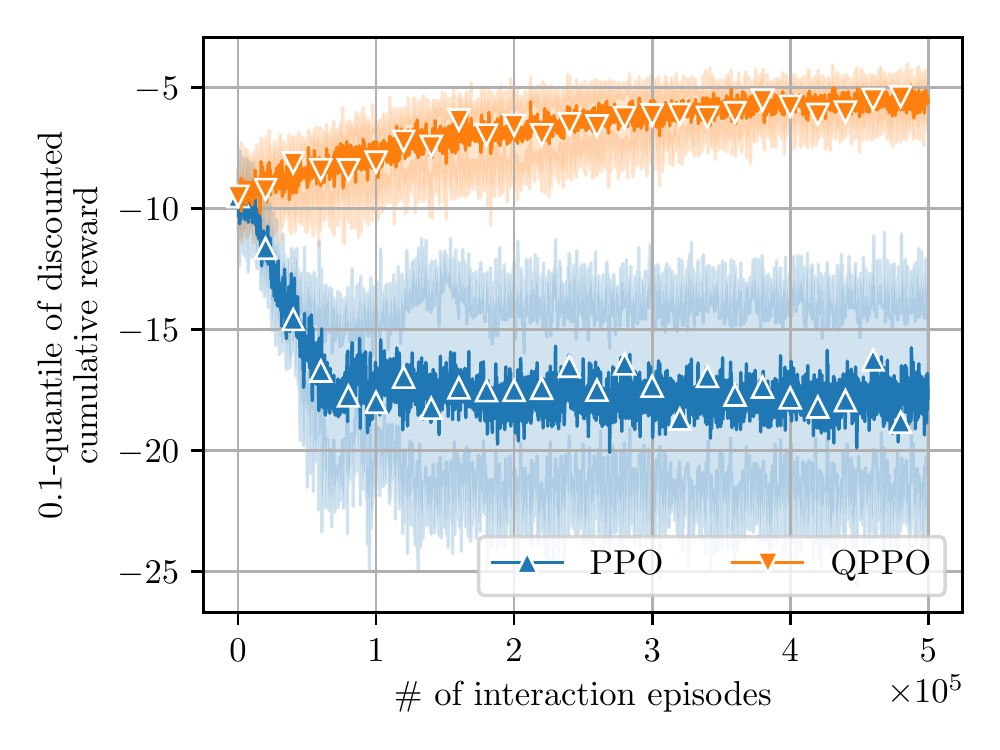}\\
        \end{minipage}
    }
    \subfigure[Average learning curves]{
        \begin{minipage}[t]{0.25\linewidth}
            \centering
            \includegraphics[width=\linewidth]{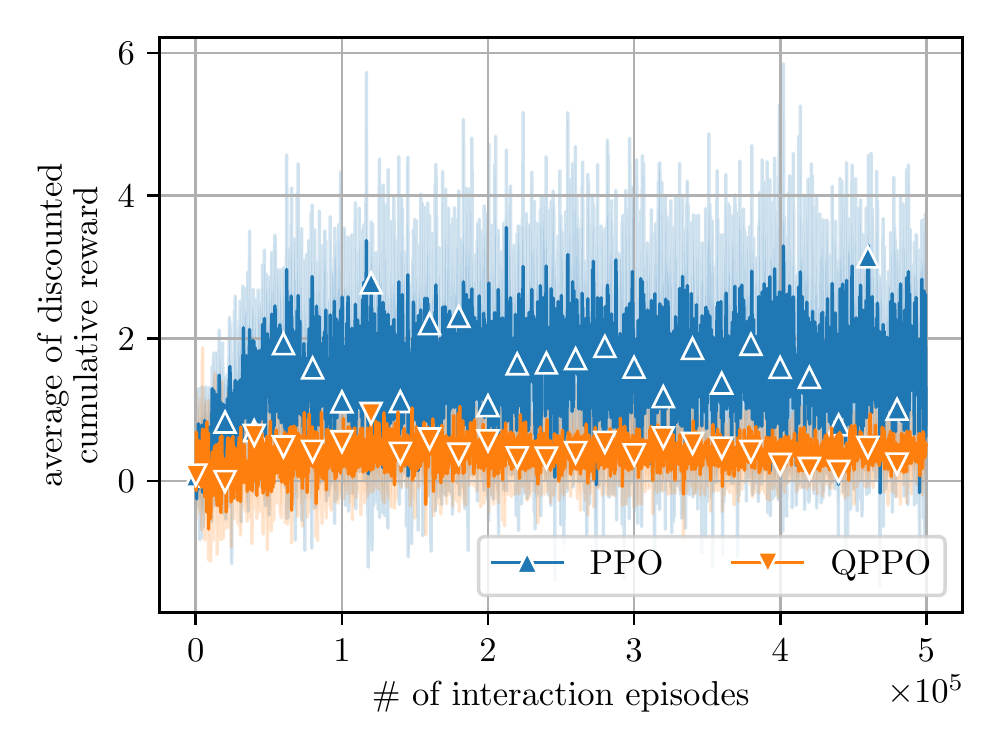}\\
        \end{minipage}
    }
    \subfigure[KDE plots]{
        \begin{minipage}[t]{0.25\linewidth}
            \centering
            \includegraphics[width=\linewidth]{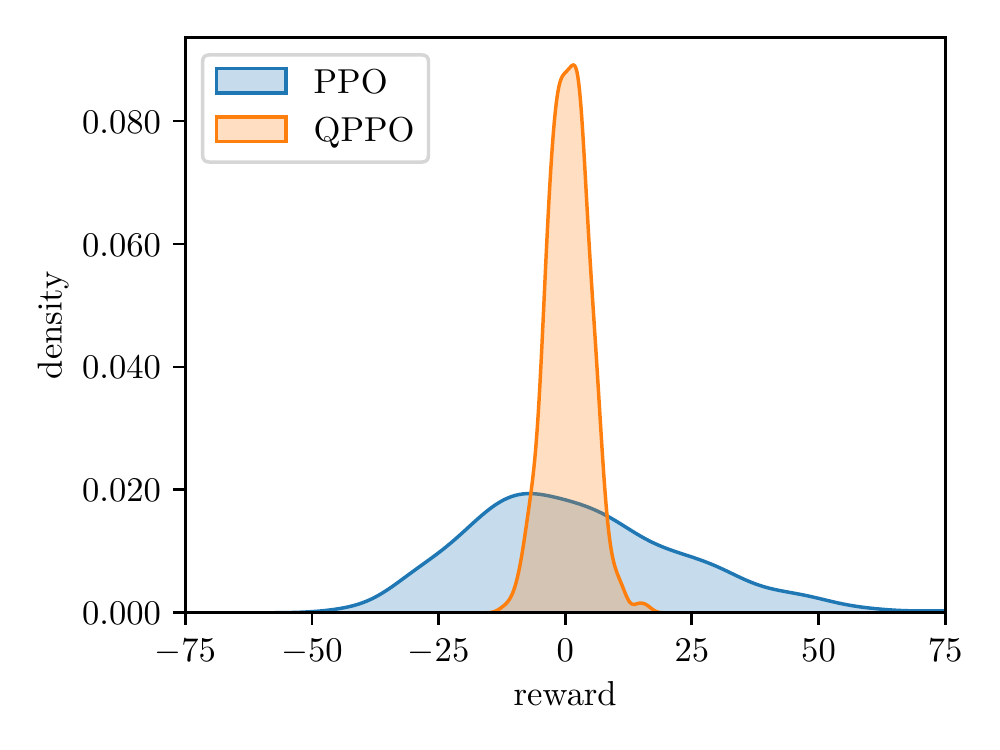}\\
        \end{minipage}
    }
    \caption{Comparison of PPO and QPPO in imperfectly hedgeable Portfolio Management example.}
    \label{fig: learning_curve_kde_5}
\end{figure}

We conducted experiments in two investment portfolio management scenarios distinguished by whether the portfolio risk can be perfectly hedged or not. The volatility for each asset increases with the growth of the drift. In the perfectly hedgeable scenario, alternative assets include a low-risk asset and a pair of high-risk ones that can be fully hedged. In the imperfectly hedgeable scenario, alternative assets include a low-risk asset, a pair of medium-risk ones, and a pair of high-risk ones, but the risk cannot be fully hedged. The policy and baseline are represented by neural networks consisting of three fully connected layers. The agent invests into $3$ or $5$ assets in $100$ time steps and the observed statistics are estimated over past $25$ steps.
The learning curves of PPO and QPPO calculated by $5$ independent experiments are presented in Figures \ref{fig: learning_curve_kde_3}(a, b) and \ref{fig: learning_curve_kde_5}(a, b), where (a) shows the $0.1$-quantile learning curve and (b) shows the average learning curve. The quantile and average performances are estimated by rewards in the past $200$ episodes. The shaded areas represent the $95\%$ confidence intervals. We then test the agents after training for $1000$ replications and plot the kernel density estimation (KDE) in Figures \ref{fig: learning_curve_kde_3}(c) and \ref{fig: learning_curve_kde_5}(c). 
Although PPO leads to a slight advantage in average rewards, the advantage of QPPO in terms of 0.1-quantile is significant. Particularly, note that the 0.1-quantile learning curves of PPO even decrease with more training episodes. The advantage of QPPO appears to be more significant in the imperfectly hedgeable scenario which is more challenging in risk management. Moreover, the KDE of rewards obtained by QPPO is much more concentrated relative to that obtained by PPO, which means that the extreme outcomes would appear much 
less often by using QPPO.

\begin{figure}[!htb]
    \centering
    \subfigure[Price curves]{
        \begin{minipage}[t]{0.3\linewidth}
            \centering
            \includegraphics[width=\linewidth]{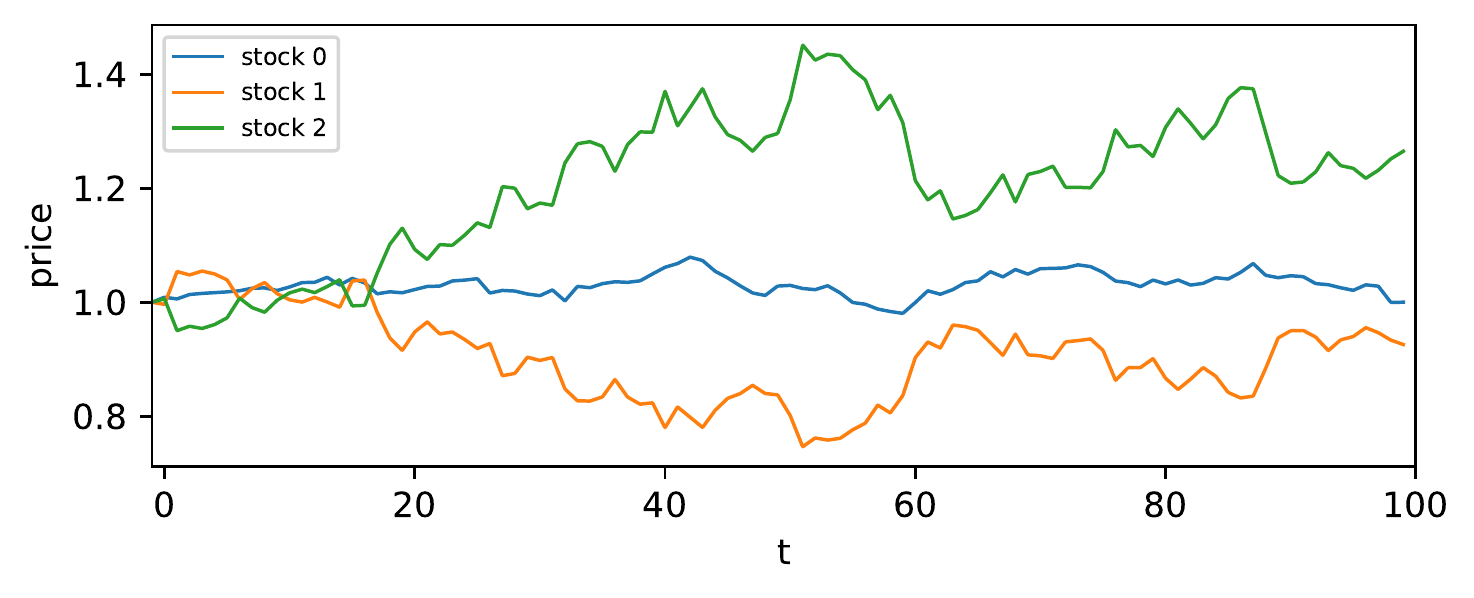}\\
        \end{minipage}
    }
    \subfigure[Optimal mean-based asset allocation solved by Makowitz model]{
        \begin{minipage}[t]{0.3\linewidth}
            \centering
            \includegraphics[width=\linewidth]{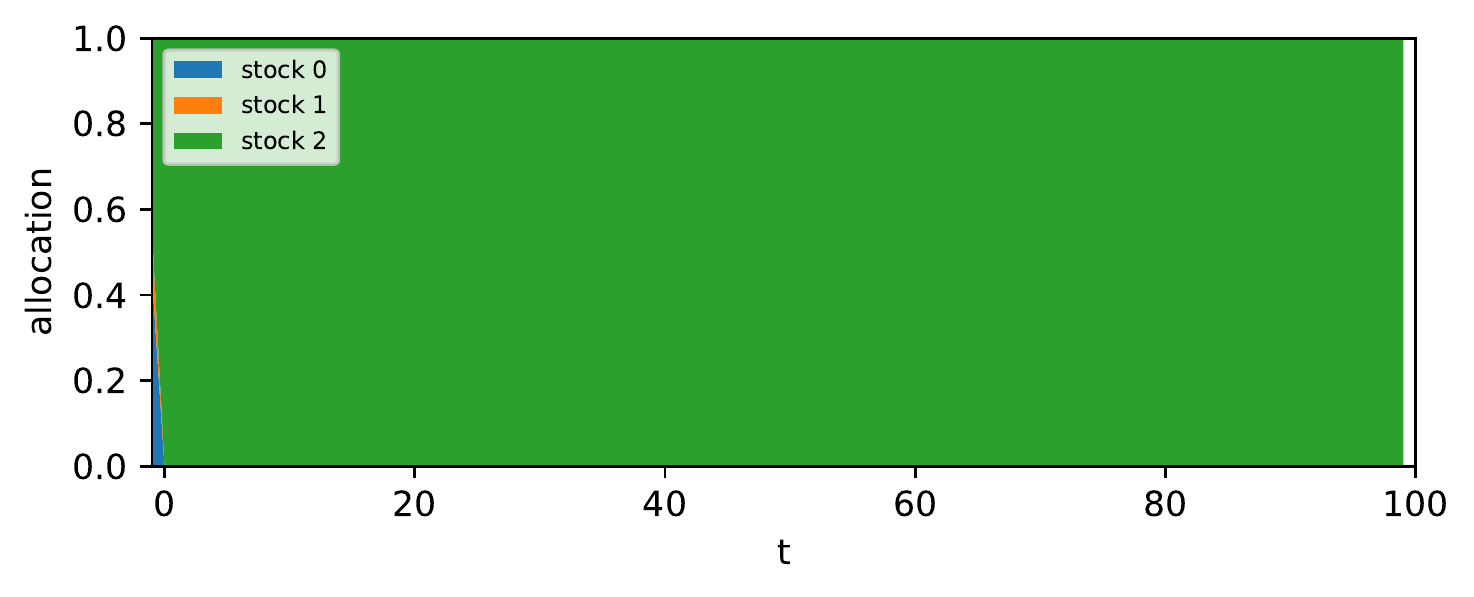}\\
        \end{minipage}
    }
    \subfigure[Optimal quantile-based asset allocation solved by Makowitz model]{
        \begin{minipage}[t]{0.3\linewidth}
            \centering
            \includegraphics[width=\linewidth]{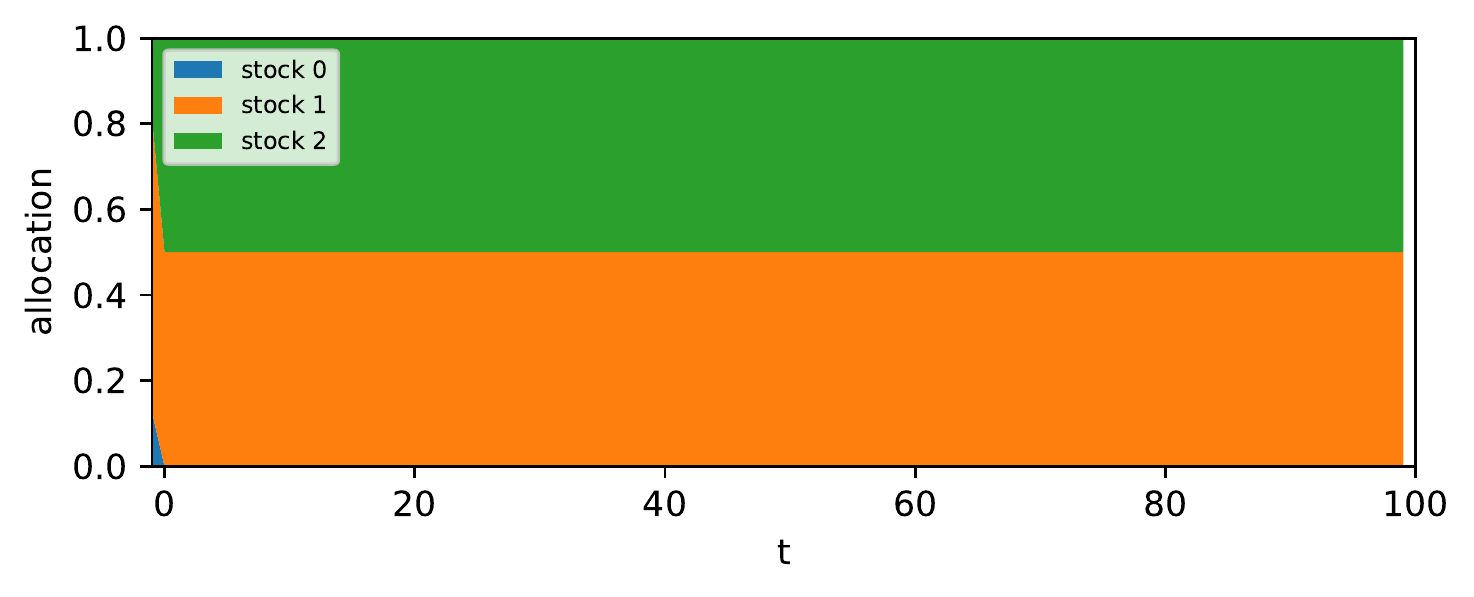}\\
        \end{minipage}
    }
    \subfigure[PPO solution]{
        \begin{minipage}[t]{0.42\linewidth}
            \centering
            \includegraphics[width=\linewidth]{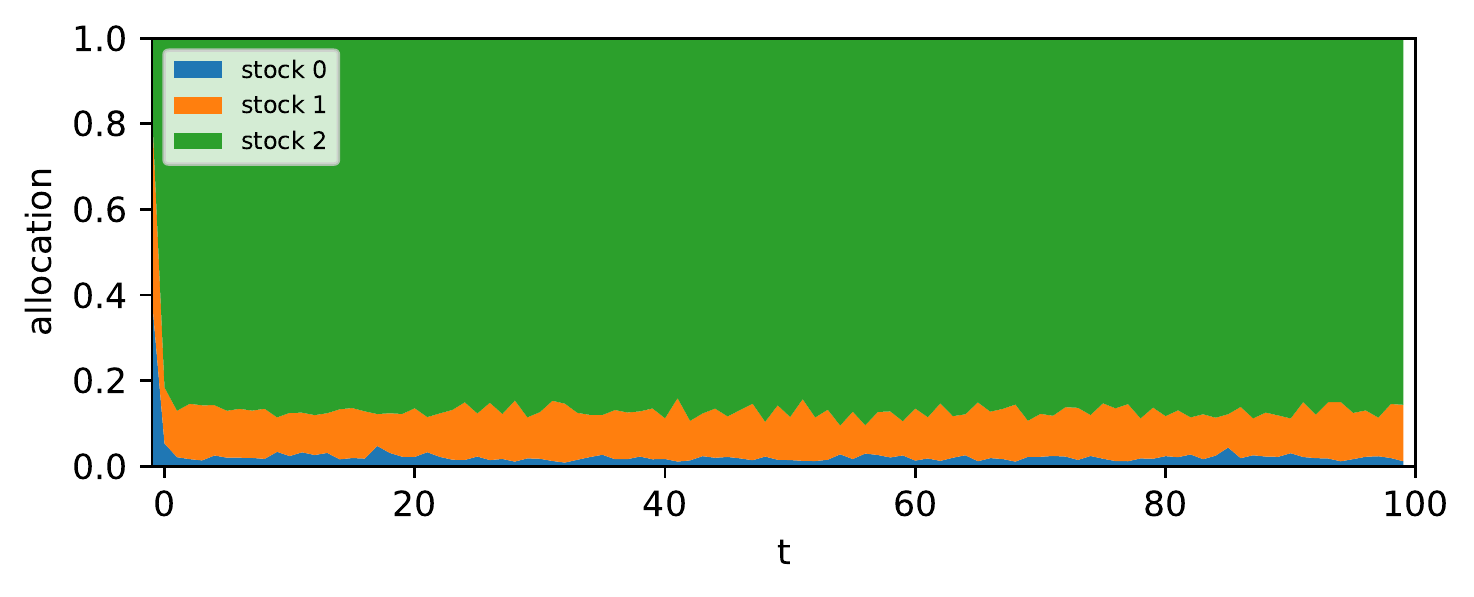}\\
            \includegraphics[width=\linewidth]{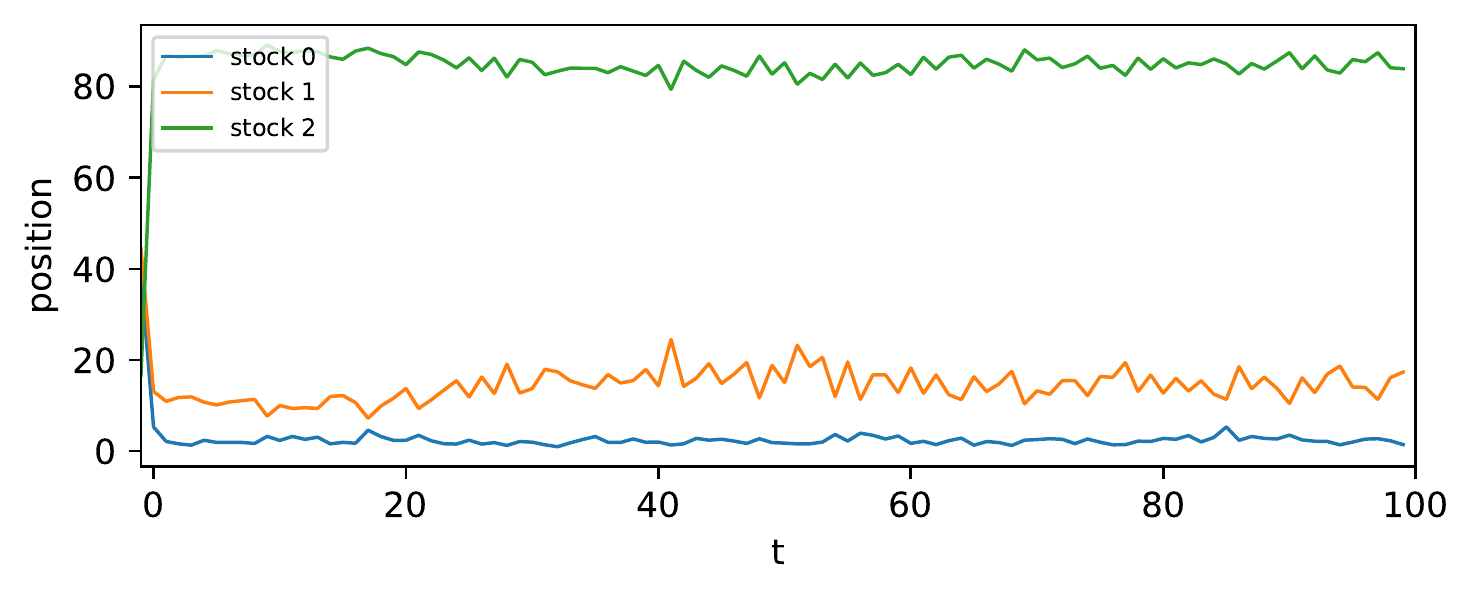}\\
        \end{minipage}
    }
    \subfigure[QPPO solution]{
        \begin{minipage}[t]{0.42\linewidth}
            \centering
            \includegraphics[width=\linewidth]{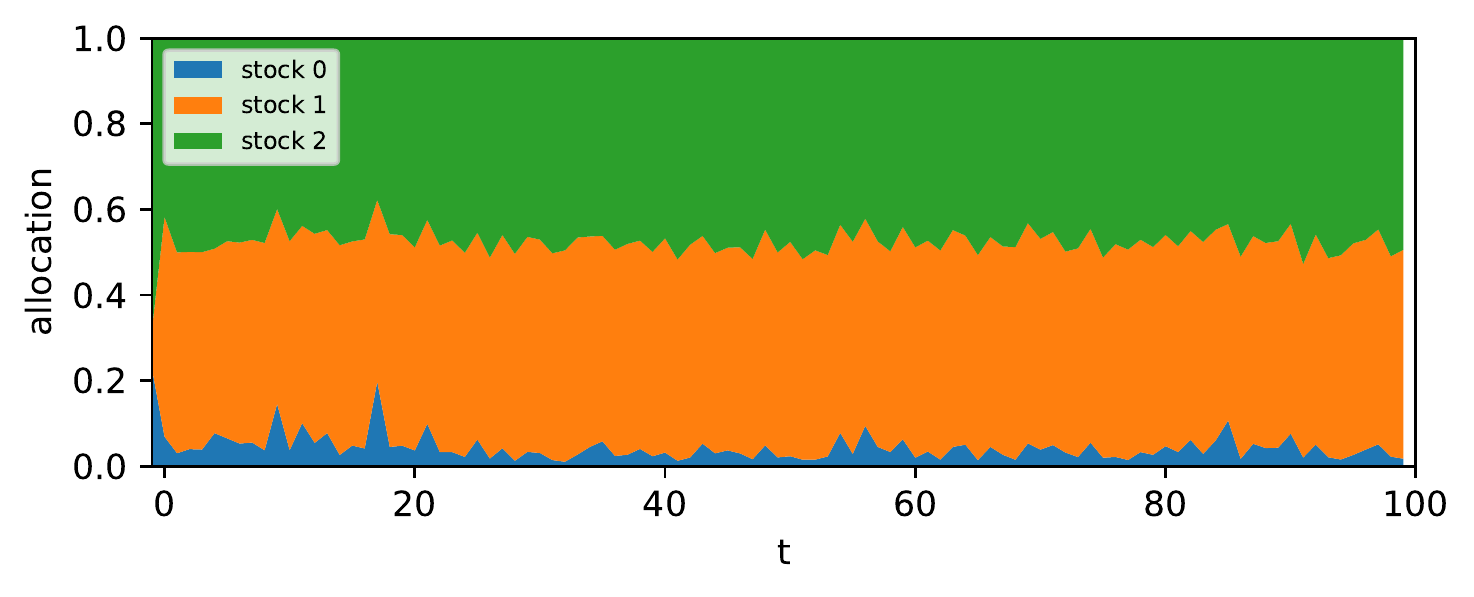}\\
            \includegraphics[width=\linewidth]{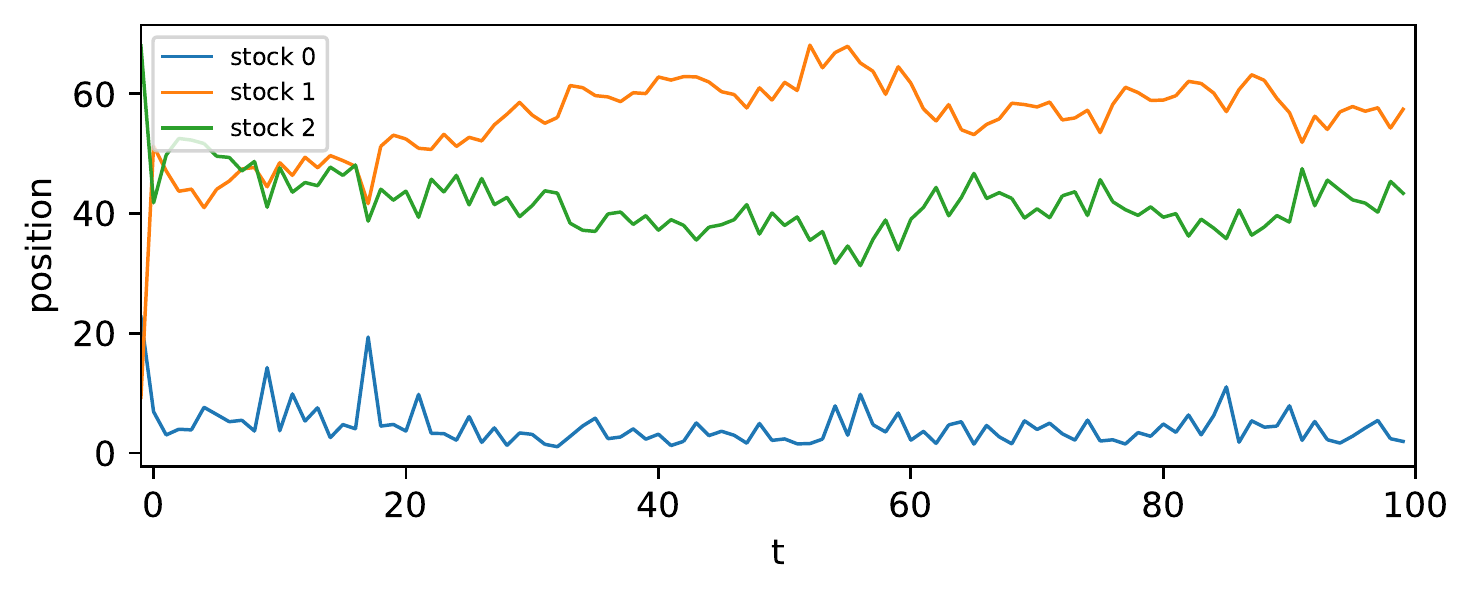}\\
        \end{minipage}
    }
    \caption{Policy visualization for Markowitz model, PPO and QPPO in perfectly hedgeable Portfolio Management example.}
    \label{fig: visual_3}
\end{figure}

\begin{figure}[!htb]
    \centering
    \subfigure[Price curves]{
        \begin{minipage}[t]{0.3\linewidth}
            \centering
            \includegraphics[width=\linewidth]{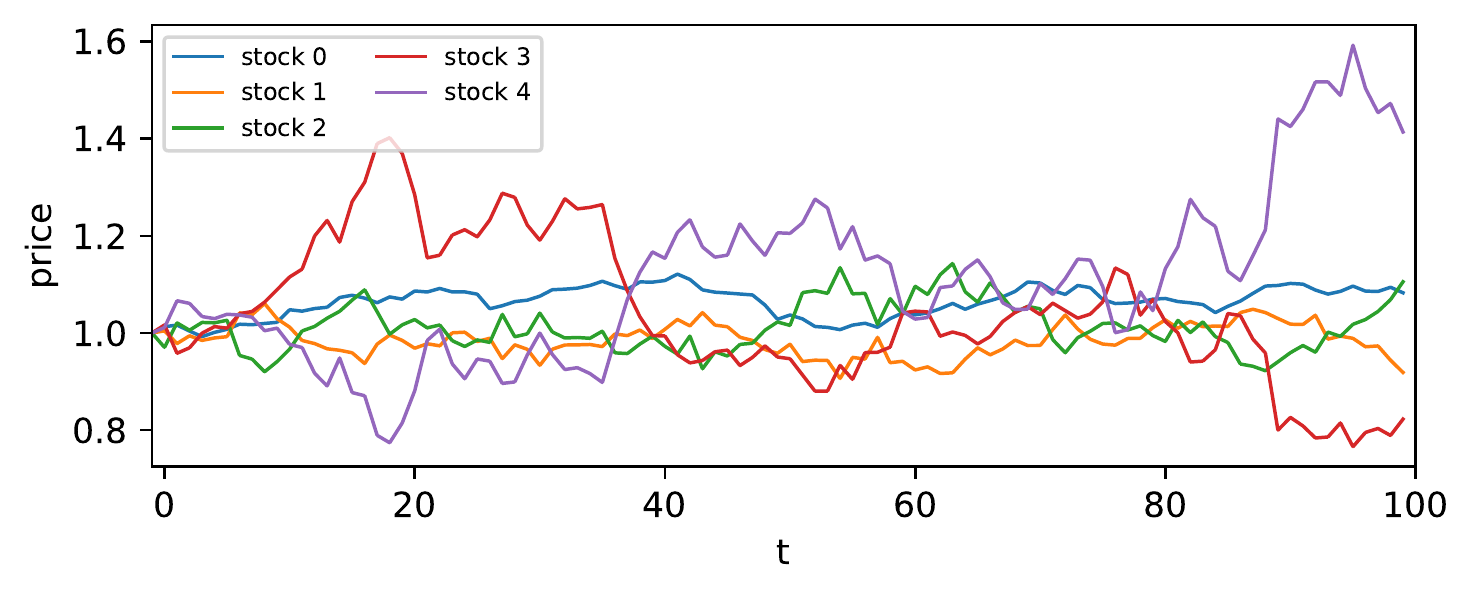}\\
        \end{minipage}
    }
    \subfigure[Optimal mean-based asset allocation solved by Makowitz model]{
        \begin{minipage}[t]{0.3\linewidth}
            \centering
            \includegraphics[width=\linewidth]{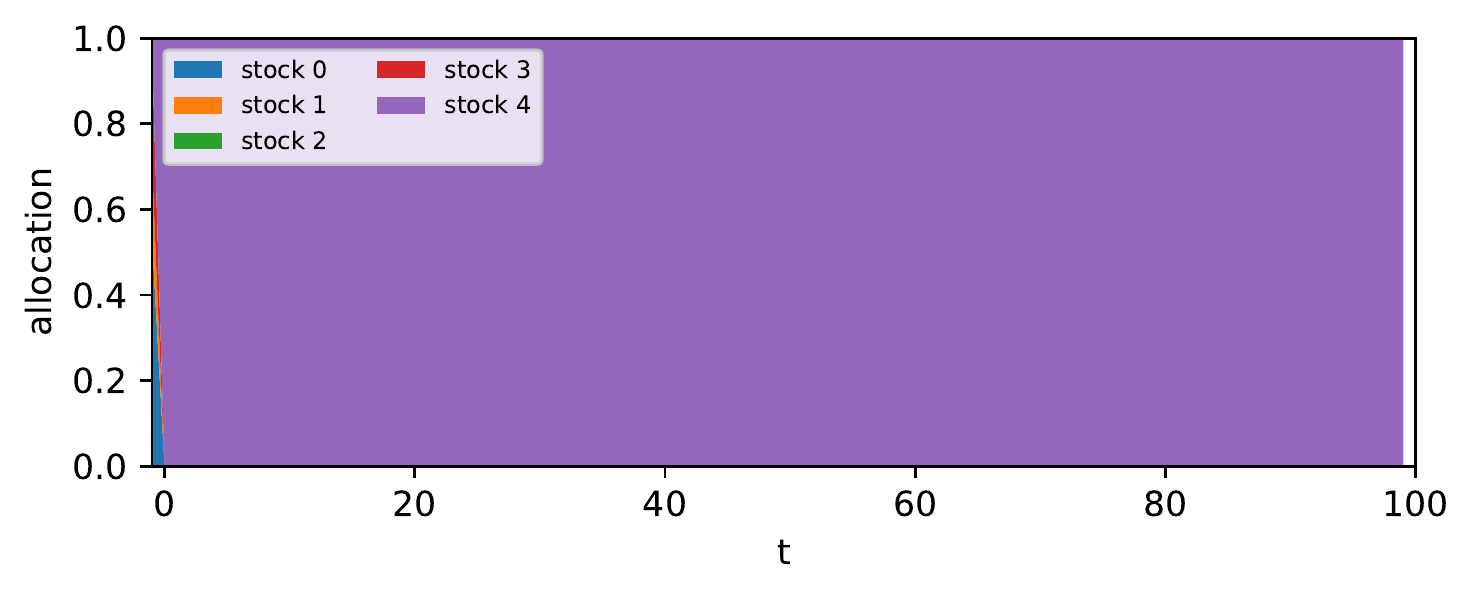}\\
        \end{minipage}
    }
    \subfigure[Optimal quantile-based asset allocation solved by Makowitz model]{
        \begin{minipage}[t]{0.3\linewidth}
            \centering
            \includegraphics[width=\linewidth]{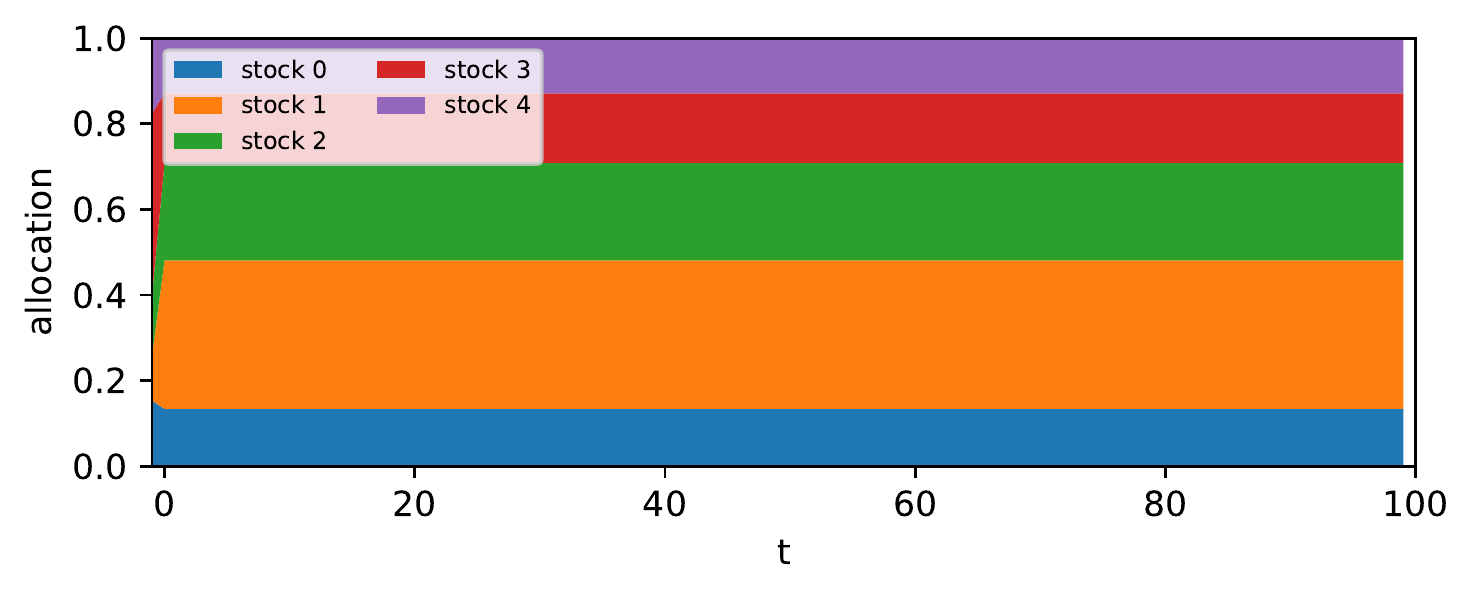}\\
        \end{minipage}
    }
    \subfigure[PPO solution]{
        \begin{minipage}[t]{0.42\linewidth}
            \centering
            \includegraphics[width=\linewidth]{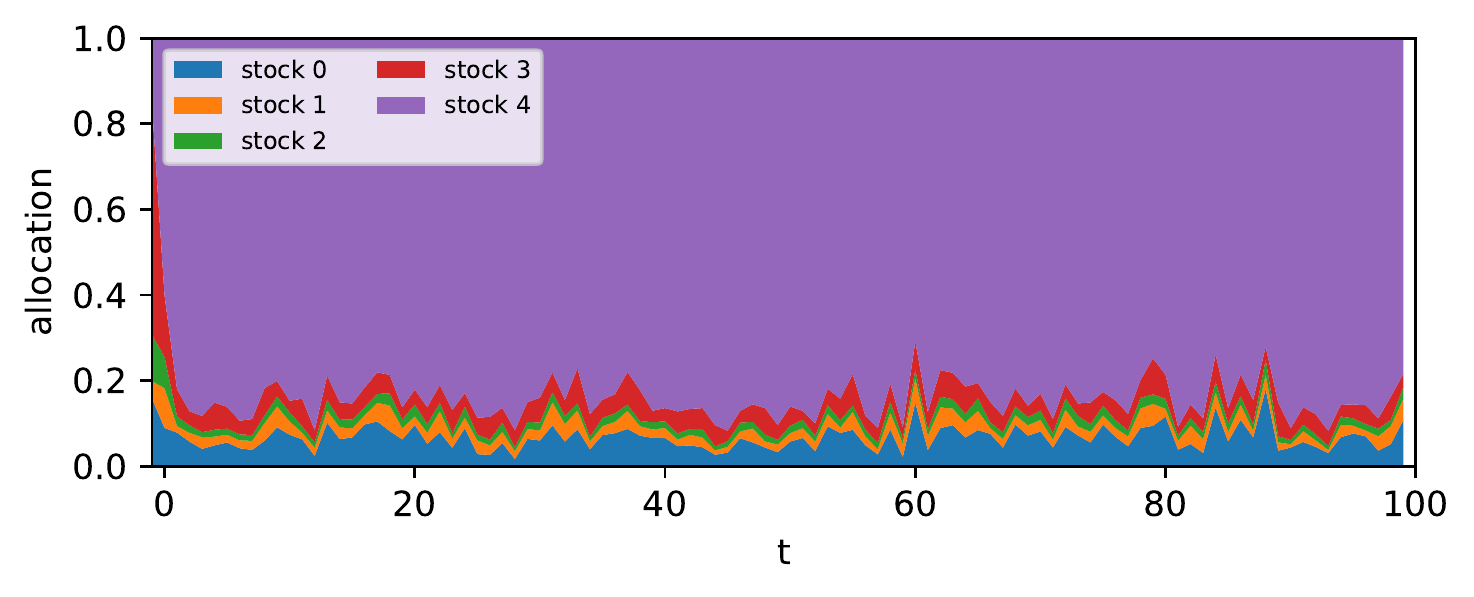}\\
            \includegraphics[width=\linewidth]{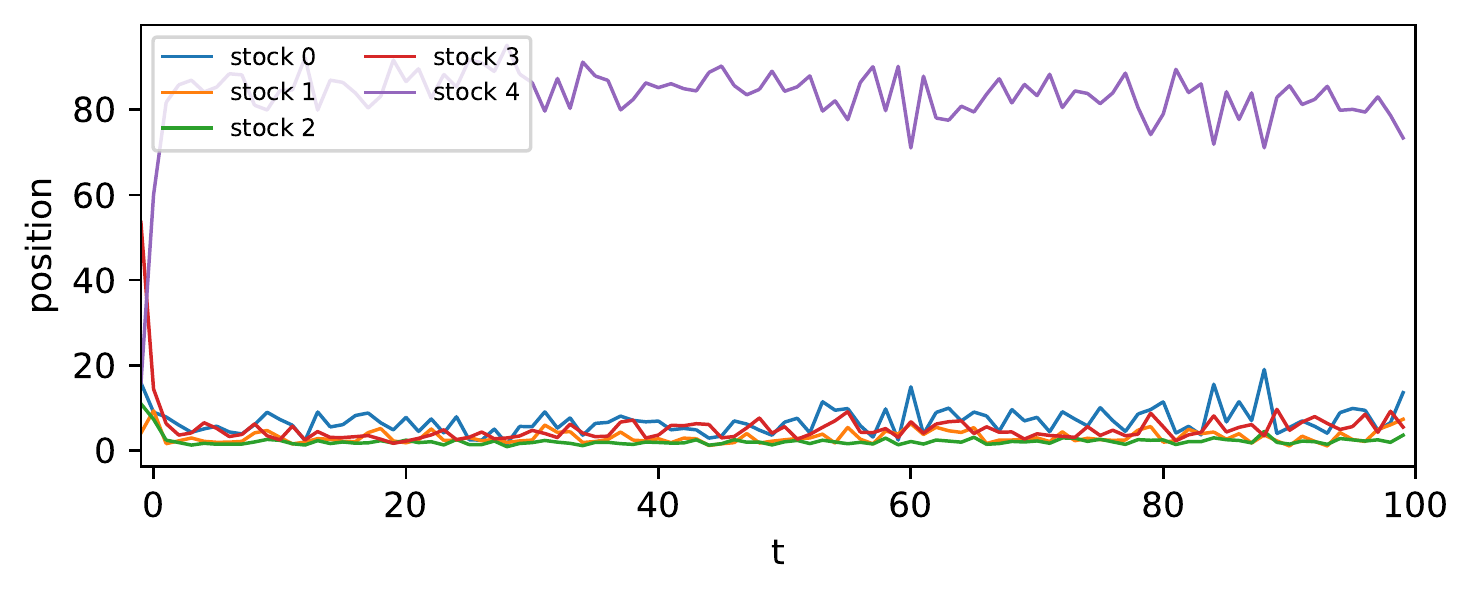}\\
        \end{minipage}
    }
    \subfigure[QPPO solution]{
        \begin{minipage}[t]{0.42\linewidth}
            \centering
            \includegraphics[width=\linewidth]{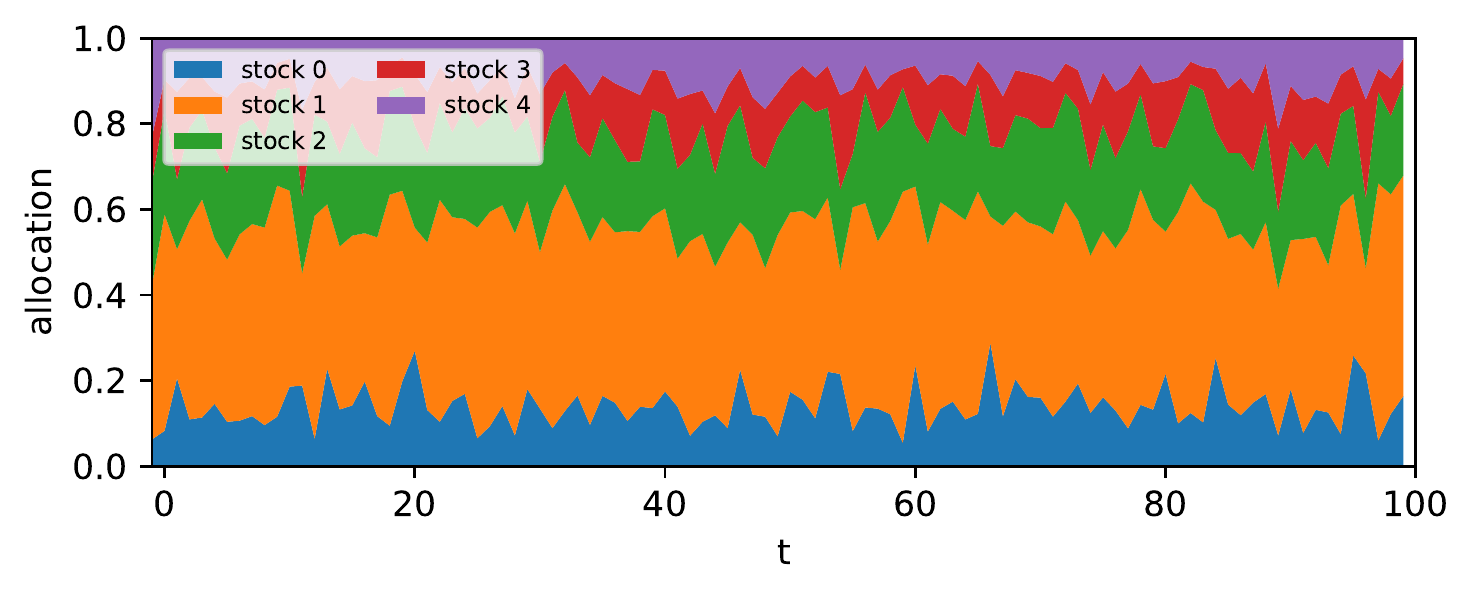}\\
            \includegraphics[width=\linewidth]{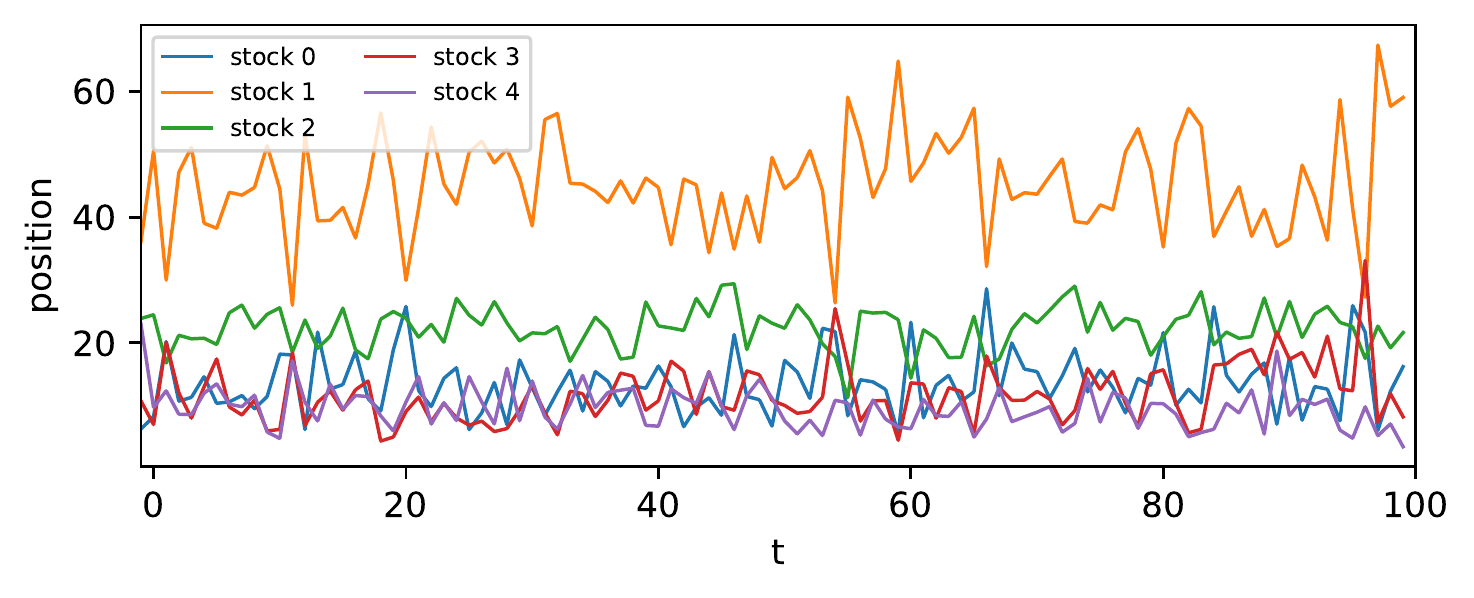}\\
        \end{minipage}
    }
    \caption{Policy visualization for Markowitz model, PPO and QPPO in imperfectly hedgeable Portfolio Management example.}
    \label{fig: visual_5}
\end{figure}

To further investigate the policies of the two agents, we fine tune their networks and visualize their behavior in Figures \ref{fig: visual_3} and \ref{fig: visual_5}. Note that the profit margin $m_t$ follows a multivariate Gaussian distribution $\mathcal{N}(\mu\Delta t ,\Sigma \Delta t)$, so the quantile can be calculated directly by a linear combination of the real mean and variance. Therefore, we can solve the best asset allocation under both criteria by Markowitz model \citep{fabozzi2008portfolio} ignoring {\color{black}the compound interest and transaction fee in reinvestment}. As shown in Figures \ref{fig: visual_3} and \ref{fig: visual_5}, both algorithms reproduce the results of Markowitz model. With the help of deep RL, we are able to solve optimal quantile-based asset allocation problem after taking market frictions into consideration. 
In both scenarios, the agent trained by PPO invests all cash into the asset with the highest expected return, whereas the agent trained by QPPO invests in assets with negative correlations to hedge against risks. 
The slight difference between the position determined by the Markowitz model and that determined by QPPO is because of  the fact that transaction fee is ignored by the the Markowitz model while it is taken into account by QPPO, and the position determined by the Markowitz model only solves a static quantile optimization while the position determined by the QPPO algorithm is the outcome of an optimal policy for an MDP under the quantile criterion.


\subsection{Inventory Management}\label{Subsection Inventory Management}
Solving multi-echelon inventory management by deep RL has attracted attention in operation management recently \citep{gijsbrechts2022can}. The existing literature considers optimizing the ordering policy under the criterion of expected inventory profit. We focus on an inventory management problem with lost sales in a multi-echelon supply chain system, which is more difficult than the backlogging case because the optimal decision depends on the entire inventory pipeline. 

We consider an $N$-echelon supply chain during $T$ periods and index echelons by integers from $0$ to $N+1$, where the first and last echelons are the customer and manufacturer, respectively, and the rest represents the intermediate echelons. We use $S_t^i$, $U_t^i$ and $I_t^i$ to denote the goods shipped by echelon $i$, the lost sales and on-hand inventory of echelon $i$ at the end of period $t$. The quantity ordered by echelon $i$ from echelon $i+1$ is denoted as $q_t^i$, with $q_t^0$ being the customer's demand.
In each period $t$, the agent can control the ordering quantities $q_t=(q_t^1,\cdots,q_t^N)$ of intermediate echelons. For $t=1,\cdots,T$, the shipped quantities and lost sales of the intermediate echelons $i=1,\cdots,N$ are given by
\begin{align*}
    S_t^i=q_t^{i-1}-\left[q_t^{i-1}-I_{t-1}^i-S_{t-L^i}^{i+1}\right]^{+},\quad 
    U_t^i=\left[q_t^{i-1}-S_t^i\right]^{+},
\end{align*}
where $L_i$ denotes the lead time for shipping goods in echelon $i$. Then the on-hand inventory is updated by 
\begin{align*}
    I_t^i=\left[I_{t-1}^i+S_{t-L^i}^{i+1}-S_t^i\right]^{+}.
\end{align*}
Note that the manufacturer $N+1$ always has unlimited resources, i.e., $S_t^{N+1} = q_t^{N}$. The profit of each echelon is calculated below:
\begin{align*}
    P_t^i=p^i S_t^i-p^{i+1} S_t^{i+1}-h^i I_t^i-l^i U_t^i,
\end{align*}
where $p^i$, $h^i$ and $l^i$ are the unit price, unit holding cost and unit penalty for lost sales of echelon $i$. The observation state consists of the inventories, lost sales, shipped and ordering quantities in past $L=\max\{L_1,\cdots,L_N\}$ periods, and the rewards of the agent are the total supply chain profits, i.e., $r_t = \sum_{i=1}^N P_t^i$.

\begin{figure}[!htb]
    \centering
    \subfigure[Uniform demand]{
        \begin{minipage}[t]{0.25\linewidth}
            \centering
            \includegraphics[width=\linewidth]{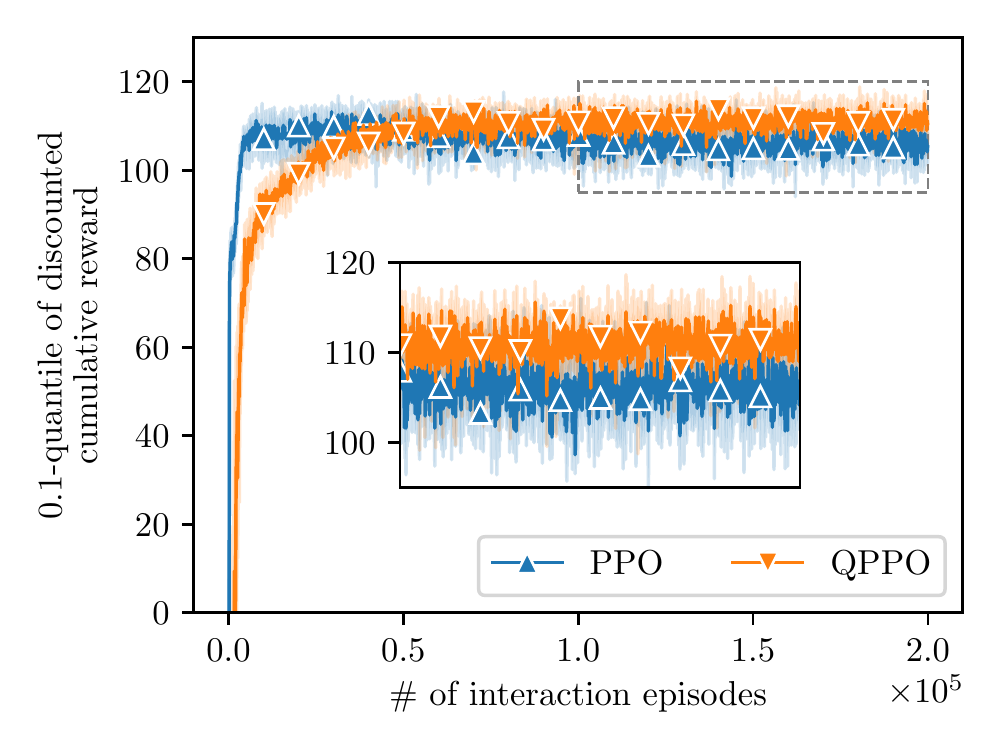}\\
            \includegraphics[width=\linewidth]{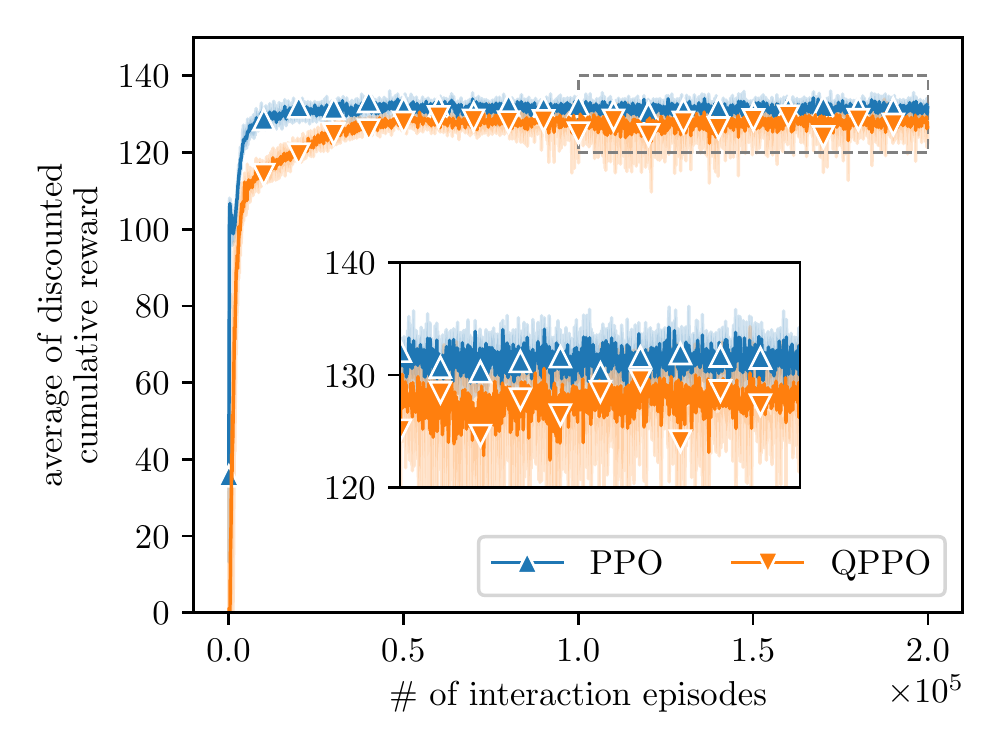}\\
        \end{minipage}
    }
    \subfigure[Merton demand]{
        \begin{minipage}[t]{0.25\linewidth}
            \centering
            \includegraphics[width=\linewidth]{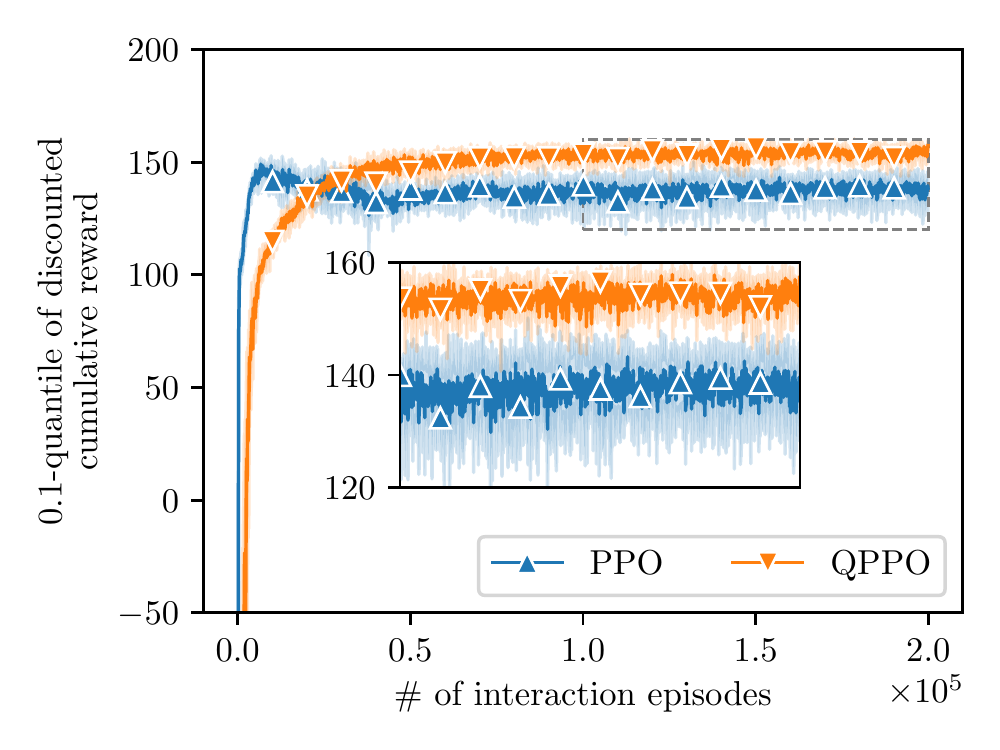}\\
            \includegraphics[width=\linewidth]{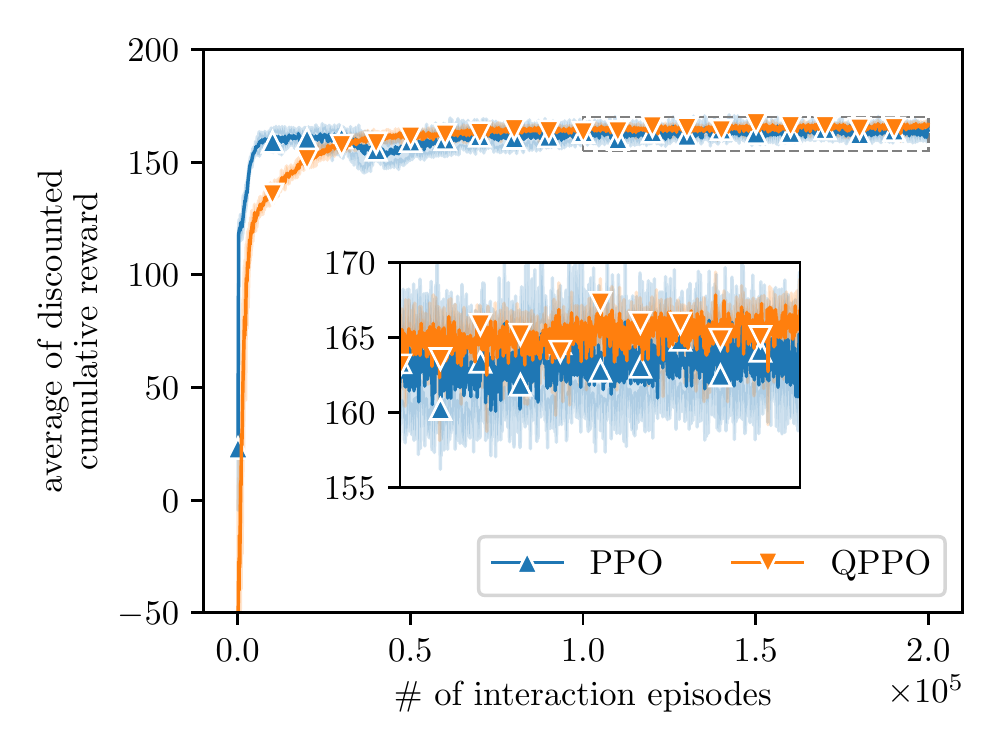}\\
        \end{minipage}
    }
    \subfigure[Periodical demand]{
        \begin{minipage}[t]{0.25\linewidth}
            \centering
            \includegraphics[width=\linewidth]{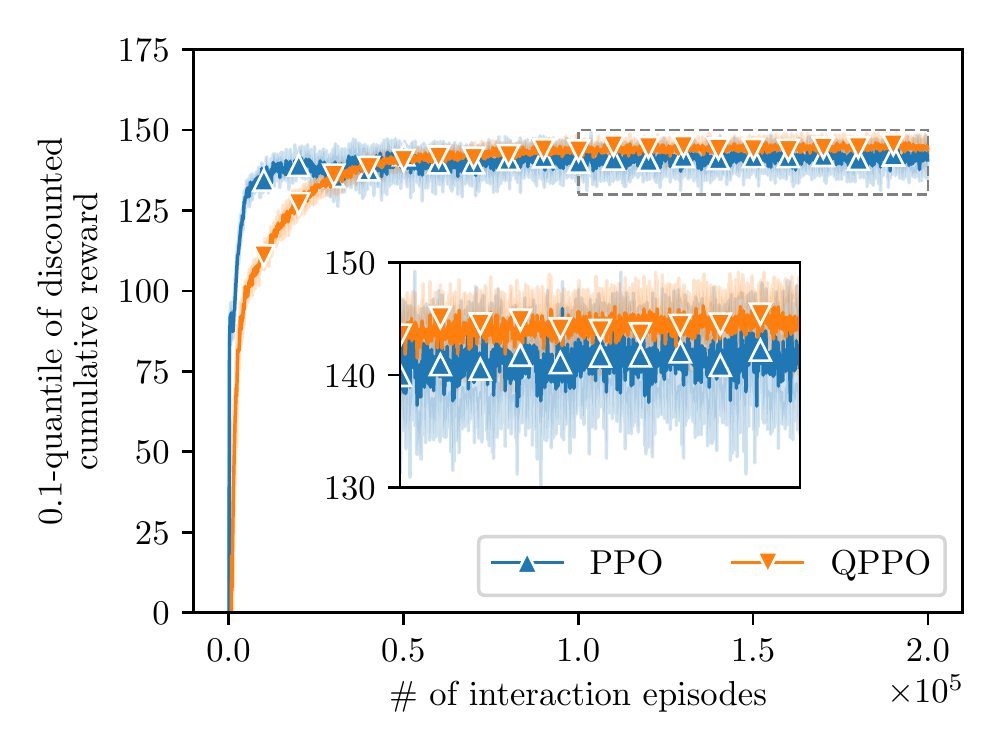}\\
            \includegraphics[width=\linewidth]{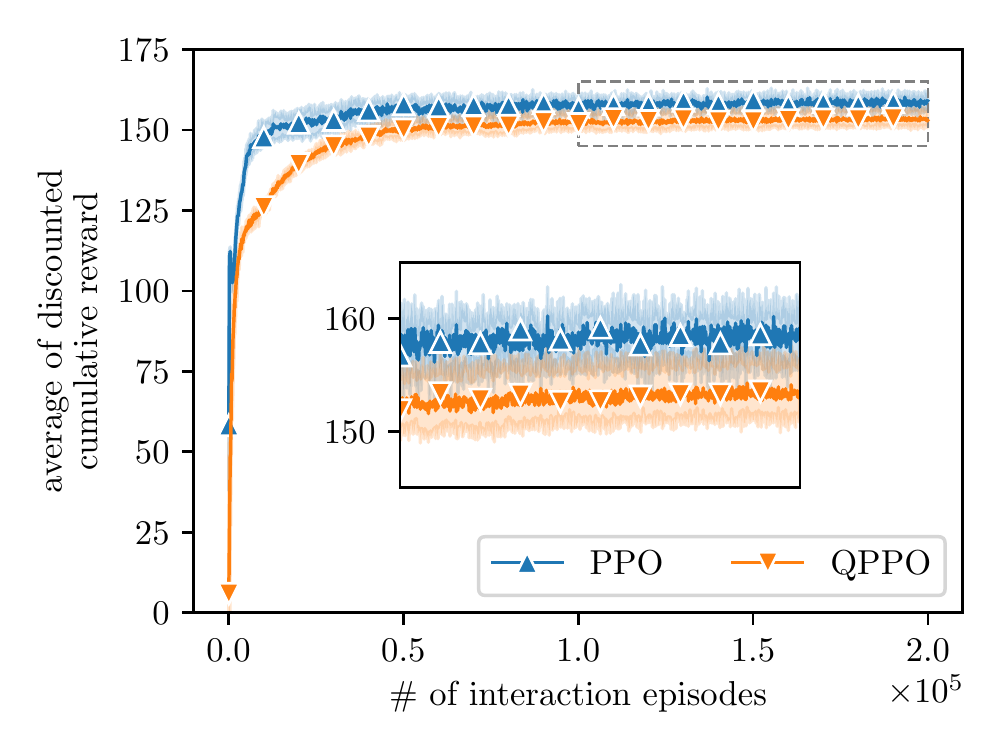}\\
        \end{minipage}
    }
    \caption{Learning curves for quantile ($\alpha=0.1$) and average of PPO and QPPO in single-echelon Inventory Management example by 5 independent experiments.}
    \label{fig: learning_curve_single_im}
\end{figure}

\begin{figure}[!htb]
    \centering
    \subfigure[Uniform demand]{
        \begin{minipage}[t]{0.25\linewidth}
            \centering
            \includegraphics[width=\linewidth]{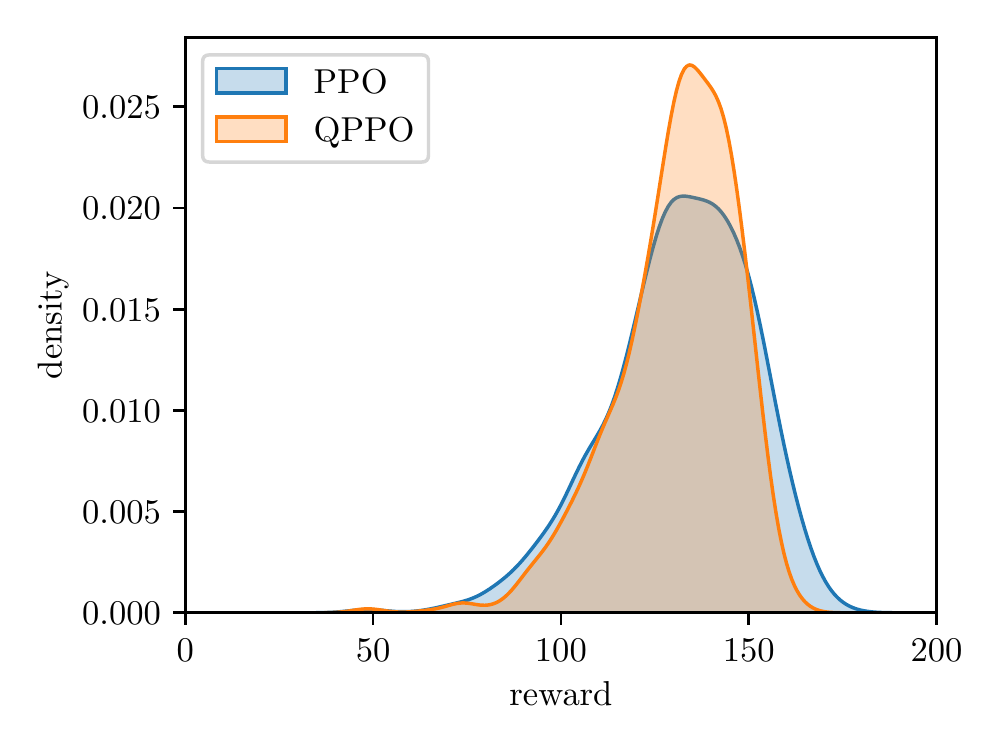}\\
        \end{minipage}
    }
    \subfigure[Merton demand]{
        \begin{minipage}[t]{0.25\linewidth}
            \centering
            \includegraphics[width=\linewidth]{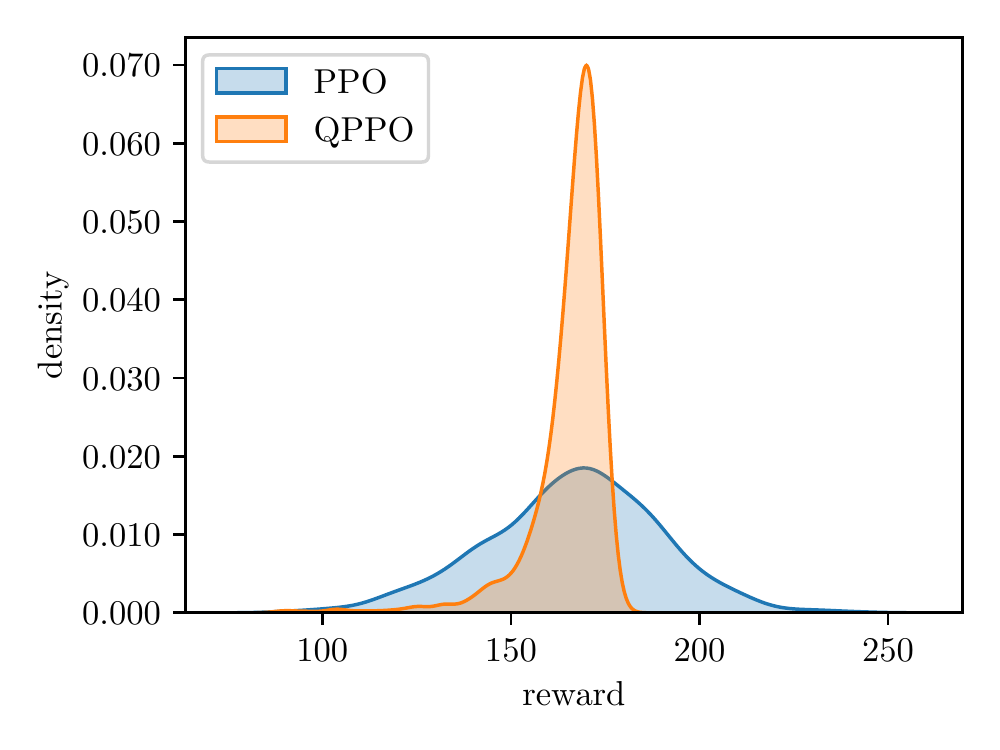}\\
        \end{minipage}
    }
    \subfigure[Periodical demand]{
        \begin{minipage}[t]{0.25\linewidth}
            \centering
            \includegraphics[width=\linewidth]{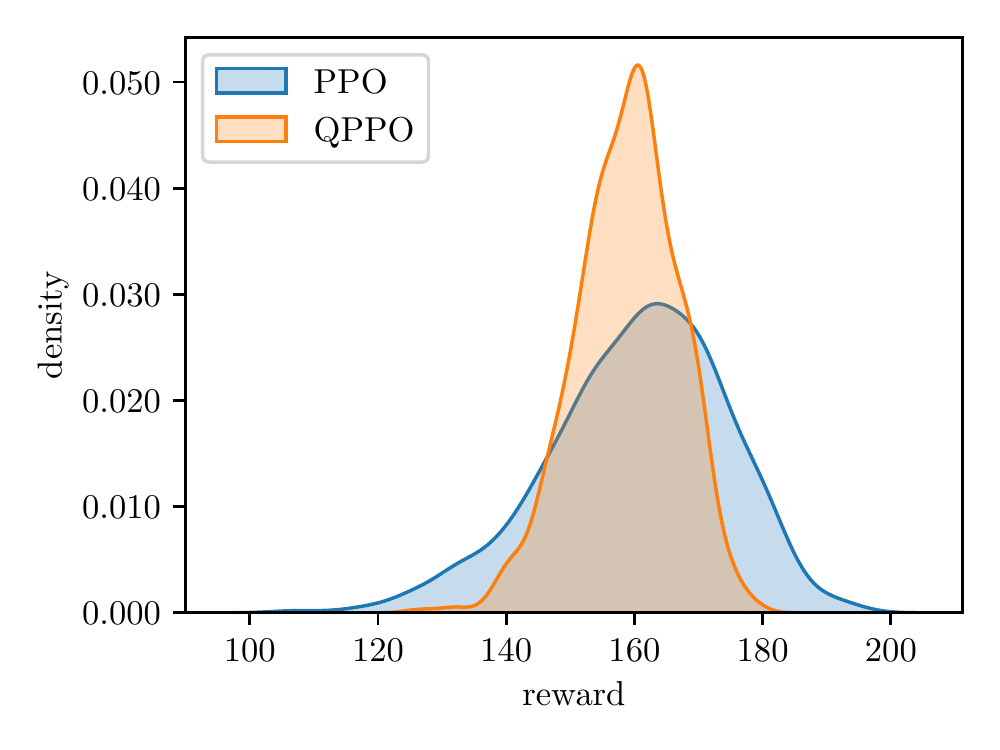}\\
        \end{minipage}
    }
    \caption{KDE plots for rewards of PPO and QPPO in single-echelon Inventory Management examples.}
    \label{fig: kde_single_im}
\end{figure}

\begin{figure}[!htb]
    \centering
    \subfigure[Uniform demand]{
        \begin{minipage}[t]{0.25\linewidth}
            \centering
            \includegraphics[width=\linewidth]{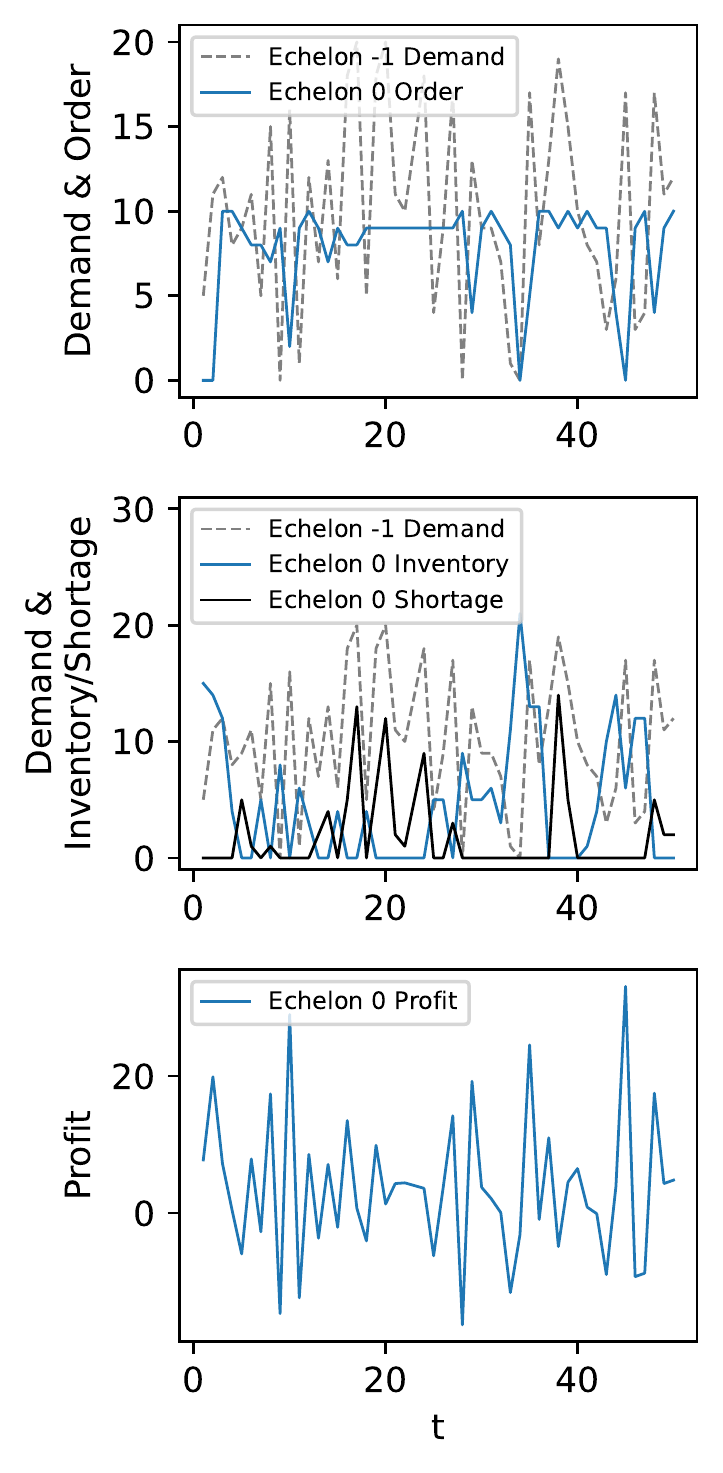}\\
        \end{minipage}
    }
    \subfigure[Merton demand]{
        \begin{minipage}[t]{0.25\linewidth}
            \centering
            \includegraphics[width=\linewidth]{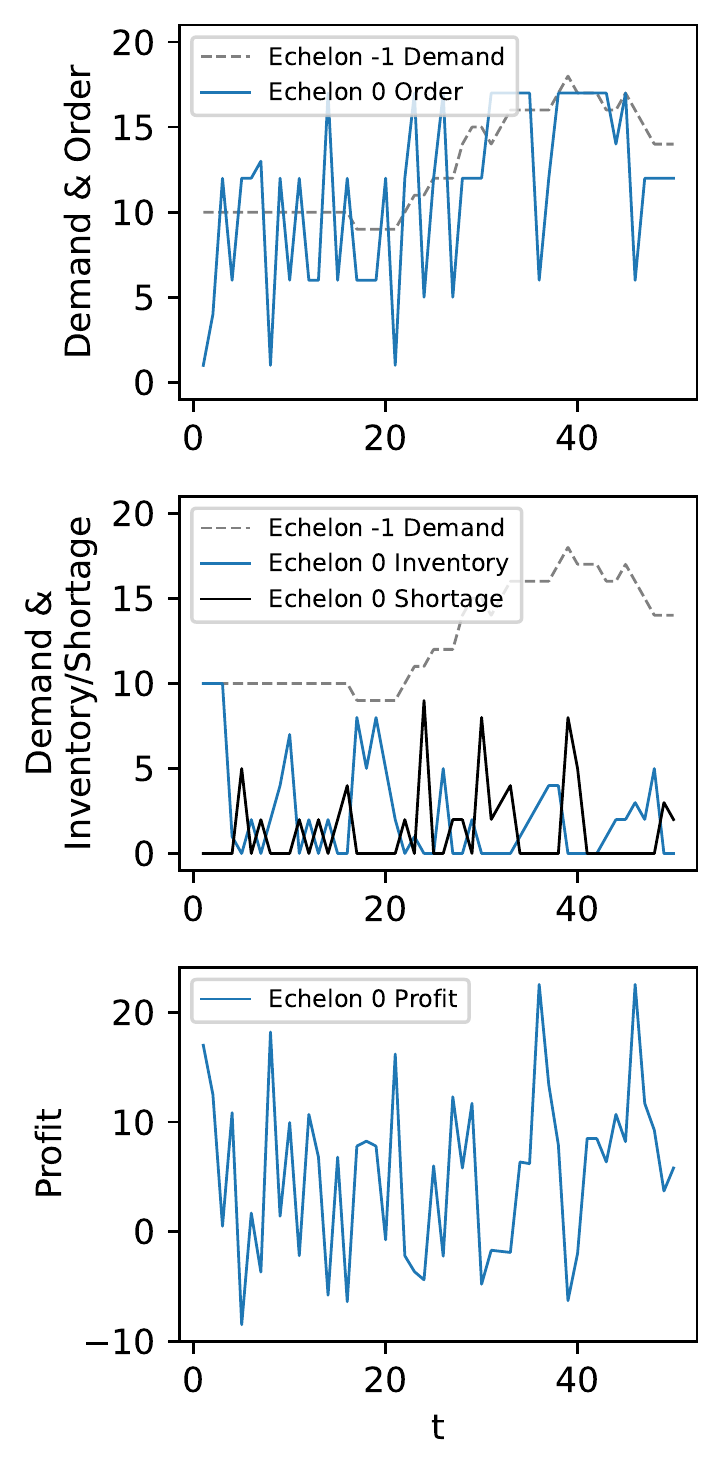}\\
        \end{minipage}
    }
    \subfigure[Periodical demand]{
        \begin{minipage}[t]{0.25\linewidth}
            \centering
            \includegraphics[width=\linewidth]{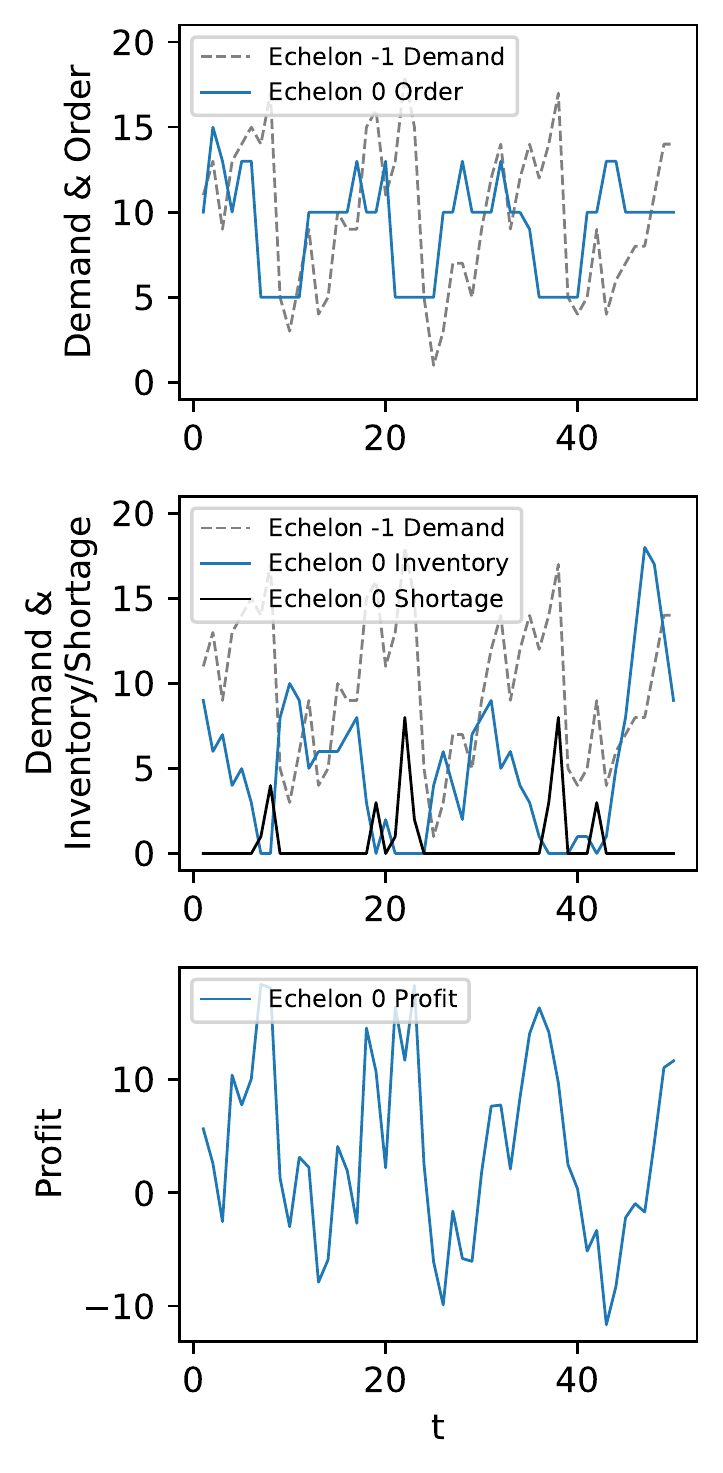}\\
        \end{minipage}
    }
    \caption{Policy visualization of PPO in single-echelon Inventory Management examples.}
    \label{fig: ppo_single_im}
\end{figure}

\begin{figure}[!htb]
    \centering
    \subfigure[Uniform demand]{
        \begin{minipage}[t]{0.25\linewidth}
            \centering
            \includegraphics[width=\linewidth]{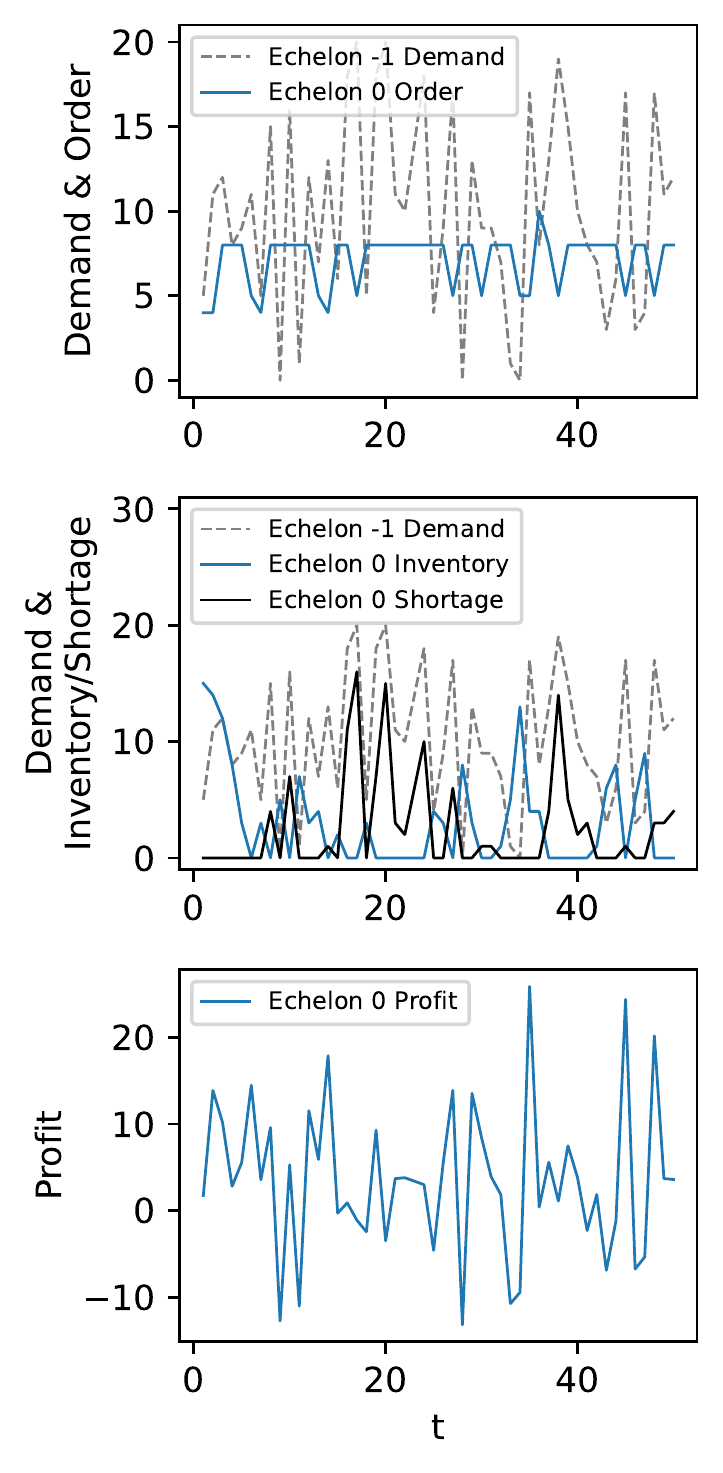}\\
        \end{minipage}
    }
    \subfigure[Merton demand]{
        \begin{minipage}[t]{0.25\linewidth}
            \centering
            \includegraphics[width=\linewidth]{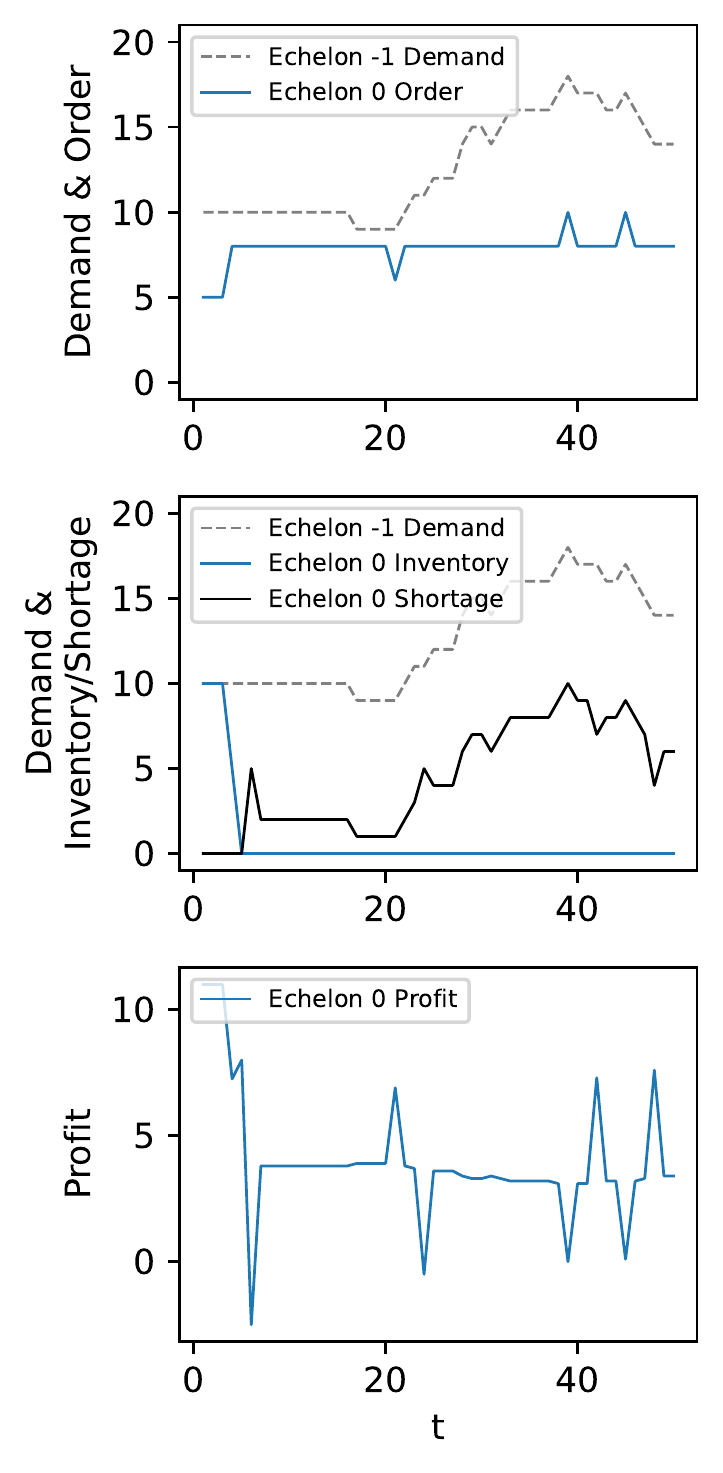}\\
        \end{minipage}
    }
    \subfigure[Periodical demand]{
        \begin{minipage}[t]{0.25\linewidth}
            \centering
            \includegraphics[width=\linewidth]{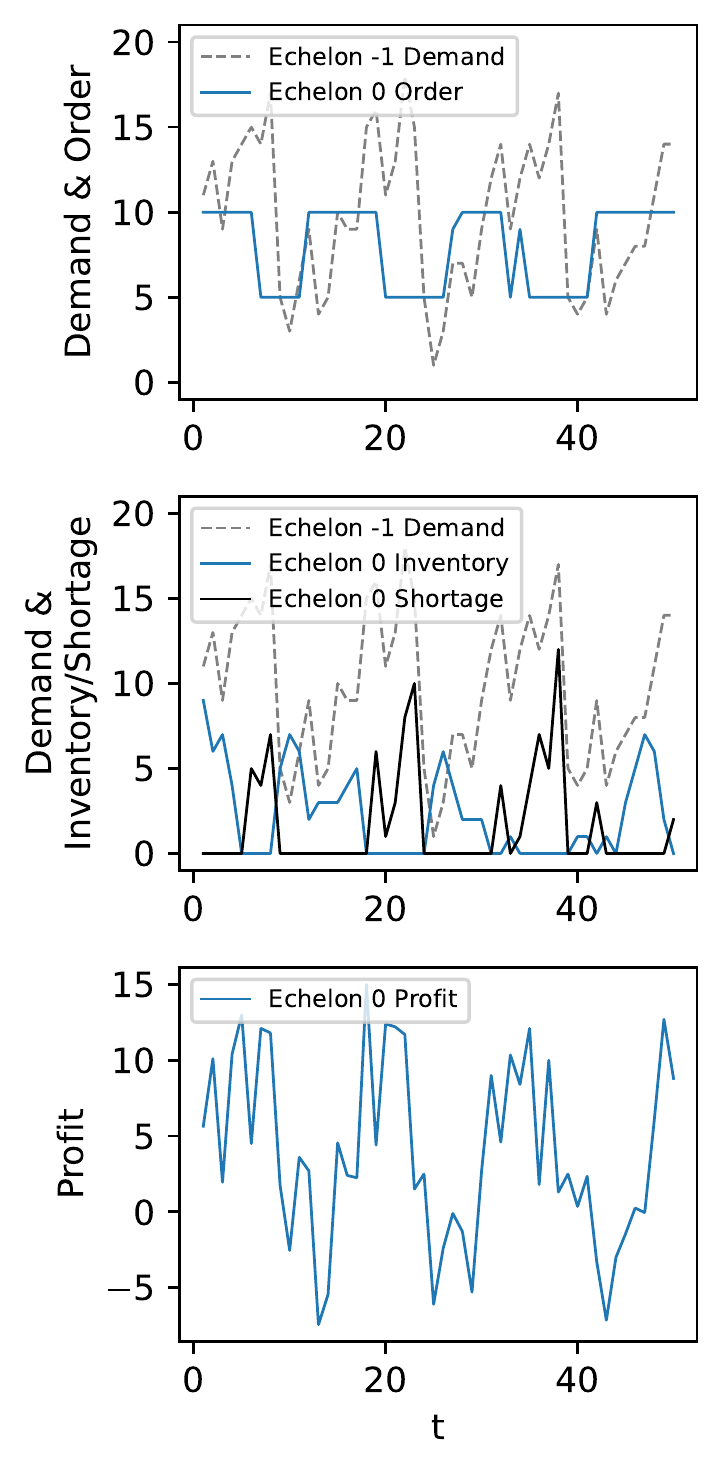}\\
        \end{minipage}
    }
    \caption{Policy visualization of QPPO in single-echelon Inventory Management examples.}
    \label{fig: qppo_single_im}
\end{figure}

We first apply PPO and QPPO to three single-echelon inventory management problems, where customer demands $q_t^0$ are generated from a uniform distribution, Merton jump diffusion model \citep{merton1976option}, and a periodic stochastic model with a saw-wave trend, respectively. 
The policy and baseline are represented by neural networks consisting of a temporal convolutional layer and a fully connected layer. The agent makes reordering decisions in $50$ time steps and can look back at information within the past $3$ steps.
The learning curves of PPO and QPPO are presented in Figure \ref{fig: learning_curve_single_im}. The quantile and average performances are estimated by rewards in the past $100$ episodes. The shaded areas represent the $95\%$ confidence intervals.
The steep initial climb in the learning curve indicates that QPPO has a comparable learning ability as the state-of-the-art mean-based algorithm, PPO. Meanwhile, QPPO achieves higher $0.1$-quantile performance at the end of the training. We test the agents for $1000$ replications after training and present the results in Figure \ref{fig: kde_single_im} and Table \ref{table: single_im}. 
In all single-echelon examples, QPPO achieves much superior $0.1$-quantile performance with only a slight loss in mean compared to PPO, and its KDE is much more concentrated than PPO.
The agents' policies after training can be visualized in Figures \ref{fig: ppo_single_im} and \ref{fig: qppo_single_im}. 
The oscillation of the profits obtained by QPPO is narrower than that of PPO.
Although the performances of PPO and QPPO in simple single-echelon problems occasionally lead to similar statistics, agents' policies are significantly different. The reordering decisions of PPO change dramatically as customer's demand changes, which are difficult to implement in practice. The decisions of QPPO are relatively stable, with limited amplitude and frequency of changes.
\begingroup
\renewcommand{\arraystretch}{0.6}
\setlength{\extrarowheight}{-0.5pt}
\begin{table}[!htb]
\centering
\caption{Testing results in single-echelon Inventory Management examples.}
\label{table: single_im}
\begin{tabular}{ccccc}
\toprule
\multicolumn{1}{c}{\multirow{2}{*}{Demand}} & \multicolumn{2}{c}{$0.1$-quantile} & \multicolumn{2}{c}{Average} \\
 & PPO         & QPPO         & PPO        & QPPO         \\ \midrule
Uniform & $105.93$ & $109.45$ & $131.55$ & $130.81$\\
Merton  & $136.71$ & $155.28$ & $166.44$ & $165.50$\\ 
Periodical  & $142.40$ & $148.56$ & $160.49$ & $158.81$\\ \bottomrule
\end{tabular}
\end{table}
\endgroup

\begin{figure}[!htb]
    \centering
    \subfigure[0.1-quantile learning curves]{
        \begin{minipage}[t]{0.25\linewidth}
            \centering
            \includegraphics[width=\linewidth]{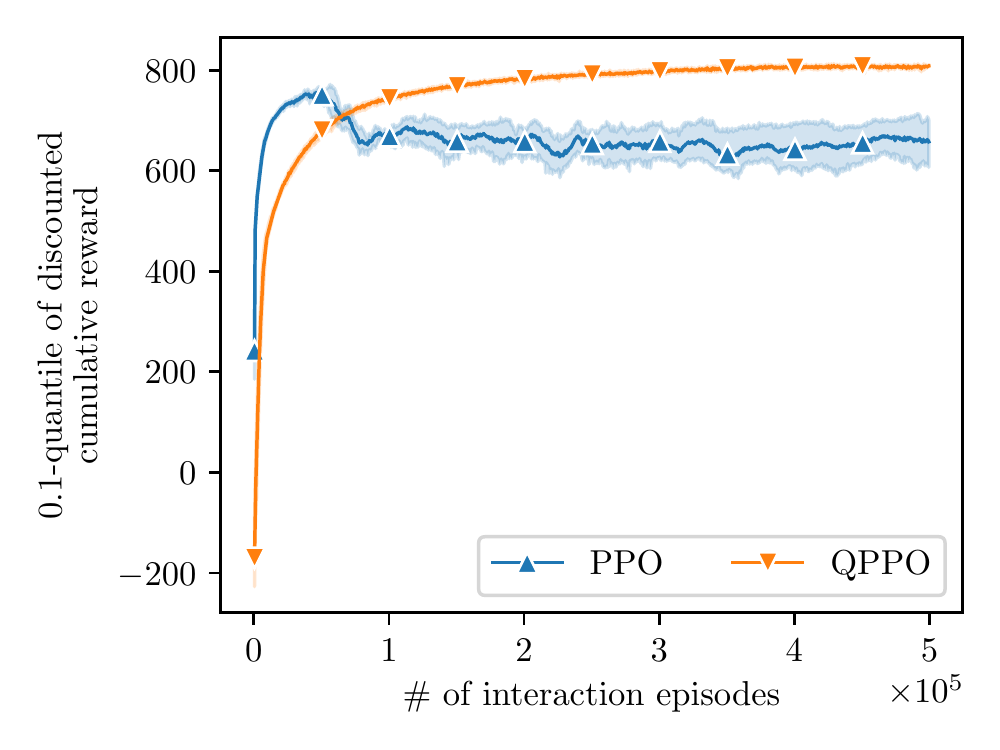}\\
        \end{minipage}
    }
    \subfigure[Average learning curves]{
        \begin{minipage}[t]{0.25\linewidth}
            \centering
            \includegraphics[width=\linewidth]{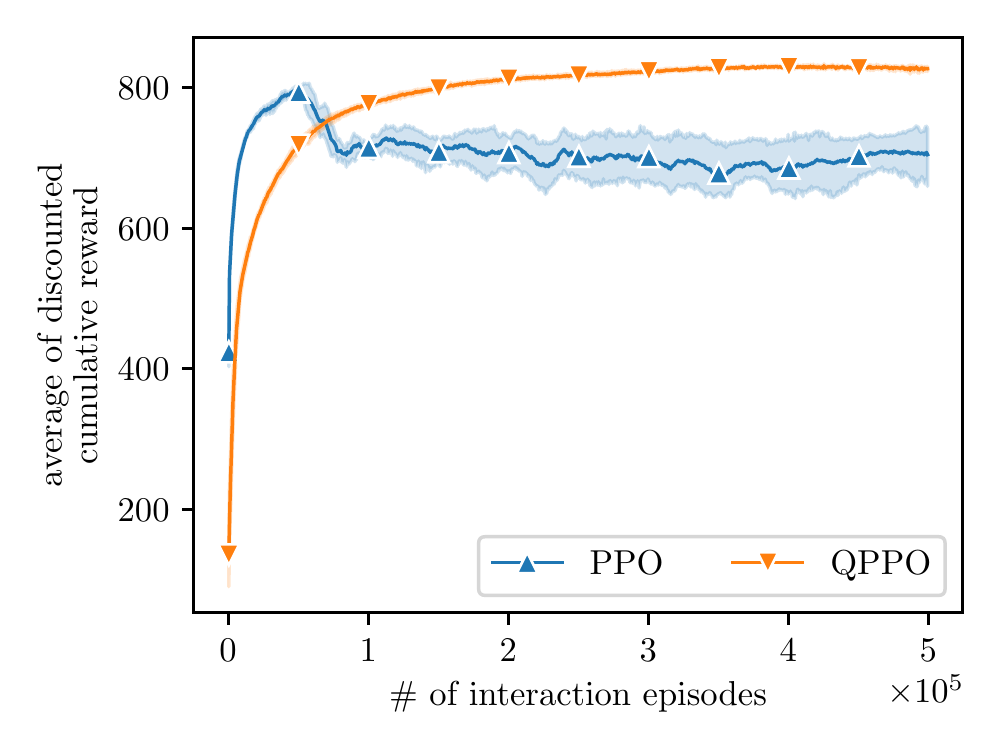}\\
        \end{minipage}
    }
    \subfigure[KDE plots]{
        \begin{minipage}[t]{0.25\linewidth}
            \centering
            \includegraphics[width=\linewidth]{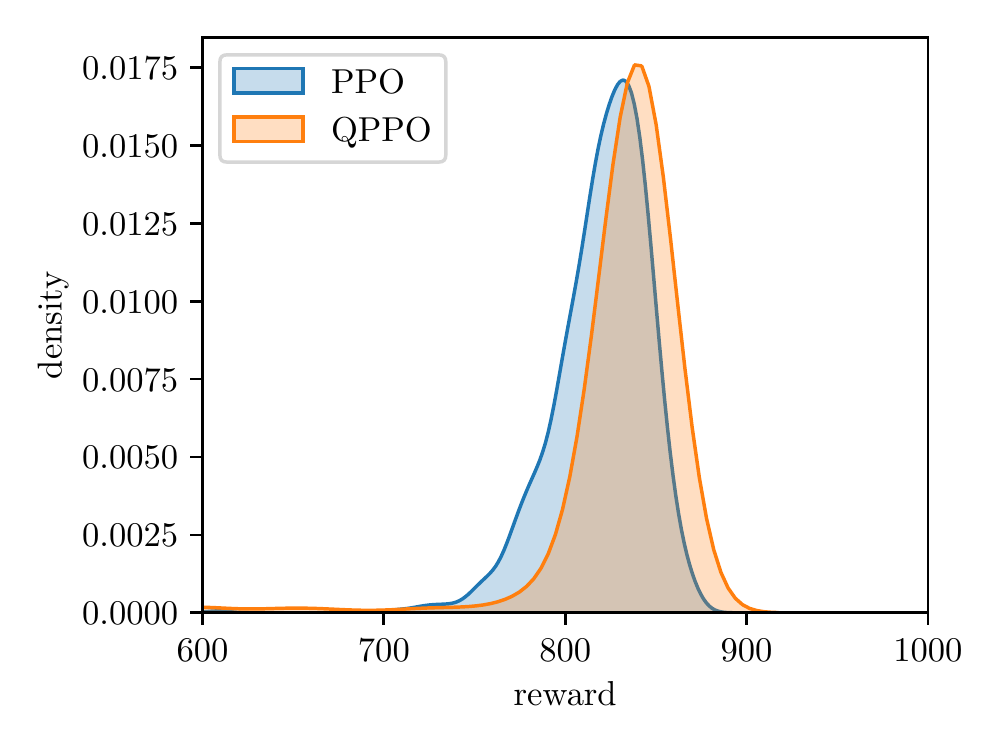}\\
        \end{minipage}
    }
    \caption{Comparison of PPO and QPPO in multi-echelon Inventory Management example.}
    \label{fig: learning_curve_kde_multi_im}
\end{figure}

\begin{figure}[!htb]
    \centering
    \subfigure[PPO]{
        \begin{minipage}[t]{0.6\linewidth}
            \centering
            \includegraphics[width=\linewidth]{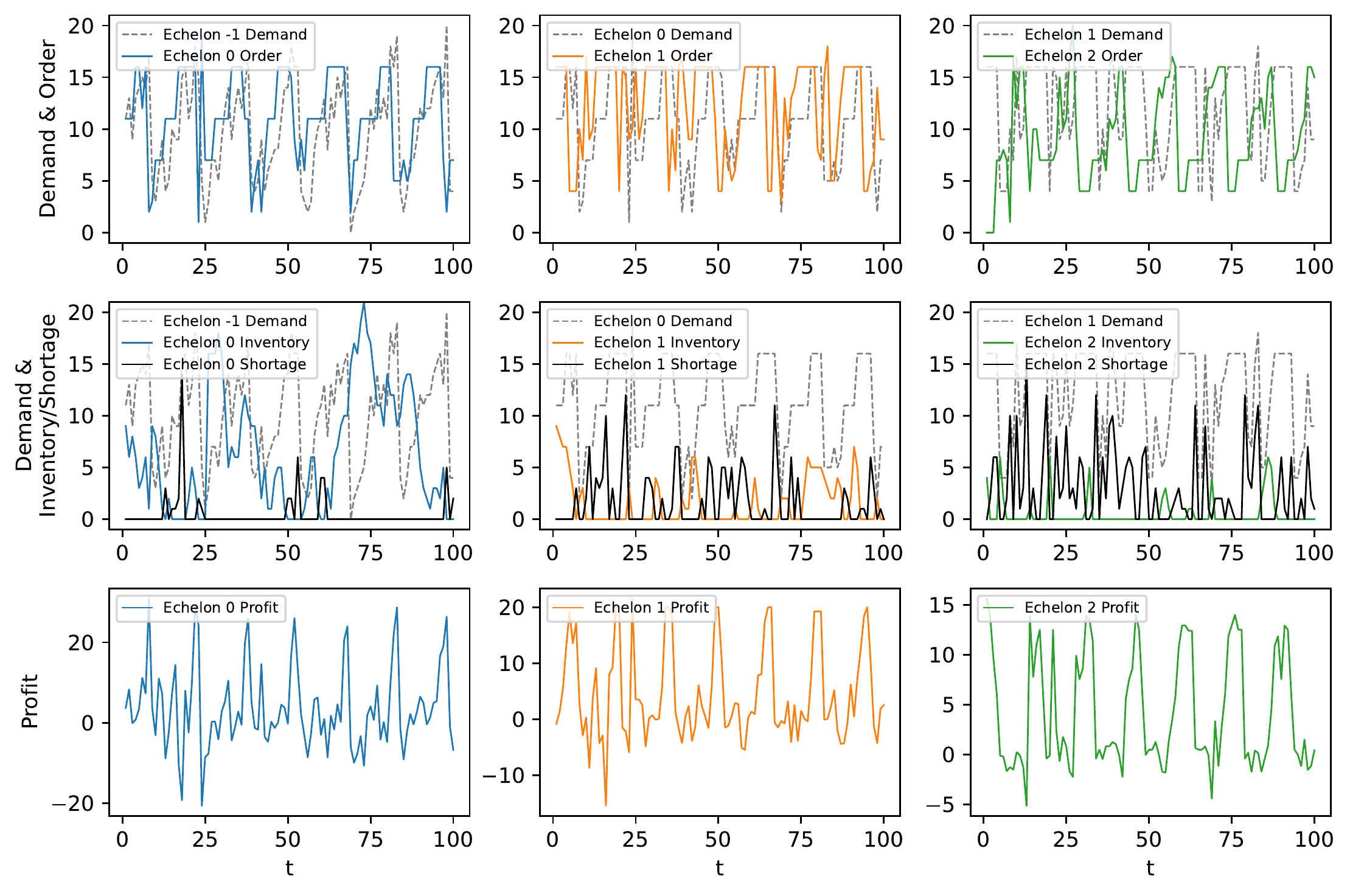}\\
        \end{minipage}
    }
    \subfigure[QPPO]{
        \begin{minipage}[t]{0.6\linewidth}
            \centering
            \includegraphics[width=\linewidth]{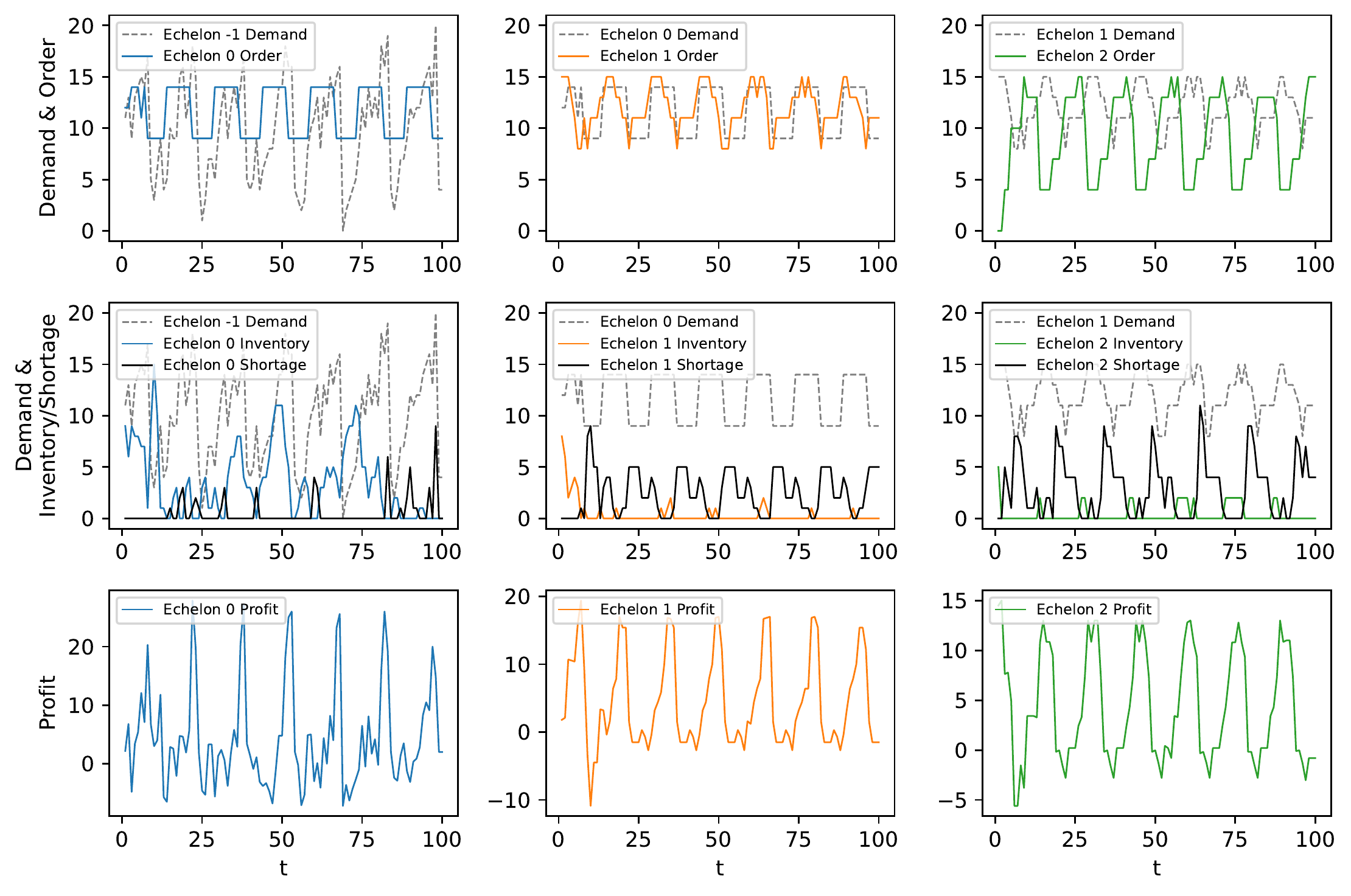}\\
        \end{minipage}
    }
    \caption{Policy visualization of PPO and QPPO in multi-echelon Inventory Management example.}
    \label{fig: ppo_qppo_multi_im}
\end{figure}

We further conduct experiments in a multi-echelon inventory management problem in which customer demands are generated by using a periodical stochastic model with a saw-wave trend. The policy and baseline are represented by neural networks consisting of two temporal convolutional layers and three parallel fully connected blocks to control each echelon. The agent makes reordering decisions in $100$ time steps and can look back at information within the past $5$ steps. The learning curves obtained through $5$ independent experiments and testing KDE plotted based on by $1000$ replications are shown in Figure \ref{fig: learning_curve_kde_multi_im} with the plotting settings the same as the single-echelon case. 
The policies generated by both algorithms are shown in Figure \ref{fig: ppo_qppo_multi_im}.
Although the ordering behavior of both agents show periodical patterns, QPPO's reordering quantities are almost the same in each period with smaller amplitude, whereas PPO leads to dramatic inventory adjustment more frequently.
We note that the performance of PPO degrades in the later stage. In addition, QPPO outperforms PPO under both criteria in this example, which suggests that it may be more desirable to solve multi-echelon supply chain inventory management problems from a quantile optimization perspective.

\section{Conclusions}
In this paper, we have proposed the QPO algorithm for RL with a quantile criterion under a policy optimization framework. 
We have shown that QPO converges to the global optimum under certain conditions and derived its rates of convergence. 
Then, we have introduced an off-policy version QPPO, which significantly improves the learning efficiency of agents. Convergence and new error bounds are established to provide insights into the theoretical performance of QPPO with new computational architecture.
Numerical experiments show that our proposed algorithms are effective for policy optimization under the quantile criterion. 
In finance, QPPO can learn how to hedge against risk based on the rewards obtained in the process of investment. 
In inventory management, QPPO can learn to control the tail risk and lead to more stable reordering decisions.
In the future work, the criterion can be extended to the distortion risk measures that are more general than quantiles \citep{glynn2021computing}.
The codes that support this study can be found in \url{https://github.com/JinyangJiangAI/Quantile-based-Policy-Optimization}.


\ACKNOWLEDGMENT{This work was supported in part by the National Natural Science Foundation of China (NSFC) under Grants 72250065, 72022001, and 71901003.}


\bibliographystyle{informs2014} 
\bibliography{mybib} 



%
%
%

\newpage
\begin{APPENDICES}
\section{Supplements for Section \ref{Section Quantile-Based Policy Optimization}}
\subsection{Proofs in “Strong Convergence of QPO"}\label{Appendix Proofs in "Strong Convergence of QPO"}
\proof{Proof of Lemma \ref{lem.ode1}.}
By definition $q(\alpha;\bar{\theta})=F_R^{-1}(\alpha;\bar{\theta})$, $q(\alpha;\bar{\theta})$ is the unique solution of $g_1(q,\bar{\theta})=0$, i.e., the unique equilibrium point of ODE (\ref{ca.eq4}).
Consider the Lyapunov function $V(x)=(x-q(\alpha;\bar{\theta}))^2$, and the derivative $\dot{V}(x)=2(x-q(\alpha;\bar{\theta}))(\alpha-F_R(x;\bar{\theta}))$ is negative for any $x\neq q(\alpha;\bar{\theta})$. Therefore, $q(\alpha;\bar{\theta})$ is global asymptotically stable by the Lyapunov Stability Theory \citep{liapounoff2016probleme}.
\Halmos\endproof

\proof{Proof of Lemma \ref{lem.ode2}.}
If $\theta^{*}\in\Theta^{\circ}$, then $\nabla_{\theta} q(\alpha;\theta^{*})=0$ and $C(\theta^{*})={0}$;  if $\theta^{*}\in\partial\Theta$, then $\nabla_{\theta} q(\alpha;\theta^{*})$ must lie in $C(\theta^{*})$, so $p(t)=-g_2(q(\alpha;\theta^{*}),\theta^{*})$. By convexity of $q(\alpha;\theta)$ on $\Theta$, $\theta^{*}$ is the unique equilibrium point. Take $V'(x)=\Vert x-\theta^{*}\Vert^2$ as the Lyapunov function, and the derivative is $\dot{V}'(x)=2(x-\theta^{*})'(g_2(q(\alpha;x),x)+p(t))$. Since $q(\alpha;\theta)$ is strictly convex, $(\theta^{*}-x)'\nabla q(\alpha;x)>q(\alpha;\theta^{*})-q(\alpha;x)>0$ for any $x\neq\theta^{*}$, which implies $(x-\theta^{*})'g_2(q(\alpha;x))<0$. Since $p(t)\in-C(x)$, we have $(x-\theta^{*})'p(t)\leq0$. Thus, $\theta^{*}$ is global asymptotically stable.
\Halmos\endproof

\proof{Proof of Lemma \ref{lem.q}.} Recursion (\ref{iter_q}) can be rewritten as
\begin{align}
    q_{k+1} = q_k +\beta_k(\alpha-F_R(q_k;\theta_k)) + \beta_k \delta_k, \label{ca.eq1}
\end{align}
where $\delta_k=F_R(q_k;\theta_k)-\mathbf{1}\{R_k\leq q_k\}$.
Let $M_k=\sum_{i=0}^k\beta_i\delta_i$. We then verify that $\{M_k\}$ is a $L^2$-bounded martingale sequence. With Assumption \ref{a3}(b) and boundedness of $\mathbf{1}\{\cdot\}$, we have
\begin{align*}
    \sum_{i=0}^k\beta_i^2\delta_i^2\leq 4 \sum_{i=0}^k\beta_i^2<\infty.
\end{align*}
By noticing $\mathbb{E}[\delta_i|\mathcal{F}_i]=0$, we have
\begin{align*}
    \mathbb{E}[\ \beta_i \delta_i\beta_j \delta_j\ ]=\mathbb{E}[\ \beta_i \delta_i\mathbb{E}[\ \beta_j \delta_j|\mathcal{F}_j\ ]\ ] = 0,
\end{align*}
for all $i<j$.
Thus, $\sup_{k\geq0}\mathbb{E}[M_k^2]<\infty$. From the martingale convergence theorem \citep{durrett2019probability}, we have $M_k\rightarrow M_{\infty}$ w.p.1. Then for any $\varepsilon>0$, there exists a constant $k_0>0$ such that for any $m>n \geq k_0$, we have
\begin{align}
    \big|\sum_{i=n+1}^{m}\beta_i\delta_i\big|<\varepsilon,\quad w.p.1. \label{ca.eq2}
\end{align}
Denote $q^{+}=\sup_{\theta\in\Theta}q(\alpha;\theta)$ and $q^{-}=\inf_{\theta\in\Theta}q(\alpha;\theta)$. Since Assumption \ref{a1} holds and $\Theta$ is compact, we have $|q^{+}|<\infty$ and $|q^{-}|<\infty$. From Assumption \ref{a3}(b), there exists $k_1>0$ such that for any $k \geq k_1$, $\beta_k<\varepsilon$.
We then establish an upper bound of $\{q_k\}$. If the tail sequence $\{q_k\}_{\max\{k_0,k_1\}}^{\infty}$ is bounded by $q^{+}$, the boundedness of $\{q_k\}$ holds. Otherwise, let $k_2\geq\max\{k_0,k_1\}$ be the first time that $\{q_k\}_{\max\{k_0,k_1\}}^{\infty}$ rises above $q^{+}$ (i.e. $q_{k_2}> q^{+}$ and $q_{k_2-1}\leq q^{+}$), and denote a segment above $q^{+}$ of $\{q_k\}_{k_2}^{\infty}$ as $\{q_k\}_{k'}^{k''}$.
We discuss all three possible situations as shown in Figure.\ref{fig:quantile_recursion}. On path 1, $\{q_k\}$ stays above $q^+$; on path 2, $\{q_k\}$ drops below $q^+$ first and then rises above $q^+$; and on path 3, once $\{q_k\}$ drops below $q^+$, it never rises above $q^+$.
\begin{figure}[t]
	\centering
	\includegraphics[scale=0.8]{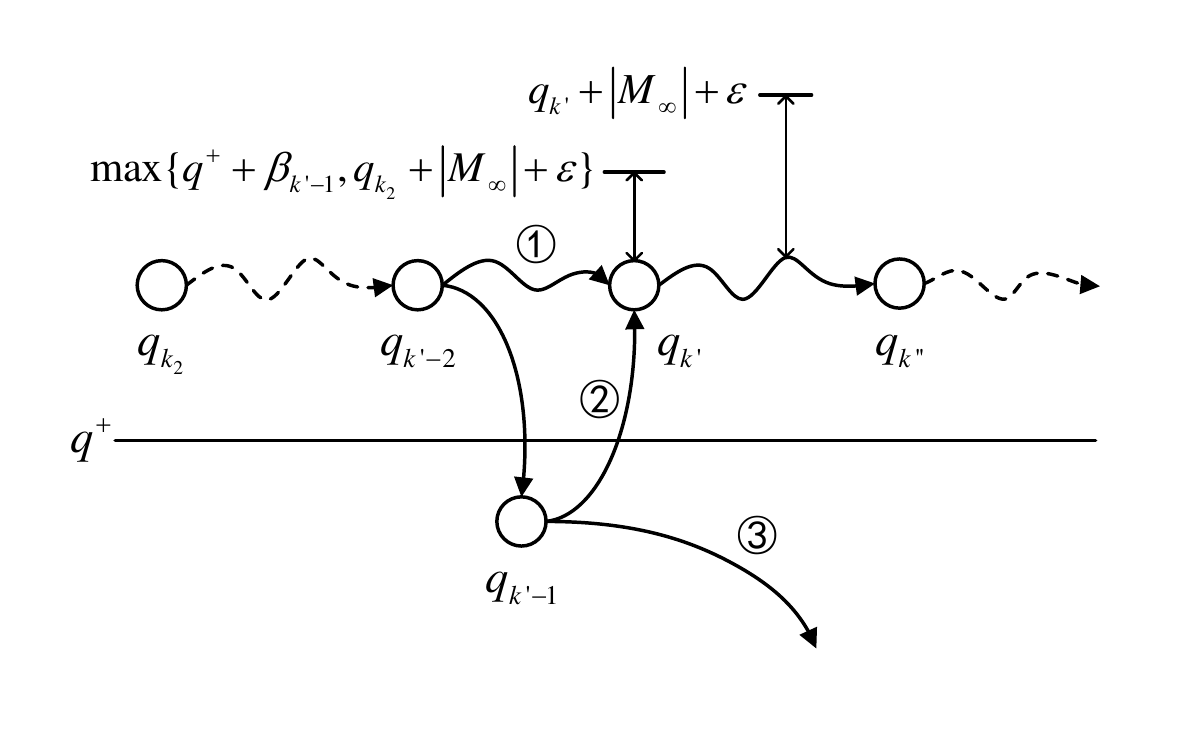}
	\caption{Illustration for possible situations of the path of $\{q_k\}$.}
	\label{fig:quantile_recursion}
\end{figure}
For the situations corresponding to path 1 and path 2, the definition of $\alpha$-quantile implies $\alpha-F_R(q_k;\theta_k)<0$ for $k\in[k',k'']$. Then with equality (\ref{ca.eq1}) and inequality (\ref{ca.eq2}), we have $q_k<q_{k'}+|M_{\infty}|+\varepsilon$ for $k\in[k',k'']$. In addition, if the situation corresponds to path 1, then $q_{k'}<q_{k_2}+|M_{\infty}|+\varepsilon$ by the previous result; and if it corresponds to path 2, there holds $q_{k'}<q^{+}+\beta_{k'-1}$ by recursion (\ref{iter_q}). By the definition of $k_1$, we have a bound for both cases:
$q_{k'}<\max\{q^{+}+\beta_{k'-1},q_{k_2}+|M_{\infty}|+\varepsilon\}=q_{k_2}+|M_{\infty}|+\varepsilon$
. Thus, we have $q_k<q_{k_2}+2|M_{\infty}|+2\varepsilon$ for all $k>k_2\geq\max\{k_0,k_1\}$. Analogously, we have $q_k>q_{k_2'}-2|M_{\infty}|-2\varepsilon$ for $k>k_2'\geq\max\{k_0,k_1\}$. For the situation corresponding to path 3, the tail sequence is naturally bounded.
In summary, the conclusion has been proved.
\Halmos\endproof

\proof{Proof of Theorem \ref{th.main}.}  With $F_R(r;\theta)\in C^1(\mathbb{R})$ and Assumption \ref{a2}, we have $g_1(q,\theta)$ and $g_2(q,\theta)$ are Lipschitz continuous. It has been verified in Lemma \ref{lem.q} that $M_k=\sum_{i=0}^k\beta_i\delta_i<\infty$. Denote $M'_k=\sum_{i=0}^k\gamma_i\delta'_i$, where $\delta'_i=D_i+\nabla_{\theta}F_R(q_i;\theta)|_{\theta=\theta_i}$ with $D_i=D(\tau_i;\theta_i,q_i)$. Since $\nabla_{\theta}F_R(q;\theta)$ is Lipschitz continuous on the compact set $\Theta$ and $\{q_k\}$ is bounded as shown in Lemma \ref{lem.q}, $\nabla_{\theta}F_R(q_i;\theta)|_{\theta=\theta_i}$ is bounded. By Assumption \ref{a4}, we have
\begin{align*}
    \Vert D_i\Vert&\leq\sum_{t=0}^{T-1}\Vert\nabla_{\theta}\log\pi(a_t^i|s_t^i;\theta_i)\Vert 
    \leq T \sup_{a,s,\theta}\Vert\nabla_{\theta}\log\pi(a|s;\theta)\Vert<\infty.
\end{align*}
By noticing $\mathbb{E}[\delta'_i|\mathcal{F}_i]=0$, Assumption \ref{a3}(a), and a similar argument in Lemma 3, we can prove $\{M'_k\}$ is a $L^2$-bounded martingale sequence, which implies that $\{M'_k\}$ is bounded w.p.1.

With Assumption \ref{a3} and the conclusions in Lemmas \ref{lem.ode1} and \ref{lem.q}, all conditions in Theorem \ref{th.bokar} are satisfied.
Therefore, it is almost sure that recursions (\ref{iter_q}) and (\ref{iter_theta}) converge to the unique global asymptotically stable equilibrium of ODE (\ref{ca.eq5}),
which is the optimal solution of problem (\ref{pf.eq3}) by the conclusion of Lemma \ref{lem.ode2}.
\Halmos\endproof

\subsection{Proofs in “Rate of Convergence"}\label{Appendix Proofs in "Rate of Convergence"}

\proof{Proof of Theorem \ref{clt}.}
Since $\theta^*\in\Theta$, we can omit the projection operator $\varphi(\cdot)$ in the recursion (\ref{iter_theta}).
The convergence of $\{q_k,\theta_k\}$ to $\{q(\alpha;\theta^*),\theta^*\}$ has already been proved in Theorem \ref{th.main}.
With Assumption \ref{clt.a1} and $Q_q:=Q_{11}-Q_{12}Q_{22}^{-1}Q_{21}=Q_{11}$, the largest eigenvalues of $Q_{22}$ and $Q_q$ are less than 0.
Since $\alpha-\mathbf{1}\{R_{k}\leq q_k\}$ and $D(\tau_i;\theta_i,q_i)$ are unbiased estimations of $g_1(q_k,\theta_k)$ and $g_2(q_k,\theta_k)$, $\mathbb{E}[\delta_k|\mathcal{F}_k] = \mathbb{E}[\delta'_k|\mathcal{F}_k] = 0$ a.s.
And we can find that
\begin{align*}
\lim_{k\rightarrow\infty}\ \mathbb{E}\left[
\begin{pmatrix} \delta_k  \\
\delta'_k\end{pmatrix}
\left(\delta_k^{\top}\ {\delta'}_k^{\top} \right)
\bigg|\mathcal{F}_k
\right] := \Gamma =\text{Cov}\left(
\begin{pmatrix}
\mathbf{1}\{R\leq q(\alpha;\theta^*)\}\\D(\tau;\theta^{*}, q(\alpha;\theta^*))
\end{pmatrix}\right),
\end{align*}
which must be a positive matrix.
By noticing $\delta_k$ and $\delta'_k$ are uniformly bounded, we immediately have the boundedness of all their finite moments. Therefore, all conditions in the central limit theorem of two-timescale stochastic approximation are satisfied, and the joint weak convergence rate of $\{q_k,\theta_k\}$ are as given in equation (\ref{clt_rate}).
\Halmos
\endproof

\proof{Proof of Theorem \ref{th.q_rate}.}
In Theorem \ref{th.main}, we have checked that $\Vert  D_k\Vert$ is bounded. Denote $C_D=T \sup_{a,s,\theta}\Vert\nabla_{\theta}\log\pi(a|s;\theta)\Vert$. By noticing Assumption \ref{a1} and compactness of $\Theta$, $q(\alpha;\theta)$ is Lipschitz continuous on $\Theta$ and denote its Lipschitz constant as $L_q$. Then we have
\begin{align*}
    \vert q(\alpha;\theta_k)-q(\alpha;\theta_{k+1})\vert
    &\leq L_q\Vert \theta_{k}-\theta_{k+1}\Vert 
    =  L_q\Vert \theta_{k}-\varphi(\theta_k+\gamma_k D_k)\Vert
    \leq 2L_q\gamma_k \Vert  D_k\Vert \leq 2L_q\gamma_k C_D,
\end{align*}
where the second inequality comes from the definition of the projection function $\varphi(\cdot)$. Define $\zeta_k = q_k-q(\alpha;\theta_k)$ and rewrite the recursion (\ref{iter_q}) as
\begin{align*}
    \zeta_{k+1}=\zeta_k+\beta_k(\alpha-\mathbf{1}\{U(\tau_{k})\leq q_{k})\})+q(\alpha;\theta_k)-q(\alpha;\theta_{k+1}).
\end{align*}
By taking square on both sides, we have
\begin{align*}
    \zeta_{k+1}^2\leq & \zeta_{k}^2+\beta_k^2+4L_q^2\gamma_k^2 C_D^2 + 4\vert \zeta_k\vert \gamma_k L_q C_D 
    +2\zeta_k\beta_k(\alpha-\mathbf{1}\{U(\tau_{k})\leq q_{k})\})\\
    & + 2\beta_k(\alpha-\mathbf{1}\{U(\tau_{k})\leq q_{k})\})(q(\alpha;\theta_k)-q(\alpha;\theta_{k+1})).
\end{align*}
With the definition of quantile and the intermediate value theorem, we can obtain
\begin{align*}
    \mathbb{E}[\zeta_{k+1}^2|\mathcal{F}_k]&\leq \zeta_{k}^2+\beta_k^2+4L_q^2\gamma_k^2 C_D^2 + 4\vert \zeta_k\vert \gamma_k L_q C_D \\&\quad
    -2\beta_k\zeta_k^2 f_R(\tilde{q_k};\theta_k)-2\beta_k\zeta_k f_R(\tilde{q_k};\theta_k) (q(\alpha;\theta_k)-q(\alpha;\theta_{k+1}))\\
    &\leq \zeta_{k}^2(1-2\beta_k C_f^{-}) + \beta_k^2+4L_q^2\gamma_k^2 C_D^2 + 4\vert \zeta_k\vert \gamma_k L_q C_D (1+\beta_k C_f^{+}),
\end{align*}
where $\tilde{q}_k$ lies in the interval between $q_k$ and $q(\alpha;\theta_k)$. By taking expectation with respect to $\mathcal{F}_k$ and applying the Cauchy Schwarz inequality, we further obtain
\begin{align*}
    \mathbb{E}[\zeta_{k+1}^2] & = \mathbb{E}[\mathbb{E}[\zeta_{k+1}^2|\mathcal{F}_k]]
    \leq \mathbb{E}[\zeta_k^2](1-2\beta_k C_f^{-}) + \beta_k^2 
    + 4L_q^2\gamma_k^2 C_D^2 + 4 \mathbb{E}[\vert \zeta_k\vert]\gamma_k L_q C_D(1+\beta_k C_f^{+})\\
    & \leq \mathbb{E}[\zeta_k^2](1-2\beta_k C_f^{-}) + \beta_k^2 + 4L_q^2\gamma_k^2 C_D^2 
    + 4 \gamma_k L_q C_D(1+\beta_k C_f^{+})\left(\mathbb{E}[\zeta_k^2]\right)^{\frac{1}{2}}\\
    & \leq \mathbb{E}[\zeta_k^2](1-2\beta_k C_f^{-}) + \beta_k^2 + 4L_q^2\gamma_k^2 C_D^2 
    + \frac{(2 \gamma_k L_q C_D(1+\beta_k C_f^{+}) )^2}{\beta_kC_f^{-}}+\beta_kC_f^{-}\mathbb{E}[\zeta_k^2]\\
    & = \mathbb{E}[\zeta_k^2](1-\beta_k C_f^{-}) + \beta_k^2 + 4\frac{\gamma_k^2}{\beta_k}L_q^2 C_D^2\left(\beta_k+\frac{(1+\beta_kC_f^{+})^2}{C_f^{-}}\right)\\
    &\leq \mathbb{E}[\zeta_k^2](1-\beta_k C_f^{-}) + \beta_k^2 + \frac{\gamma_k^2}{\beta_k}\tilde{C},
\end{align*}
where $\tilde{C} = 4L_q^2 C_D^2\left(\beta_1+\frac{(1+\beta_1C_f^{+})^2}{C_f^{-}}\right)$. By assuming $\beta_kC_f^{-}<1$ without lossing generality and repeatedly applying the inequality above, we have
\begin{align}
    & \mathbb{E}[\zeta_{k+1}^2] \leq \prod_{i=1}^k (1-\beta_iC_f^{-})\mathbb{E}[\zeta_1^2]
    +\sum_{i=1}^{k-1}\left[\prod_{j=i+1}^k (1-\beta_jC_f^{-}) \right]\beta_i(\beta_i+\frac{\gamma_i^2}{\beta_i^2}\tilde{C}) + (\beta_{k+1}^2+\frac{\gamma_{k+1}^2}{\beta_{k+1}}\tilde{C})\nonumber\\
    &= \prod_{i=1}^k (1-\beta_iC_f^{-})(\mathbb{E}[\zeta_1^2]+\frac{\beta_{1}^2+\gamma_{1}^2\beta_{1}^{-1}\tilde{C}}{1-\beta_1C_f^{-}}) +\sum_{i=2}^{k-1}\left[\prod_{j=i+1}^k (1-\beta_jC_f^{-}) \right]\beta_i(\beta_i+\frac{\gamma_i^2}{\beta_i^2}\tilde{C})
    + (\beta_{k}^2+\frac{\gamma_{k}^2}{\beta_{k}}\tilde{C}). \label{order1}
\end{align}
We then bound the order of the second term on the right hand side of inequality (\ref{order1}). For any $p>0$,
\begin{align*}
    \sum_{i=2}^{k-1}&\left[\prod_{j=i+1}^k (1-C_f^{-}j^{-\beta}) \right]i^{-\beta}i^{-p} \leq \sum_{i=2}^{k-1}\exp\{-C_f^{-}\sum_{j=i+1}^k j^{-\beta}\}i^{-(\beta+p)}\\
    &\leq \sum_{i=2}^{k-1}\exp\{-C_f^{-}\int_{i+1}^{{\color{black}k+1}} x^{-\beta}dx\}i^{-(\beta+p)}
    = \sum_{i=2}^{k-1}\exp\{c((i+1)^{1-\beta}-{\color{black}(k+1)}^{1-\beta})\}i^{-(\beta+p)}\\
    & \leq {\color{black}c'}\int_2^{{\color{black}k}} \exp\{c((x+1)^{1-\beta}-{\color{black}(k+1)}^{1-\beta})\}x^{-(\beta+p)}dx,
\end{align*}
where $c=C_f^{-}(1-\beta)^{-1}$ {\color{black}and $c'$ is a sufficiently large constant}, and the last inequality holds because $\exp\{c((x+1)^{1-\beta}-k^{1-\beta})\}$ is monotonically increasing {\color{black}when $x\gg0$}.
Let
\begin{align*}
    f_k(x) &= \exp\{c((x+1)^{1-\beta}-{\color{black}(k+1)}^{1-\beta})\}(x-1)^{-(\beta+p)},\\
    g_k(x) &= \exp\{c((x+1)^{1-\beta}-{\color{black}(k+1)}^{1-\beta})\}\frac{(x-1)^{-(\beta+p+1)}(1+x)^{-\beta}-(x-1)^{-(\beta+p)}(1+x)^{-(1+\beta)}}{c(x+1)^{-2\beta}}.
\end{align*}
Then by noticing that $x^{-(\beta+p)}\geq0$ is monotonically decreasing and $\exp\{c((x+1)^{1-\beta}-k^{1-\beta})\}\geq0$, we have
\begin{align}
    {\color{black}\frac{1}{c'}}\sum_{i=2}^{k-1} \left[\prod_{j=i+1}^k (1-C_f^{-}j^{-\beta}) \right] i^{-\beta}i^{-p} \leq \int_2^{{\color{black}k}} f_k(x)dx = \frac{f_k(x)}{c(1+x)^{-\beta}}\bigg|_2^{{\color{black}k}}-\int_2^{{\color{black}k}} g_k(x)dx, \label{ineq1}
\end{align}
where the equality comes from integration by parts.
The order of the first term on the right hand side of inequality (\ref{ineq1}) is
\begin{align}
    \frac{f_k(x)}{c(1+x)^{-\beta}}\bigg|_2^{{\color{black}k}}= \frac{({\color{black}k-1})^{-(\beta+p)}}{c {\color{black}(k+1)}^{-\beta}}-\frac{\exp\{c(3^{1-\beta}-{\color{black}(k+1)}^{1-\beta})\}}{c 3^{-\beta}}
    \leq\frac{({\color{black}k-1})^{-(\beta+p)}}{c {\color{black}(k+1)}^{-\beta}} = O(k^{-p}). \label{ineq2}
\end{align}
For $x$ large enough, there exist constant $c''>0$ such that $g'(x)>0$ and $g(x)\leq c''x^{\beta-1}f(x)$. Then for large enough $k$,
\begin{align*}
    \frac{\int_2^{{\color{black}k}} g_k(x)dx}{\int_2^{{\color{black}k}} f_k(x)dx}
    \leq \frac{2\int_{\frac{{\color{black}k}+2}{2}}^{{\color{black}k}} g_k(x)dx}{\int_{\frac{{\color{black}k}+2}{2}}^{{\color{black}k}} f_k(x)dx}
    \leq \frac{2c''\int_{\frac{k+{\color{black}2}}{2}}^{{\color{black}k}} f_k(x)x^{\beta-1}dx}{\int_{\frac{k+{\color{black}2}}{2}}^{{\color{black}k}} f_k(x)dx}
    \leq 2c''\left(\frac{k+{\color{black}2}}{2}\right)^{\beta-1},
\end{align*}
which goes to zero as $k\rightarrow0$. Combining this with inequality (\ref{ineq1}) and (\ref{ineq2}), we have
\begin{align}
    \sum_{i=2}^{k-1} \left[\prod_{j=i+1}^k (1-cj^{-\beta}) \right] i^{-\beta}i^{-p}= O(k^{-p}).\label{order_inequality}
\end{align}
Therefore, the second term on the right hand side of inequality (\ref{order1}) is in the order of $O(k^{-\beta})+O(k^{2\beta-2\gamma})=O(\beta_k)+O(\gamma_k^2\beta_k^{-2})$. Note that
$\mathbb{E}[\zeta_1^2]\leq\mathbb{E}[\zeta_0^2](1-\beta_0C_f^{-})+\beta_0^2+4\frac{\gamma_0^2}{\beta_0}L_q^2 C_D^2\left(\beta_0+\frac{(1+\beta_0C_f^{+})^2}{C_f^{-}}\right)$ and
\begin{align*}
    \prod_{i=0}^k (1-\beta_iC_f^{-})\leq \exp\{-C_f^{-}\sum_{i=0}^k\beta_i\}\leq \exp\{-C_f^{-}bk^{1-\beta}\},
\end{align*}
so the first term on the right hand side of inequality (\ref{order1}) decays exponentially so that it can be absorbed into the second term.
Finally, we can conclude that $\mathbb{E}[\zeta_k^2] = O(\beta_k)+O(\gamma_k^2\beta_k^{-2})$, which completes the proof.\Halmos
\endproof

\proof{Proof of Theorem \ref{th.theta_rate}.}
Rewrite the recursion (\ref{iter_theta}) as
\begin{align}
    \theta_{k+1}=\theta_k+\gamma_k D_k+\gamma_k P_k, \label{iter_theta_proj}
\end{align}
where $P_k\in -C(\theta_{k+1})$ is the vector with the shortest $L^2$ norm needed to project $\theta_k+\gamma_k D_k$ onto $\Theta$.
Let $\Delta_k = \theta_k-\theta^*$. From the equality (\ref{iter_theta_proj}) and Assumption \ref{a2}, we have
\begin{align*}
    \Vert \Delta_{k+1} \Vert^2 =&\Vert \Delta_k \Vert^2 + 2\gamma_k\Delta_k^{\top}(D_k+P_k)+\gamma_k^2\Vert D_k+P_k \Vert^2\\
     \leq& \Vert \Delta_k \Vert^2 - 2\gamma_k\Delta_k^{\top}\nabla_{\theta} F_R(q_k;\theta_k) + 2\gamma_k\Delta_k^{\top}\delta'_k
    +2\gamma_k\Delta_k^{\top}P_k+4\gamma_k^2C_D^2\\
    \leq & \Vert \Delta_k \Vert^2 - 2\gamma_k\Delta_k^{\top}\nabla_{\theta} F_R(r;\theta_k)\big|_{r=q(\alpha;\theta_k)} + 2\gamma_k\Delta_k^{\top}\delta'_k
    +2\gamma_k\Delta_k^{\top}P_k+4\gamma_k^2C_D^2
    + 2\gamma_kC \Vert\Delta_k\Vert \Vert q(\alpha;\theta_k) - q_k\Vert .
\end{align*}
By noticing the strict convexity of $q(\alpha;\theta)$, we have $\Delta_k^{\top}\nabla_{\theta}q(\alpha;\theta_k){\color{black}<0}$, which implies $\Delta_k^{\top}(-\nabla_{\theta} F_R(r;\theta_k)\big|_{r=q(\alpha;\theta_k)})/{\color{black}C_f^{-}}\leq \Delta_k^{\top}\nabla_{\theta} q(\alpha;\theta_k)$ from Assumption \ref{a5} and the equality (\ref{pf.eq4}).
Then, we can obtain
\begin{align*}
    \Vert \Delta_{k+1} \Vert^2
    & \leq \Vert \Delta_k \Vert^2 {\color{black}+} 2\gamma_k{\color{black}C_f^{-}}\Delta_k^{\top}\nabla_{\theta}q(\alpha;\theta_k) + 2\gamma_k\Delta_k^{\top}\delta'_k
    +2\gamma_k\Delta_k^{\top}P_k+4\gamma_k^2C_D^2
    + 2\gamma_kC \Vert\Delta_k\Vert \Vert q(\alpha;\theta_k) - q_k\Vert \\
    & \leq \Vert \Delta_k \Vert^2 {\color{black}+} 2\gamma_k{\color{black}C_f^{-}}\Delta_k^{\top}H(\tilde{\theta}_k)\Delta_k + 2\gamma_k\Delta_k^{\top}\delta'_k
    +2\gamma_k\vert\Delta_k^{\top}P_k\vert+4\gamma_k^2C_D^2
    + 2\gamma_kC \Vert\Delta_k\Vert \Vert q(\alpha;\theta_k) - q_k\Vert ,
\end{align*}
where $\tilde{\theta}$ lies between $\theta_k$ and $\theta^*$; the second equality comes from the Taylor expansion of $\nabla_{\theta}q(\alpha;\theta_k)$ around $\theta^*$. Using the Rayleigh-Ritz inequality \citep{rugh1996linear} and Assumption \ref{a6}, we have
\begin{align*}
    \Vert \Delta_{k+1} \Vert^2
    & \leq (1-2\gamma_k{\color{black}C_f^{-}}C_{\lambda})\Vert \Delta_k \Vert^2 +2\gamma_k\Delta_k^{\top}\delta'_k
    +2\gamma_k\vert\Delta_k^{\top}P_k\vert+4\gamma_k^2C_D^2
    + 2\gamma_kC \Vert\Delta_k\Vert \Vert q(\alpha;\theta_k) - q_k\Vert .
\end{align*}
Note that $\mathbb{E}[\Delta_k^{\top}\delta'_k|\mathcal{F}_k]=\Delta_k^{\top}\mathbb{E}[\delta'_k|\mathcal{F}_k]=\Delta_k^{\top}\mathbf{0}=0$.
And by Theorem \ref{th.q_rate}, we have $\mathbb{E}[\Vert q(\alpha;\theta_k) - q_k\Vert^2]\leq C_{\beta}(\frac{\gamma_k^2}{\beta_k^2}+\beta_k)$.
Taking expectation on both sides and applying the Cauchy Schwarz inequality, we have
\begin{align}
    \mathbb{E}[\Vert \Delta_{k+1} \Vert^2]\leq& (1-2\gamma_k{\color{black}C_f^{-}}C_{\lambda})\mathbb{E}[\Vert \Delta_k \Vert^2] + 2\gamma_k \mathbb{E}[\mathbb{E}[\Delta_k^{\top}\delta'_k)|\mathcal{F}_k]] \nonumber
    + 2\gamma_k\left( \mathbb{E}[\Vert \Delta_k \Vert^2] \mathbb{E}[\Vert P_k \Vert^2]\right)^{\frac{1}{2}} + 4\gamma_k^2C_D^2 \\
    &+ 2\gamma_kC \left(\mathbb{E}[\Vert\Delta_k\Vert^2] \mathbb{E}[\Vert q(\alpha;\theta_k) - q_k\Vert^2]  \right)^{\frac{1}{2}} \nonumber\\
     \leq& (1-2\gamma_k{\color{black}C_f^{-}}C_{\lambda})\mathbb{E}[\Vert \Delta_k \Vert^2] 
    + 2\gamma_k\left( \mathbb{E}[\Vert \Delta_k \Vert^2] \mathbb{E}[\Vert P_k \Vert^2]\right)^{\frac{1}{2}} + 4\gamma_k^2C_D^2\nonumber\\
    &+ 2\gamma_kC \left(\mathbb{E}[\Vert\Delta_k\Vert^2] \right)^{\frac{1}{2}} \left(C_{\beta}(\gamma_k^2\beta_k^{-2}+\beta_k) \right)^{\frac{1}{2}}. \label{ordertheta1}
\end{align}
Now we derive a bound for $\mathbb{E}[\Vert P_k \Vert^2]$. Assume the interior of $\Theta$ is not empty, so that is a constant $C_{\Theta}>0$ such that $\mathcal{B}(\theta^*, 2C_{\Theta})\subseteq\Theta$, where $\mathcal{B}(\theta^*,\delta)$ represents a round neighborhood of $\theta^*$ with radius $\delta$. Let $\mathcal{E}_k=\{\theta_{k}+\gamma_kD_k\notin \mathcal{B}(\theta^*, 2C_{\Theta})\}$. By the definition of the projection function $\varphi(\cdot)$, the occurrence of $\mathcal{E}_k^c$ implies $P_k=0$. Thus, we have
\begin{align*}
    \mathbb{E} [\Vert P_k \Vert^2]& =\mathbb{E}[\Vert P_k \Vert^2|\mathcal{E}_k]P(\mathcal{E}_k)
    \leq \mathbb{E}[\Vert D_k \Vert^2]P(\mathcal{E}_k)  \\
    & \leq \mathbb{E}[\Vert D_k \Vert^2](P(\Vert (\theta_{k}+\gamma_kD_k)-\theta_k \Vert\geq C_{\Theta})+P(\Vert \theta_{k}-\theta^* \Vert\geq C_{\Theta})).
\end{align*}
Using the Markov's inequality, we further have
\begin{align}
    \mathbb{E} [\Vert P_k \Vert^2] &\leq \mathbb{E}[\Vert D_k \Vert^2]\frac{\gamma_k^2\mathbb{E}[\Vert D_k \Vert^2]+\mathbb{E}[\Vert \Delta_k \Vert^2]}{C_{\Theta}^2} \nonumber\\
    & \leq \frac{\gamma_k^2\left(\mathbb{E}[\Vert D_k \Vert^2]\right)^2}{C_{\Theta}^2}
    +\frac{\mathbb{E}[\Vert D_k \Vert^2]\mathbb{E}[\Vert \Delta_k \Vert^2]}{C_{\Theta}^2} 
    \leq \frac{\gamma_k^2C_D^4}{C_{\Theta}^2}
    +\frac{C_D^2\mathbb{E}[\Vert \Delta_k \Vert^2]}{C_{\Theta}^2}. \label{ordertheta2}
\end{align}
Next, we apply inequality (\ref{ordertheta2}) to inequality (\ref{ordertheta1}) and obtain:
\begin{align*}
    \mathbb{E}[\Vert \Delta_{k+1} \Vert^2] \leq& \left(1-2\gamma_k\left({\color{black}C_f^{-}}C_{\lambda}-\frac{C_D^2}{C_{\Theta}^2}\right)\right)\mathbb{E}[\Vert \Delta_k \Vert^2] 
    + 2\gamma_k \left( \mathbb{E}[\Vert \Delta_k \Vert^2]\right)^{\frac{1}{2}} \frac{\gamma_kC_D^2}{C_{\Theta}} + 4\gamma_k^2C_D^2\\
    &+ 2\gamma_kC \left(\mathbb{E}[\Vert\Delta_k\Vert^2] \right)^{\frac{1}{2}} \left(C_{\beta}(\gamma_k^2\beta_k^{-2}+\beta_k) \right)^{\frac{1}{2}}\\
    \leq &\left(1-2\gamma_k\left({\color{black}C_f^{-}}C_{\lambda}-\frac{C_D^2}{C_{\Theta}^2}\right)\right)\mathbb{E}[\Vert \Delta_k \Vert^2] 
    + \gamma_k \frac{C_D^2}{C_{\Theta}^2}\mathbb{E}[\Vert \Delta_{k} \Vert^2] + \gamma_k^3C_D^2 + 4\gamma_k^2C_D^2\\
    &+\gamma_k\mathbb{E}[\Vert \Delta_{k} \Vert^2]+\gamma_kC^2C_{\beta}(\gamma_k^2\beta_k^{-2}+\beta_k) \\
     =& \left(1-\gamma_k\tilde{C}'\right)\mathbb{E}[\Vert \Delta_k \Vert^2] + \gamma_k^2C_D^2(\gamma_k+4)+\gamma_kC^2C_{\beta}(\gamma_k^2\beta_k^{-2}+\beta_k),
\end{align*}
where $\tilde{C}'=2{\color{black}C_f^{-}}C_{\lambda}-3\frac{C_D^2}{C_{\Theta}^2}-1$. Let $\tilde{C}''=C_D^2(\gamma_1+4)$. From the inequality above, we have
\begin{align}
    \mathbb{E}[\Vert \Delta_{k+1} \Vert^2] \leq& \left(1-\gamma_k\tilde{C}'\right)\mathbb{E}[\Vert \Delta_k \Vert^2] + \gamma_k(\tilde{C}''\gamma_k+C^2C_{\beta}(\gamma_k^2\beta_k^{-2}+\beta_k)) \nonumber\\
    \leq& \prod_{i=1}^k (1-\gamma_i\tilde{C}')\mathbb{E}[\Vert \Delta_1\Vert ^2] +\sum_{i=1}^{k-1}\left[\prod_{j=i+1}^k (1-\gamma_j\tilde{C}') \right]\gamma_i(\tilde{C}''\gamma_i+C^2C_{\beta}(\gamma_i^2\beta_i^{-2}+\beta_i)) \nonumber\\
    &+\gamma_k(\tilde{C}''\gamma_k+C^2C_{\beta}(\gamma_k^2\beta_k^{-2}+\beta_k)) \nonumber\\
    =& \prod_{i=1}^k (1-\gamma_i\tilde{C}')(\mathbb{E}[\Vert \Delta_1\Vert ^2]+\frac{\gamma_1(\tilde{C}''\gamma_1+C^2C_{\beta}(\gamma_1^2\beta_1^{-2}+\beta_1))}{1-\gamma_1\tilde{C}'}) \nonumber\\
    &+\sum_{i=2}^{k-1}\left[\prod_{j=i+1}^k (1-\gamma_j\tilde{C}') \right]\gamma_i(\tilde{C}''\gamma_i+C^2C_{\beta}(\gamma_i^2\beta_i^{-2}+\beta_i))+\gamma_k(\tilde{C}''\gamma_k+C^2C_{\beta}(\gamma_k^2\beta_k^{-2}+\beta_k)). \label{order2}
\end{align}
By an analogous analysis in Theorem \ref{th.q_rate}, the second term on the right hand side of inequality (\ref{order2}) can be proved to be in the order of $O(\gamma_k)+O(\frac{\gamma_k^2}{\beta_k^2})+O(\beta_k)$.
Similarly, the first term decays exponentially so that it can be absorbed into the second term. Therefore, the conclusions of the theorem holds.
\Halmos
\endproof

\section{Supplements for Section \ref{Section Acceleration Technique}}\label{Appendix Acceleration Technique}

\proof{Proof of Lemma \ref{lem.ode3}.}
See the proof of Lemma \ref{lem.ode1} and \ref{lem.ode2}.
\Halmos\endproof

\proof{Proof of Lemma \ref{lem.aq}.}
Since Lemma \ref{lem.q} is derived based on Assumption \ref{a3}(b), the  boundedness of the distribution function and $C^1$-continuity of the quantile function on a compact parameter set,
the sequence $\{q_k^l\}_{l=T_0}^T$ can be proved bounded w.p.1 in the same manner under Assumptions \ref{a3}(b) and \ref{a7}.
\Halmos\endproof

\proof{Proof of Lemma \ref{lem.aqpo}.}
$M_k''$ can be rewritten as $M_k'' =M_{k1}''+M_{k2}''$, where
\begin{align}
    M_{k1}''& = \sum_{i=0}^k\gamma_i(D(\tau_i^{l_i};\theta_i,q_i^{l_i})+\nabla_{\theta'}F_{R^{l_i}}(q_i^{l_i};\theta')\big|_{\theta'=\theta_i}) \label{bias1}\\
    M_{k2}''& = \sum_{i=0}^k \gamma_i\left(\frac{1}{T-T_0+1}\sum_{l=T_0}^T \nabla_{\theta'}F_{R^l}(q_i^l;\theta')\big|_{\theta'=\theta_i}-\nabla_{\theta'}F_{R^{l_i}}(q_i^{l_i};\theta')\big|_{\theta'=\theta_i}\right) \label{bias2}
\end{align}
Similar to the proof of Theorem \ref{th.main}, since $D(\tau_i^{l_i};\theta_i,q_i^{l_i})$ and $\nabla_{\theta'}F_{R^{l_i}}(q_i^{l_i};\theta')\big|_{\theta'=\theta_i}$ are bounded and $\mathbb{E}[D(\tau_i^{l_i};\theta_i,q_i^{l_i})|\mathcal{F}_i]=-\nabla_{\theta'}F_{R^{l_i}}(q_i^{l_i};\theta')\big|_{\theta'=\theta_i}$, $\{M_{k1}''\}$ is an $L^2$-bounded martingale sequence and bounded w.p.1. Considering that $\mathbb{E}[\nabla_{\theta'}F_{R^{l_i}}(q_i^{l_i};\theta')\big|_{\theta'=\theta_i}|\mathcal{F}_i]=\frac{1}{T-T_0+1}\sum_{l=T_0}^T \nabla_{\theta'}F_{R^l}(q_i^l;\theta')\big|_{\theta'=\theta_i}$, we can conclude that $\{M_{k2}''\}$ is also an $L^2$-bounded martingale sequence and almost surely bounded. Further, the sequence $\{M_{k}''\}$ is bounded w.p.1.
\Halmos\endproof

\proof{Proof of Theorem \ref{main.aqpo}.}  With $F_R^l(r;\theta)\in C^1(\mathbb{R})$ and Assumption \ref{a8}, $g_1^l(q,\theta)$ and $g_2^l(q,\theta)$ are Lipschitz continuous. Let $M_{kl}'''=\sum_{i=0}^k \beta_{i}\mathbf{1}\{l=l_i\}(F_{R^{l_i}}(q_{i}^{l_i};\theta_{i})-\mathbf{1}\{U(\tau^{l_i}_{i})\leq q_{i}^{l_i}\})$, for $l=T_0,\cdots,T$. It can be verified by Assumption \ref{a3}(b) and the martingale convergence theorem that $\{M_{kl}'''\}$ is a $L^2$-bounded martingale sequence and bounded w.p.1.
With Assumption \ref{a3} and the conclusions in Lemmas \ref{lem.ode3}, \ref{lem.aq} and \ref{lem.aqpo}, all conditions in Theorem \ref{th.bokar} are satisfied.
Therefore, it is almost sure that recursions (\ref{iter_q_trun}) and (\ref{iter_theta_trun}) converge to the unique global asymptotically stable equilibrium of ODE (\ref{thetaode.aqpo}),
which is the optimal solution of problem (\ref{prob.aqpo}) by the conclusion of Lemma \ref{lem.ode3}.
\Halmos\endproof

\proof{Proof of Theorem \ref{q_error_1}.}
Using the definition of $\theta^*$ and $\theta^{**}$, we have $q^T(\alpha;\theta^*)\geq q^T(\alpha;\theta^{**})$ and $\bar{q}(\alpha;\theta^{**})\geq \bar{q}(\alpha;\theta^*)$.
Note that for any $\tau^T$ and its subsequence $\tau^l\subseteq\tau^T$, $l\leq T$, we have
\begin{align}
    \sup_{\theta\in\Theta}|q^l(\alpha;\theta)-q^T(\alpha;\theta)|
    &= \sup_{\theta\in\Theta} |F_{R^l}^{-1}(\alpha;\theta)-F_{R^T}^{-1}(\alpha;\theta)|
    \leq \sup_{u\in(0,1),\theta\in\Theta} |F_{R^l}^{-1}(u;\theta)-F_{R^T}^{-1}(u;\theta)|\nonumber\\
    &=\sup_{\tau^T} |U(\tau^T)-U(\tau^l)|
    \leq\sum_{t=l}^{T-1}\eta^t \sup_{(s',a,s)}|u(s',a,s)|\leq \frac{\eta^l}{1-\eta}C_r.\label{q_diff}
\end{align}
Then, using the inequality (\ref{q_diff}), we have
\begin{align*}
    &\sum_{l=T_0}^T q^T(\alpha;\theta^*)\geq  \sum_{l=T_0}^T q^T(\alpha;\theta^{**})\geq \sum_{l=T_0}^T \left(q^l(\alpha;\theta^{**}) - \frac{\eta^l}{1-\eta}C_r\right) \geq \bar{q}(\alpha;\theta^{**})- \frac{\eta^{T_0}}{(1-\eta)^2}C_r,\\
    &\bar{q}(\alpha;\theta^{**}) + \frac{\eta^{T_0}}{(1-\eta)^2}C_r \geq
    \bar{q}(\alpha;\theta^*) + \frac{\eta^{T_0}}{(1-\eta)^2}C_r \geq \sum_{l=T_0}^T \left(q^l(\alpha;\theta^{*}) + \frac{\eta^l}{1-\eta}C_r\right) \geq \sum_{l=T_0}^T q^T(\alpha;\theta^*),
\end{align*}
which imply $|q^T(\alpha;\theta^*)-\frac{1}{T-T_0+1}\bar{q}(\alpha;\theta^{**})|\leq \frac{1}{T-T_0+1}\cdot\frac{\eta^{T_0}}{(1-\eta)^2}C_r$. Further, we can obtain
\begin{align*}
    |q^T(\alpha;\theta^*)&-q^T(\alpha;\theta^{**})|
    \leq |q^T(\alpha;\theta^*)-\frac{1}{T-T_0+1}\bar{q}(\alpha;\theta^{**})|+ \frac{1}{T-T_0+1}\sum_{l=T_0}^T |q^l(\alpha;\theta^{**}) - q^T(\alpha;\theta^{**})|\\
    &\leq \frac{1}{T-T_0+1}\cdot\frac{\eta^{T_0}}{(1-\eta)^2}C_r + \frac{1}{T-T_0+1}\sum_{l=T_0}^T \frac{\eta^l}{1-\eta}C_r
    \leq \frac{2}{T-T_0+1}\cdot\frac{\eta^{T_0}}{(1-\eta)^2}C_r.
\end{align*}
\begin{flushright}
\Halmos
\end{flushright}
\endproof

\proof{Proof of Theorem \ref{q_error_2}.}
Let $\mu_t:=\mu(s_{t+1},a_t,s_t)$, $\sigma^2_t:=\sigma^2(s_{t+1},a_t,s_t)$. Since $u_t:=u(s_{t+1},a_t,s_t)\sim\mathcal{N}(\mu_t,\sigma^2_t)$, for any subsequence $\tau^l$ of $\tau^T$, $l\leq T$, we have $$U(\tau^l)=\sum_{t=0}^{l-1}u_t\sim\mathcal{N}(\sum_{t=0}^{l-1}\eta^t\mu_t,\sum_{t=0}^{l-1}\eta^{2t}\sigma^2_t)=\sqrt{\sum_{t=0}^{l-1}\eta^{2t}\sigma^2_t}\mathcal{N}(0,1)+\sum_{t=0}^{l-1}\eta^t\mu_t.$$
which implies $q^l(\alpha;\theta)=\sqrt{\sum_{t=0}^{l-1}\eta^{2t}\sigma^2_t}q_{\mathcal{N}(0,1)}(\alpha)+\sum_{t=0}^{l-1}\eta^t\mu_t$. Thus, we can obtain
\begin{align*}
    \sup_{\theta\in\Theta}|q^l(\alpha;\theta)-q^T(\alpha;\theta)|
    &= \sup_{\tau^T}\left|\left(\sqrt{\sum_{t=0}^{T-1}\eta^{2t}\sigma^2_t}-\sqrt{\sum_{t=0}^{l-1}\eta^{2t}\sigma^2_t}\right)q_{\mathcal{N}(0,1)}(\alpha) + \sum_{t=l}^{T-1}\eta^t\mu_t\right|\nonumber\\
    &\leq C_{\sigma}\sqrt{\sum_{t=l}^{T-1}\eta^{2t}}|q_{\mathcal{N}(0,1)}(\alpha)| + C_{\mu}\sum_{t=l}^{T-1}\eta^t\nonumber\\
    &\leq C_{\sigma}\sqrt{\frac{\eta^{2l}}{1-\eta^2}}|q_{\mathcal{N}(0,1)}(\alpha)| + C_{\mu}\frac{\eta^{l}}{1-\eta}\leq C_r' \frac{\eta^{l}}{1-\eta},
\end{align*}
where $C_r'=C_{\sigma}|q_{\mathcal{N}(0,1)}(\alpha)|+C_{\mu}$ with $C_{\mu},C_{\sigma}>0$ being upper bounds of $|\mu_t|$ and $\sigma_t$ on $\mathcal{S}\times\mathcal{A}\times\mathcal{S}$ respectively,
which has the same form as the inequality (\ref{q_diff}). Therefore, in the same manner of Theorem \ref{q_error_1}, we have $|q^T(\alpha;\theta^*)-\frac{1}{T-T_0+1}\bar{q}(\alpha;\theta^{**})|\leq \frac{1}{T-T_0+1}\cdot\frac{\eta^{T_0}}{(1-\eta)^2}C_r'$ and  $|q^T(\alpha;\theta^*)-q^T(\alpha;\theta^{**})|\leq \frac{2}{T-T_0+1}\cdot\frac{\eta^{T_0}}{(1-\eta)^2}C_r'$.
\Halmos
\endproof

\proof{Proof of Theorem \ref{q_error_3}.}
Without loss of generality, we first consider centralized reward $u_t:=u(s_{t+1},a_t,s_t)$, where $\mu_t:=\mu(s_{t+1},a_t,s_t)=0$, such that $P(|u_t|\geq \xi)\leq 2\exp(-c\xi^2)$. From the general Hoeffding’s inequality \citep{vershynin2018high}, for any $\xi>0$, $\tau^T$ and its subsequence $\tau^l\subseteq\tau^T$, $l\leq T$, we have
\begin{align*}
    P(|U(\tau^T)-U(\tau^l)|\geq \xi) = P(|\sum_{t=l}^{T-1} \eta^t u_t|\geq \xi)
    \leq 2\exp\left(-\frac{C_H\xi^2}{K^2\sum_{t=l}^{T-1} \eta^{2t} }\right)
    \leq 2\exp\left(-\frac{C_H(1-\eta^2)\xi^2}{K^2\eta^{2l}}\right),
\end{align*}
where $C_H>0$ is an absolute constant independent of other parameters, $K = \max_t \Vert u_t-\mu_t \Vert_{\psi_2} $ and
$\Vert Z \Vert_{\psi_2} = \inf\{\xi>0:\ \mathbb{E}\exp(Z^2/\xi)\leq 2\}$. The norm $\Vert \cdot \Vert_{\psi_2}$ is finite if and only if $Z$ is sub-Gaussian, which implies $K$ is finite.

Let $U(\tau^T)=X+Y$, where $X:=U(\tau^l)$, $Y:=U(\tau^T)-U(\tau^l)=\sum_{t=l}^{T-1} \eta^t u_t$. Thus, $q^l(\alpha;\theta)$ and $q^T(\alpha;\theta)$ are the $\alpha$-quantiles of $F_X(\cdot;\theta)$ and $F_{X+Y}(\cdot;\theta)$ respectively. For $\xi\gg0$, such that $2\exp\left(\frac{C_H(1-\eta^2)\xi^2}{K^2}\right)\geq1$, denote the double truncated random variable of $Y$ with interval $[-\eta^l\xi,\eta^l\xi]$ as $\hat{Y}$ and the $\alpha$-quantile of $X+\hat{Y}$ as $\hat{q}^T(\alpha;\theta,\eta^l\xi)$, where $\hat{Y}$ has a density given by
$$\hat{f}_{Y}(y;\theta,\eta^l\xi) = \frac{f_{Y}(y;\theta)\mathbf{1}\{|y|\leq \eta^l\xi\}}{F_{Y}(\eta^l\xi;\theta) - F_{Y}(-\eta^l\xi;\theta)}.$$
The boundedness of $\hat{Y}$ implies $|q^l(\alpha;\theta)-\hat{q}^T(\alpha;\theta,\eta^l\xi)|\leq \eta^l\xi$.
\begin{figure}[t]
	\centering
	\includegraphics[scale=0.6]{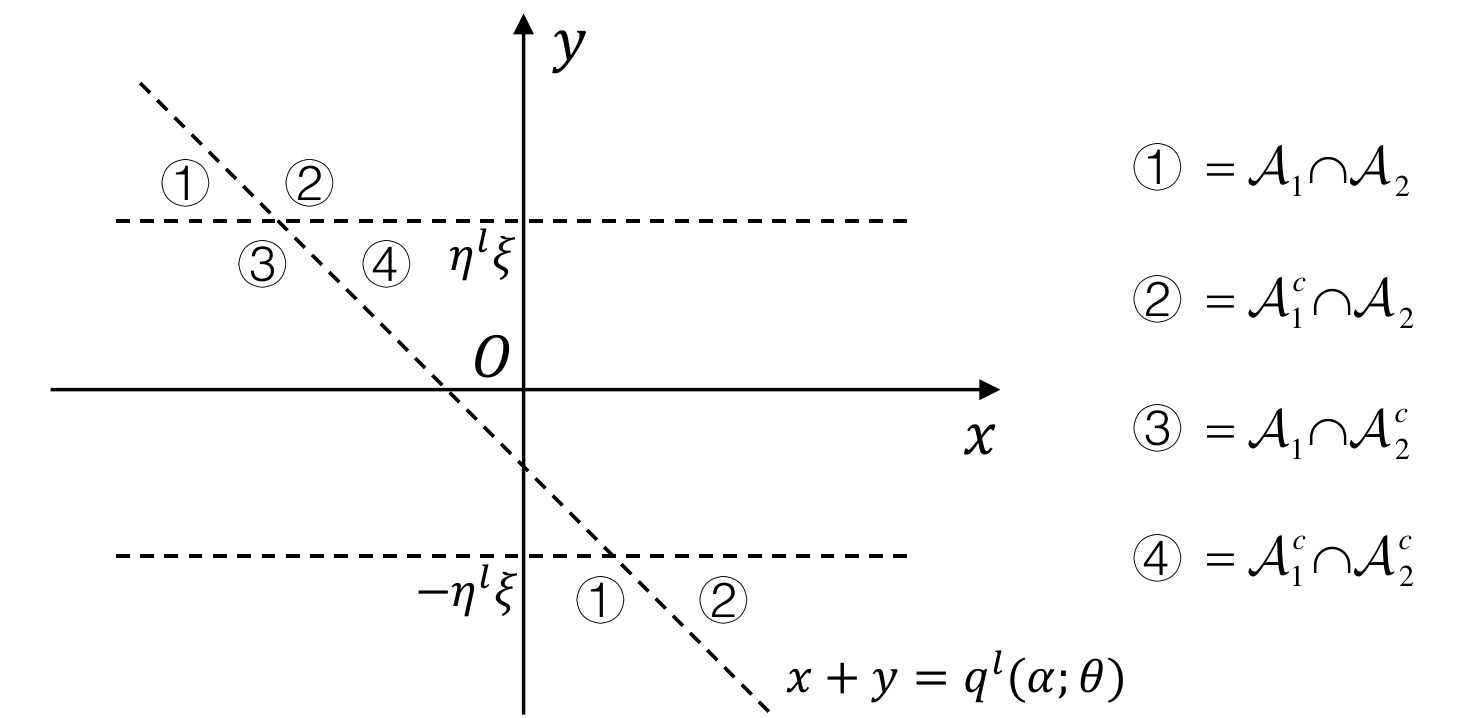}
	\caption{Division of the (x,y) space.}
	\label{fig:reward_q}
\end{figure}
By $C^1$-continuity of reward distribution function, there exists a unique
$\alpha':=\hat{F}_{X+Y}(q^T(\alpha;\theta);\theta,\eta^l\xi)$, i.e., $q^T(\alpha;\theta)=\hat{q}^T(\alpha';\theta,\eta^l\xi)$.
Let $\mathcal{A}_1:=\{X+Y\leq q^l(\alpha;\theta)\}$ and $\mathcal{A}_2:=\{|Y|\geq \eta^l\xi\}$.
The image space of $X$ and $Y$ is divided as shown in Figure.\ref{fig:reward_q}. Hence, by the definition of $\alpha'$, we have
$$\alpha'=\frac{P(\{X+Y\leq q^T(\alpha;\theta)\}\ \cap\ \{|Y|<\eta^l\xi\})}{P(|Y|<\eta^l\xi)}=\frac{P(\mathcal{A}_1\cap \mathcal{A}_2^c)}{P(\mathcal{A}_1\cap \mathcal{A}_2^c)+P(\mathcal{A}_1^c\cap \mathcal{A}_2^c)}
=\frac{1}{1+ {P(\mathcal{A}_1^c\cap \mathcal{A}_2^c)}/{P(\mathcal{A}_1\cap \mathcal{A}_2^c)}}.$$
By noticing
\begin{align*}
    \left\{
    \begin{array}{ll}
       P(\mathcal{A}_1\cap \mathcal{A}_2)+P(\mathcal{A}_1\cap \mathcal{A}_2^c)=\alpha\\
       P(\mathcal{A}_1\cap \mathcal{A}_2)+P(\mathcal{A}_1^c\cap \mathcal{A}_2)+P(\mathcal{A}_1\cap \mathcal{A}_2^c)+P(\mathcal{A}_1^c\cap \mathcal{A}_2^c) = 1\\
       P(\mathcal{A}_1\cap \mathcal{A}_2)+P(\mathcal{A}_1^c\cap \mathcal{A}_2)=P(\mathcal{A}_2)\leq 2\exp(-c_1\xi^2)
    \end{array}\right.,
\end{align*}
where $c_1=C_H(1-\eta^2)K^{-2}>0$,
we have $\frac{P(\mathcal{A}_1^c\cap \mathcal{A}_2^c)}{{P(\mathcal{A}_1\cap \mathcal{A}_2^c)}}\in\left[\frac{1-\alpha-2\exp(-c_1\xi^2)}{\alpha},\frac{1-\alpha}{\alpha-2\exp(-c_1\xi^2)}\right].$ Therefore, we can obtain $|\alpha-\alpha'|\leq\frac{ \max\{\alpha,1-\alpha\} }{2\exp(c_1\xi^2)-1}$.
Using the Lagrange's mean value theorem, we have
\begin{align*}
    |q^T(\alpha;\theta) - \hat{q}^T(\alpha;\theta,\eta^l\xi)|
    &= |\hat{q}^T(\alpha';\theta,\eta^l\xi)- \hat{q}^T(\alpha;\theta,\eta^l\xi)|
    = \left|\hat{F}_{X+Y}^{-1}(\alpha';\theta,\eta^l\xi) - \hat{F}_{X+Y}^{-1}(\alpha;\theta,\eta^l\xi)\right|\\
    & = \frac{|\alpha-\alpha'|}{\hat{f}_{X+Y}(r;\theta,\eta^l\xi)}
    \leq \frac{|\alpha-\alpha'|}{f_{X+Y}(r;\theta)}
    \leq \frac{1}{f_{R^T}(r;\theta)} \cdot \frac{ \max\{\alpha,1-\alpha\} }{2\exp(c_1\xi^2)-1},
\end{align*}
where $r\in\mathcal{D}_l:=[\hat{q}^T(\alpha-|\alpha-\alpha'|;\theta,\eta^l\xi),\hat{q}^T(\alpha+|\alpha-\alpha'|;\theta,\eta^l\xi)]$. Since $f_{R^T}(\cdot;\theta)$ is continuous, then $f_{R^T}(\cdot;\theta)$ has a lower bound on $\mathcal{D}_l$. Let $C_f>0$ be the uniform lower bound of $f_{R^T}(\cdot;\theta)$ on a sufficiently large neighborhood of $q^T(\alpha;\theta)$ that contains $\mathcal{D}_l$, where $l\leq T$ and $\theta \in \Theta$. And we can obtain
\begin{align}
    \sup_{\theta\in\Theta}|q^l(\alpha;\theta)-q^T(\alpha;\theta)|
    \leq & \sup_{\theta\in\Theta}|q^l(\alpha;\theta)-\hat{q}^T(\alpha;\theta,\eta^l\xi)|+
    \sup_{\theta\in\Theta}|q^T(\alpha;\theta)-\hat{q}^T(\alpha;\theta,\eta^l\xi)| \nonumber\\
    &\leq \min_{\xi} \left(\eta^l\xi +\frac{c_2}{\exp\left(c_1\xi^2\right)-2}\right),\label{single_bound}
\end{align}
where $c_2=2 C_f^{-1}\max\{\alpha,1-\alpha\}$.
Since both terms in the third line of the inequality (\ref{single_bound}) are monotonic, the minimizer $\xi^{l,*}$ satisfies the first order condition
$\eta^l-\frac{2c_1c_2\xi\exp\left(c_2\xi^2\right)}{(\exp\left(c_2\xi^2\right)-2)^2}=0$. By plug-in, we further have
\begin{align}
    \sup_{\theta\in\Theta}|q^l(\alpha;\theta)-q^T(\alpha;\theta)|
    \leq \eta^l\xi^{l,*} + c_2 \sqrt{\frac{\eta^l\exp(-c_1(\xi^{l,*})^2)}{2c_1c_2\xi^{l,*}}}
    \leq c_3\eta^l + c_4\sqrt{\eta^l}, \label{error_subgaussian}
\end{align}
where $c_3=\max_{l\leq T}\xi^{l,*}$, $c_4=\max_{l\leq T} c_2 \sqrt{\frac{\exp(-c_1(\xi^{l,*})^2)}{2c_1c_2\xi^{l,*}}}$.
For rewards without zero mean, similar to the proof of Theorem \ref{q_error_2}, there will be an additional term $C_{\mu}\frac{\eta^l}{1-\eta}$ that can be absorbed by $c_3$ in the inequality (\ref{error_subgaussian}). Finally, in the same manner as in the proof Theorem \ref{q_error_1}, we have
\begin{align*}
    |q^T(\alpha;\theta^*)&-q^T(\alpha;\theta^{**})|
    \leq |q^T(\alpha;\theta^*)-\frac{1}{T-T_0+1}\bar{q}(\alpha;\theta^{**})|+ \frac{1}{T-T_0+1}\sum_{l=T_0}^T |q^l(\alpha;\theta^{**}) - q^T(\alpha;\theta^{**})|\\
    &\leq  \frac{2}{T-T_0+1}\sum_{l=T_0}^T (c_3\eta^l + c_4\sqrt{\eta^l} )
    \leq \frac{2}{T-T_0+1} \left(c_3\frac{\eta^{T_0}}{1-\eta} + c_4\frac{\sqrt{\eta^{T_0}}}{1-\sqrt{\eta}}\right) \rightarrow 0,
\end{align*}
as $T_0\rightarrow \infty$.
\Halmos
\endproof

\section{Supplements for Section \ref{Section Numerical Experiments}}\label{Appendix Numerical Experiments}

\subsection{Experiment Settings in “Zero Mean"}\label{Appendix Zero Mean}
Except for the last four parameters in Table \ref{table: exp1_hyper}, all parameters are shared by all algorithms. Note that the learning rate for SPSA is the same as the other algorithms, whereas the learning rate for SPSA+ is 50 and 100 times that of the other algorithms for the simple and hard examples, respectively.
\begingroup
\renewcommand{\arraystretch}{0.6}
\setlength{\extrarowheight}{-0.5pt}
\begin{table}[!hbp]
\centering
\caption{Standard hypherparameters of
five deep RL algorithms in Zero Mean examples.}
\label{table: exp1_hyper}
\begin{tabular}{ccc}
\toprule
Hyperparameter & Simple example & Hard example\\
\midrule
Network width & $8$-$8$ & $64$-$64$-$64$\\
Reward discount factor & $0.99$ & $0.99$\\
Learning rate & $1\times10^{-3}$ & $5\times10^{-4}$\\
Learning rate decay factor & $0.8$ & $0.8$\\
Learning rate decay interval & $2.5\times10^3$ episodes & $2.5\times10^3$ episodes\\
Clip parameter (PPO \& QPPO) & $0.2$ & $0.2$\\
Update interval (PPO \& QPPO) & $2\times10^3$ steps & $2\times10^3$ steps\\
Quantile learning rate (QPO \& QPPO) & $0.01$ & $0.001$\\
Truncated trajectory length (QPPO only) & $\{16,\cdots,20\}$ & $\{16,\cdots,20\}$\\
\bottomrule
\end{tabular}
\end{table}
\endgroup

\subsection{Experiment Settings in “Financial Investment"}\label{Appendix Financial Investment}
In both examples, the initial prices of all assets are set to $1$. The initial value allocation is randomly generated. And the transaction fee is $0.1\%$ of the trading value. The policy and baseline networks consist of three fully connected layer with a width of 64.
\begingroup
\renewcommand{\arraystretch}{0.6}
\setlength{\extrarowheight}{-0.5pt}
\begin{table}[htbp]
\centering
\caption{Exogenous parameters in perfectly hedgeable portfolio.}
\begin{tabular}{ccccc}
\toprule
\multirow{2}{*}{Asset} & \multirow{2}{*}{Drift $\mu$} & \multicolumn{3}{c}{Volatility $\Sigma$} \\ &  & 1& 2& 3\\ 
\midrule
1&  0.01& 0.01& 0& 0\\
2&  0.08& 0& 0.08& -0.08\\
3&  0.16& 0& -0.08& 0.08\\
\bottomrule
\end{tabular}
\end{table}
\endgroup
\begingroup
\renewcommand{\arraystretch}{0.6}
\setlength{\extrarowheight}{-0.5pt}
\begin{table}[htbp]
\centering
\caption{Exogenous parameters in imperfectly hedgeable portfolio.}
\begin{tabular}{ccccccc}
\toprule
\multirow{2}{*}{Asset} & \multirow{2}{*}{Drift $\mu$} & \multicolumn{5}{c}{Volatility $\Sigma$} \\ &  & 1& 2& 3& 4& 5\\ 
\midrule
1&  0.01& 0.01& 0& 0& 0& 0\\
2&  0.02& 0& 0.04& -0.055& 0& 0\\
3&  0.03& 0& -0.055& 0.09& 0& 0\\
4&  0.04& 0& 0& 0& 0.16& -0.19\\
5&  0.05& 0& 0& 0& -0.19& 0.25\\
\bottomrule
\end{tabular}
\end{table}
\endgroup
\begingroup
\renewcommand{\arraystretch}{0.6}
\setlength{\extrarowheight}{-0.5pt}
\begin{table}[htbp]
\centering
\caption{Hypherparameters of
PPO and QPPO in Portfolio Management examples.}
\begin{tabular}{cc}
\toprule
Hyperparameter & Value\\
\midrule
Reward discount factor & $0.99$\\
Learning rate & $2\times10^{-5}$\\
Learning rate decay factor & $0.9$\\
Learning rate decay interval & $1\times10^4$ episodes\\
Clip parameter & $0.2$\\
Update interval & $5\times10^3$ steps\\
Quantile learning rate (QPPO only) & $0.01$\\
Truncated trajectory length (QPPO only) & $\{91,\cdots,100\}$\\
\bottomrule
\end{tabular}
\end{table}
\endgroup

\subsection{Experiment Settings in “Inventory Management"}\label{Appendix Inventory Management}
In the uniform case, the customers' demand $q_t^0$ is uniformly sampled from $\{0,1,\cdots,20\}$. In the Merton case, the customers' demand is generated by a discretized version of Merton jump diffusion model in \cite{glasserman2004monte}, i.e.,
\begin{align*}
q_t^0 =\left\lfloor B \exp \left(J_t\right)\right\rfloor,\quad
J_t =J_{t-1}+(\mu-\frac{1}{2} \sigma^2)+\sigma Z_t+a N_t+b \sqrt{N_t} Z_t^{\prime},
\end{align*}
where $Z_t$ and $Z_t^{\prime}$ are independently generated from $\mathcal{N}(0,1)$, $N_t$ is independently generated from $\text{Poisson}(15)$, $\mu=5\times10^{-5}$, $\sigma=0.01$, $a=0$, $b=0.01$, and $B=10$. In the periodical case, the customers' demand is generated by
\begin{align*}
q_t^0 = x_t + [(t+t_0)\text{ mod }T],
\end{align*}
where $x_t$ is uniformly sampled from $\{0,1,\cdots,7\}$, $t_0=6$, and $T=15$.
\begingroup
\renewcommand{\arraystretch}{0.6}
\setlength{\extrarowheight}{-0.5pt}
\begin{table}[htbp]
\centering
\caption{Exogenous parameters in single-echelon supply chain system.}
\begin{tabular}{cccccc}
\toprule
Echelon & Lead time $L$ & Price $p$ & Holding cost $h$ & Penalty of lost sale $l$ & Initial inventory $I_0$\\
\midrule
1 & 3 & 2 & 0.15 & 0.10 & 10\\
2 & / & 1.5 & / & / & $+\infty$ \\
\bottomrule
\end{tabular}
\end{table}
\endgroup
\begingroup
\renewcommand{\arraystretch}{0.6}
\setlength{\extrarowheight}{-0.5pt}
\begin{table}[htbp]
\centering
\caption{Exogenous parameters in multi-echelon supply chain system.}
\begin{tabular}{cccccc}
\toprule
Echelon & Lead time $L$ & Price $p$ & Holding cost $h$ & Penalty of lost sale $l$ & Initial inventory $I_0$\\
\midrule
1 & 2 & 2   & 0.2 & 0.125 & 10\\
2 & 3 & 1.5 & 0.15 & 0.1 & 10 \\
3 & 5 & 1   & 0.1 & 0.075 & 10\\
4 & / & 0.5 & / & / & $+\infty$ \\
\bottomrule
\end{tabular}
\end{table}
\endgroup

In the single-echelon example, the policy and baseline networks consist of a temporal convolutional layer with a kernel size of 3 and 64 output channels, and an output fully connected layer.
In the multi-echelon example, the policy and baseline networks include a convolutional block composed of a temporal convolutional layer with a kernel size of 3 and 32 output channels, and another one with 64 output channels, and three parallel output fully connected layers.
\begingroup
\renewcommand{\arraystretch}{0.6}
\setlength{\extrarowheight}{-0.5pt}
\begin{table}[htbp]
\centering
\caption{Hypherparameters of
PPO and QPPO in Inventory Management examples.}
\begin{tabular}{ccc}
\toprule
Hyperparameter & Sinle-echelon example & Multi-echelon example\\
\midrule
Reward discount factor & $0.99$ & $0.99$\\
Learning rate & $1\times10^{-4}$ & $1\times10^{-4}$\\
Learning rate decay factor & $0.9$ & $0.9$\\
Learning rate decay interval & $5\times10^3$ episodes & $2\times10^4$ episodes\\
Clip parameter & $0.2$ & $0.2$\\
Update interval & $2\times10^3$ steps & $1\times10^4$ steps\\
Quantile learning rate (QPPO only) & $2.0$ & $0.2$\\
Truncated trajectory length (QPPO only) & $\{46,\cdots,50\}$ & $\{91,\cdots,100\}$\\
\bottomrule
\end{tabular}
\end{table}
\endgroup

\end{APPENDICES}

\end{document}